\documentclass{article}
\usepackage{microtype}
\usepackage{graphicx}  
\usepackage{booktabs}
\usepackage{hyperref}

\usepackage[preprint]{main}
\usepackage{amsmath}
\usepackage{amssymb}
\usepackage{enumitem}
\usepackage{mathtools}
\usepackage{amsthm}

\usepackage[capitalize,noabbrev]{cleveref}

\theoremstyle{plain}
\newtheorem{theorem}{Theorem}[section]

\newtheorem{corollary}[theorem]{Corollary}
\theoremstyle{definition}
\newtheorem{definition}[theorem]{Definition}

\theoremstyle{remark}

\usepackage{algorithm}
\usepackage{algorithmic}
\usepackage{xcolor} 
\usepackage{amsmath} 
\usepackage[utf8]{inputenc}
\usepackage{hyperref}
\usepackage{url}
\usepackage{multirow}
\usepackage{subcaption}
\usepackage{caption}
\usepackage[table]{xcolor} 
\usepackage[normalem]{ulem} 
\usepackage[dvipsnames]{xcolor}
\usepackage{xurl}
\usepackage{longtable}
\usepackage{array} 
\usepackage{tabularx}
\usepackage{ltablex}
\keepXColumns
\definecolor{blockgray}{RGB}{248, 248, 250} 
\newcommand{\best}[1]{\textcolor{WildStrawberry!75}{\textbf{#1}}}
\newcommand{\secondbest}[1]{\textcolor{CornflowerBlue}{\uline{\textbf{#1}}}}
\definecolor{softpink}{RGB}{232, 116, 145} 
\definecolor{paleorange}{RGB}{230, 130, 110} 
\newcommand{\inc}[1]{{\tiny\color{paleorange} $\uparrow$#1}}

\icmltitlerunning{Breaking the Regional Barrier: Inductive Semantic Topology Learning for Worldwide Air Quality Forecasting}

\begin{document}

\twocolumn[
\icmltitle{Breaking the Regional Barrier: Inductive Semantic Topology Learning for Worldwide Air Quality Forecasting}

\icmlsetsymbol{equal}{*}

\begin{icmlauthorlist}
\icmlauthor{Zhiqing Cui}{equal,hkust}
\icmlauthor{Siru Zhong}{equal,hkust}

\icmlauthor{Ming Jin}{griffith}
\icmlauthor{Shirui Pan}{griffith}
\icmlauthor{Qingsong Wen}{squirrel}
\icmlauthor{Yuxuan Liang}{hkust}
\end{icmlauthorlist}

\icmlaffiliation{hkust}{The Hong Kong University of Science and Technology (Guangzhou), China}
\icmlaffiliation{griffith}{Griffith University, Australia}
\icmlaffiliation{squirrel}{Squirrel AI, USA}

\icmlcorrespondingauthor{Yuxuan Liang}{yuxliang@outlook.com}

\icmlkeywords{Air Quality, Spatio-Temporal Forecasting, AI4Earth}

\vskip 0.3in
]

\printAffiliationsAndNotice{\icmlEqualContribution}
\begin{abstract}
Global air quality forecasting grapples with extreme spatial heterogeneity and the poor generalization of existing transductive models to unseen regions. To tackle this, we propose \textbf{OmniAir}, a semantic topology learning framework tailored for global station-level prediction. By encoding invariant physical environmental attributes into generalizable station identities and dynamically constructing adaptive sparse topologies, our approach effectively captures long-range non-Euclidean correlations and physical diffusion patterns across unevenly distributed global networks. We further curate \textbf{WorldAir}, a massive dataset covering over 7,800 stations worldwide. Extensive experiments show that OmniAir achieves state-of-the-art performance against 18 baselines, maintaining high efficiency and scalability with speeds nearly 10 times faster than existing models, while effectively bridging the monitoring gap in data-sparse regions.
\end{abstract}

\section{Introduction}

Air pollution, a pressing global crisis, impairs human health and hinders sustainable urban development. According to the latest WHO monitoring data \cite{who2022billions}, 99\% of the global population breathes air that exceeds WHO air quality guidelines, with populations in low- and middle-income countries facing the highest exposure risks. Thus, developing high-precision, globally scalable predictive models is critical for protecting public health \cite{horn2024air,liang2023airformer,kumar2023critical}. While deep learning has advanced substantially in environmental science, existing methods suffer from two critical limitations \cite{asudani2023impact,han2023survey}.

\begin{figure}[ht]
    \centering
    \includegraphics[width=\linewidth]{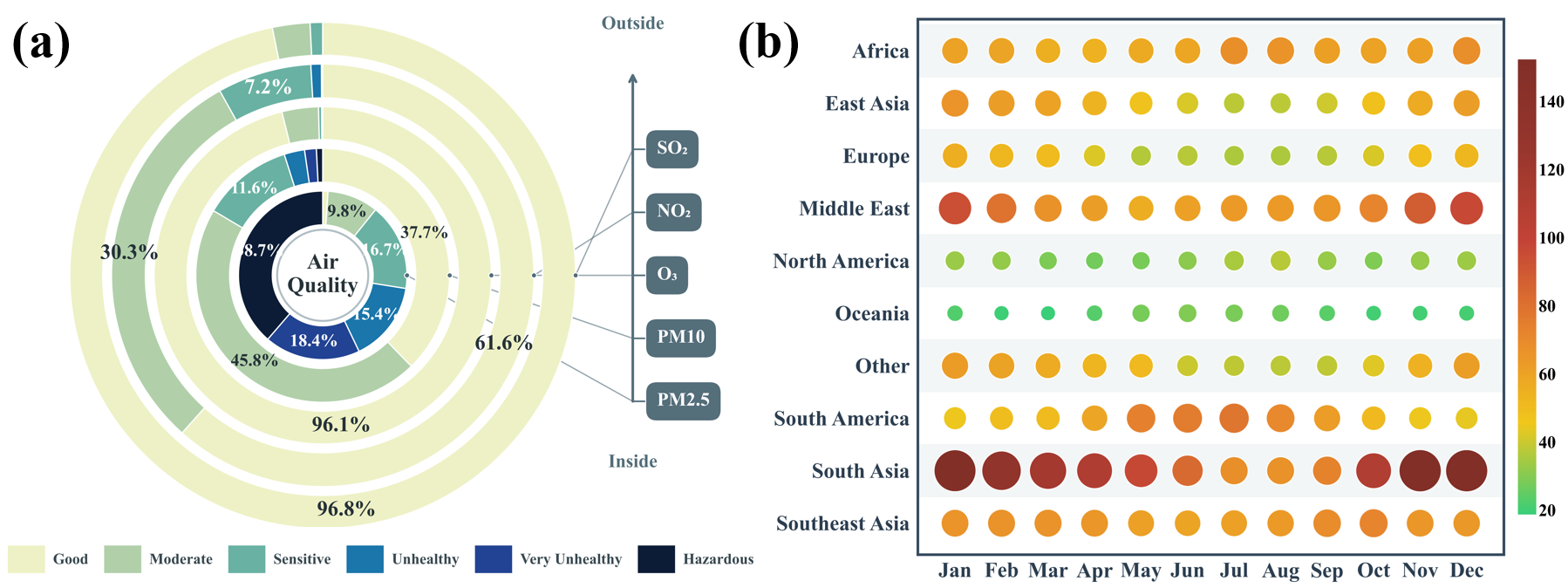}
    \caption{(a) Air quality level distribution across pollutants. (b) Regional-monthly PM$_{2.5}$ concentrations revealing cross-regional and seasonal heterogeneity.}
    \vspace{-1em}
    \label{fig:temporal_patterns}
\end{figure}
\textit{First}, the \textbf{single-pollutant} limitation: existing models focus predominantly on individual pollutants (e.g., $\text{PM}_{2.5}$) \cite{wang2020pm2,zhang2022deep} in local regions, ignoring the global transport of air pollutants and their complex chemical couplings (e.g., $\text{PM}_{10}$, $\text{CO}$, $\text{NO}_2$, $\text{SO}_2$, $\text{O}_3$). Since the Air Quality Index (AQI) is a composite metric integrating multiple pollutants \cite{kumar2022critical}, single-indicator analyses fail to support comprehensive air quality management.

\textit{Second}, the \textbf{single-region} perspective: most models are tailored for data-rich regions but fail to transfer to sparse regions with diverse environmental contexts \cite{zhang2022deep}, leading to redevelopment costs and impeding global governance deployment. Notably, \textit{global-scale air quality forecasting remains largely underexplored} \cite{seinfeld2016atmospheric,lu2025causality}. Local models cannot capture the global interconnectivity of atmospheric phenomena and suffer poor cross-context generalization, stemming from three critical performance-limiting challenges:

\begin{table*}[h!]
\caption{Comparison of spatial scale, dataset size, modeling paradigm and computational complexity for air quality forecasting models. Traditional $\mathcal{O}(N^2)$/$\mathcal{O}(N \log N)$ methods are computationally prohibitive at global scale ($N \approx 8,000$), while our OmniAir achieves linear complexity $\mathcal{O}(N \cdot K)$ ($K$: neighbors) for efficient global station-level prediction.}
\vspace{-1em}
\label{tab:comparison}
\begin{center}
\begin{small}
\begin{sc}
\renewcommand{\arraystretch}{1}
\resizebox{\textwidth}{!}{%
\begin{tabular}{llc|lll}
\toprule
\multicolumn{3}{c|}{\textbf{Dataset \& Spatial Scale}} & \multicolumn{3}{c}{\textbf{Modeling Paradigm \& Complexity}} \\
\midrule
\textbf{Scope} & \textbf{Dataset} & \textbf{Nodes ($N$)} & \textbf{Paradigm} & \textbf{Representative Models} & \textbf{Complexity} \\
\midrule
City & London & 13 & Spectral GNN & GWNet \cite{wu2019gwnet} & $\mathcal{O}(N^2)$ \\
Nation & KnowAir & 184 & Recurrent GNN & PM2.5-GNN \cite{wang2020pm2} & $\mathcal{O}(N^2)$ \\
Nation & LargeAQ & 1,341 & Linear Trans. & Informer \cite{zhou2021informer} & $\mathcal{O}(N \log N)$ \\
Global & \textbf{WorldAir (Ours)} & \textbf{7,861} & Standard Trans. & AirFormer \cite{liang2023airformer} & $\mathcal{O}(N^2)$ \\
\midrule
\midrule
\multicolumn{3}{l|}{\textbf{Proposed: OmniAir (Ours)}} & \multicolumn{2}{l}{\textbf{Inductive Semantic Topology}} & \textbf{$\mathcal{O}(N \cdot K)$} \\
\bottomrule
\end{tabular}}
\vspace{-1em}
\end{sc}
\end{small}
\end{center}
\end{table*}

\vspace{-0.5em}
\begin{itemize}[leftmargin=*]
    \item \textbf{Cross-Spatial Heterogeneity.} Global spatial dependencies are far more complex than local ones, exhibiting non-Euclidean characteristics \cite{he2022spatial}. Driven by large-scale atmospheric circulation, source-receptor regions thousands of kilometers apart may show strong temporal correlations \cite{wang2023review}, while geographically adjacent stations across complex terrain can exhibit entirely distinct pollution patterns \cite{uno2009asian}.
 
    \item \textbf{Cross-Temporal Dynamics.} Unlike static sensor networks, global monitoring systems exhibit high dynamics \cite{yao2022wild}. Data availability is not consistent year-round, and pollutant distributions evolve temporally. A robust model must accommodate such temporal irregularities without perfect data continuity, capturing intrinsic temporal patterns instead of memorizing fixed sequences.
    
    \item \textbf{Cross Spatio-Temporal (ST) Distribution Shifts.} The core challenge is the dynamic ST coupling \cite{ma2025causal}. Air quality topology is inherently dynamic: wind variations instantly reconfigure dependency graphs, invalidating static adjacency matrices (widely used in traffic forecasting) \cite{marobust,wang2023pattern}. Besides, global modeling confronts severe distribution shifts, requiring models trained on historical data from data-rich regions to generalize to future scenarios and data-sparse regions \cite{tian2024air,karniadakis2021physics}.
\end{itemize}
\vspace{-0.5em}

To address these challenges as empirically illustrated in Figure \ref{fig:temporal_patterns}, we present \textbf{OmniAir}, an inductive framework approximating the latent causal structure of global atmospheric dynamics. The pipeline begins by projecting raw physical attributes into semantic identities, enabling the model to decouple station representations from transductive indices and support zero-shot generalization. Guided by these inductive identities, the framework reconstructs a differentiable transport manifold, dynamically connecting local geographic neighbors with long-range semantic counterparts. Upon this learned topology, we implement a specialized propagation mechanism capturing the physical duality of pollutant diffusion and source generation. Collectively, this end-to-end design empowers OmniAir to model global atmospheric dynamics, transfer cross-region knowledge, and maintain scale efficiency. As quantified in \autoref{tab:comparison}, our approach achieves linear computational complexity $\mathcal{O}(N \cdot K)$, overcoming the scalability bottleneck of traditional quadratic-complexity models and enabling efficient global-scale prediction across 7,861 stations.

Our main contributions are summarized as follows:

\vspace{-0.5em}
\begin{itemize}[leftmargin=*]
    \item \textbf{First Global-Scale Air Quality Dataset.} We develop WorldAir, the largest open-source air quality dataset (7,800+ stations across varied climate/development levels), integrating ground truth  and geospatial context as a rigorous benchmark for global generalization evaluation.
    \vspace{-0.5em}

    \item \textbf{Physics-Informed Inductive Graph Framework.} We propose OmniAir, an inductive graph architecture, enables spatial extrapolation to data-sparse regions; its physics-inspired components address global density heterogeneity, capture pollutant diffusion/source generation duality, and solve long-range non-Euclidean correlations.  
    \vspace{-0.5em}
    
    \item \textbf{Superior Performance and Efficiency.} Extensive experiments show OmniAir outperforms 18 SOTA baselines, with high scalability and inference speeds nearly \textbf{10×} faster than complex baselines, offering a practical solution for real-time global monitoring.
\end{itemize}
\vspace{-0.5em}

\section{Preliminaries}

\subsection{Problem Statement}

Let $\mathcal{G}=(\mathcal{V},\mathcal{E})$ be a spatial graph where $\mathcal{V}$ denotes $N$ air quality monitoring stations and $\mathcal{E}$ encodes spatial dependency edges; at time step $t$, pollutant concentrations $\mathbf{X}_t\in\mathbb{R}^{N\times c}$ (with $c$ the number of measured pollutants) are observed, and each station $i\in\mathcal{V}$ is assigned static attributes $\mathbf{s}_i$ capturing geographic coordinates, elevation, climate zone and other environmental features. Given the historical observation sequence $\mathcal{X}_{h} = \{\mathbf{X}_{t-T+1}, \dots, \mathbf{X}_t\}$ over $T$ past time steps and static features $\mathbf{S} = \{\mathbf{s}_1, \dots, \mathbf{s}_N\}$, air quality forecasting aims to learn a mapping $\mathcal{F}$ that predicts future concentrations for the next $\tau$ steps:
\begin{equation}
    \hat{\mathbf{Y}}_{t+1:t+\tau} = \mathcal{F}(\mathcal{X}_{h}, \mathbf{S}, \mathcal{G}), \quad \hat{\mathbf{Y}} \in \mathbb{R}^{\tau \times N \times c}
\end{equation}

\subsection{Related Work}

\paragraph{Traditional Methods for Air Quality Prediction.} 
Air quality prediction is a fundamental task in environmental monitoring. However, traditional numerical simulation methods based on atmospheric physics require extensive domain knowledge and computational resources, limiting their applicability to real time global monitoring~\cite{daly2007air,li2023physics}. Although recent large scale weather foundation models~\cite{bi2023accurate,lam2023learning} have demonstrated remarkable capabilities in global meteorological forecasting, these approaches predominantly rely on coarse grained grid data~\cite{gao2025oneforecast,brown2025alphaearth}. This reliance inevitably smooths out high frequency spatial heterogeneity, rendering them less effective for capturing localized pollution events (such as traffic related emissions monitored at specific sites) compared to station based methods~\cite{sun2025can,yang2025local}.

\paragraph{Spatio-Temporal Deep Models.} Deep learning has greatly advanced air quality forecasting, with spatio-temporal graph neural networks (STGNNs) as the dominant paradigm. STGNNs model monitoring stations as nodes, capturing temporal dependencies via recurrent networks or Transformers and spatial correlations through graph convolutions \cite{wang2020pm2}, fueling rapid progress in station-level prediction. Representative works embed meteorological knowledge into Transformers \cite{ma2025causal,hettige2024airphynet,zhangeulerian,tian2024air}; AirFormer \cite{liang2023airformer} enhances fine-grained prediction with decoupled determinism for practical challenges. Recent studies further move beyond pure data-driven frameworks by integrating domain knowledge \cite{chen2023gagnn,feng2024spatio,xu2023dynamic}, and AirRadar \cite{wang2025airradar} infers air quality in unmonitored regions using learnable mask tokens. By contrast, our \textbf{OmniAir} pioneers an inductive semantic topology for zero-shot generalization to unseen regions, and achieves linear complexity (\autoref{tab:comparison})—a critical advantage over traditional quadratic-complexity models—enabling efficient, scalable global station-level prediction via physical diffusion-source duality modeling.

\section{Methodology}

\subsection{Model Overview}

\begin{figure}[h!]
    \centering

    \includegraphics[width=0.5\textwidth]{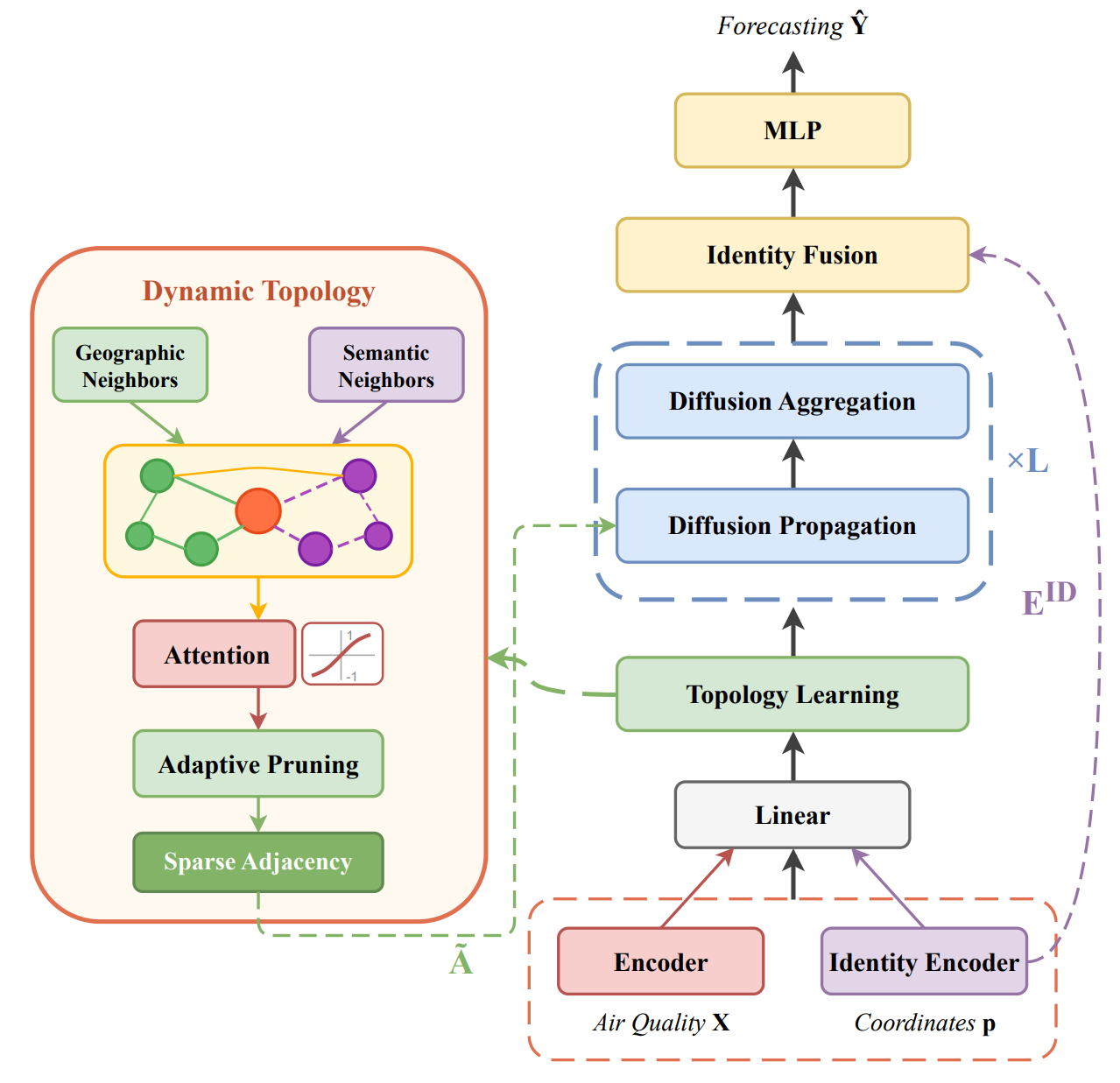}
    \caption{Overall framework of OmniAir, which models the spatio-temporal evolution of atmospheric pollutants via inductive station identities and adaptive sparse topologies.}
    \label{fig:framework} 

\end{figure}

As shown in Fig.~\ref{fig:framework}, OmniAir approximates the latent causal structure of global atmospheric dynamics via an inductive pipeline: physical attributes are projected into invariant semantic identities for zero-shot generalization, guiding the reconstruction of a differentiable manifold modeling local and global dependencies; atop this topology, the model propagates to simulate pollutant dispersion and source generation, ensuring robust inference under sparse setting.
\vspace{0.5em}

\subsection{Inductive Semantic Identity Encoder}
\label{sec:isie}

\begin{figure}[h]

    \centering
    \includegraphics[width=0.5\textwidth]{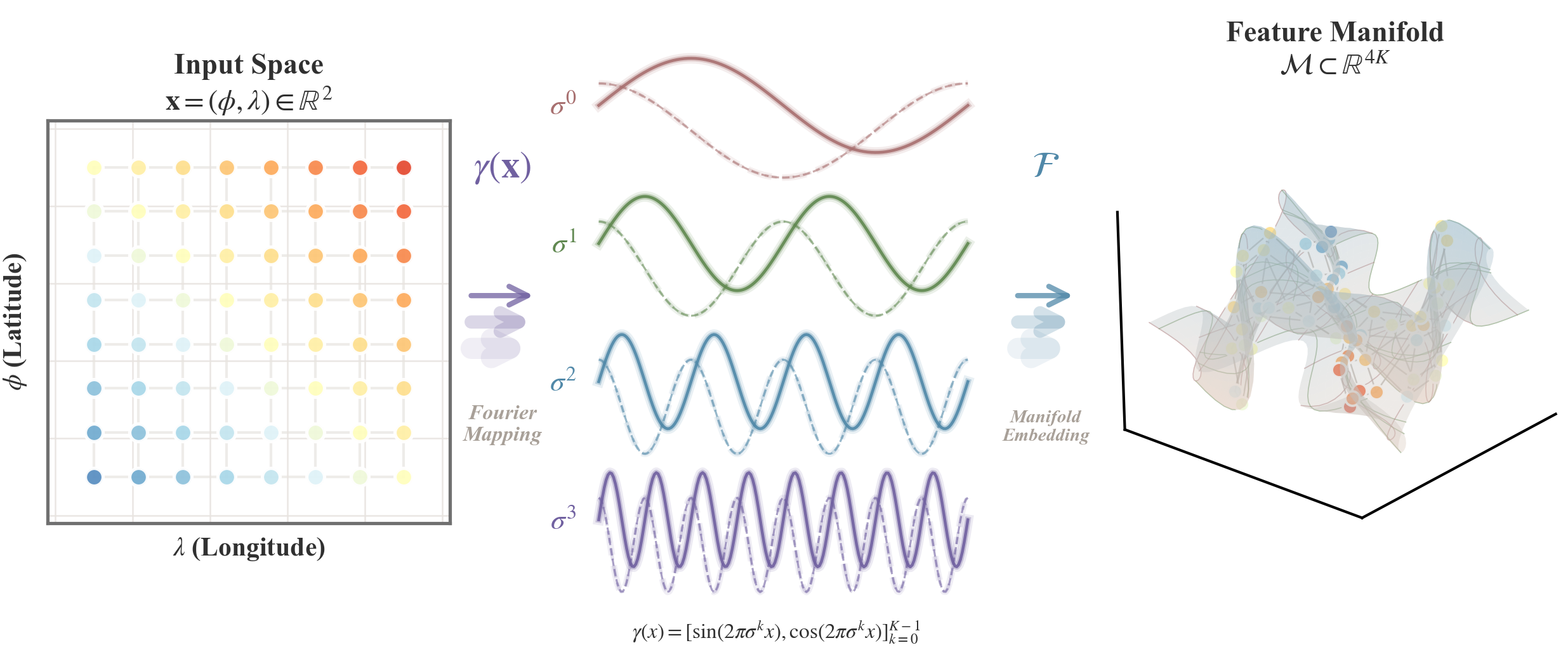}
    \caption{Proposed encoding converts raw geographic coordinates to semantically rich, high dimensional representations for inductive generalization to unseen stations.}
    \label{fig:fourier_encoding}
\end{figure}

Most spatio-temporal forecasting methods rely on transductive learnable node embeddings $\mathbf{E} \in \mathbb{R}^{N \times D}$ to capture station-specific characteristics \cite{huang2025std,uddinscnode,shao2022spatial}, which inherently limits generalization. As shown in Fig.~\ref{fig:fourier_encoding}, we adopt Fourier feature mapping to encode geographic coordinates into high-dimensional representations that capture multi-scale spatial patterns.

To enable inductive inference, our encoder computes a continuous station identity $\mathbf{e}_i^{ID}$ by synthesizing diverse observable metadata instead of memorizing discrete indices. Specifically, we construct a composite representation fusing: (1) high-frequency spatial variations from the deterministic multi-scale Fourier mapping $\gamma(\mathbf{p}_i) = \bigoplus_{j=0}^{M-1} [\sin(2\pi f_j \tilde{\mathbf{p}}_i), \cos(2\pi f_j \tilde{\mathbf{p}}_i)]$; (2) local environmental context $\mathbf{f}_i^{nbr}$ from geographic neighbors, including historical moments $\mu_i^{nbr}, \sigma_i^{nbr}$ and directional centroid offsets $\delta_i^{c}$; (3) physical station attributes $\mathbf{f}_i^{geo}$ (e.g., elevation, climate zones). These multi-source features are projected into a unified embedding space via:
\vspace{0.5em}
\begin{equation}
    \mathbf{e}_i^{ID} = \operatorname{MLP}\left( \gamma(\mathbf{p}_i) \oplus \mathbf{f}_i^{nbr} \oplus \mathbf{f}_i^{geo} \right)
\end{equation}
This formulation relies exclusively on observable attributes, enabling zero-shot initialization for newly deployed stations.

\subsection{Inductive Representation Learning}
\begin{figure}[htbp]
    \centering
    \includegraphics[width=\linewidth]{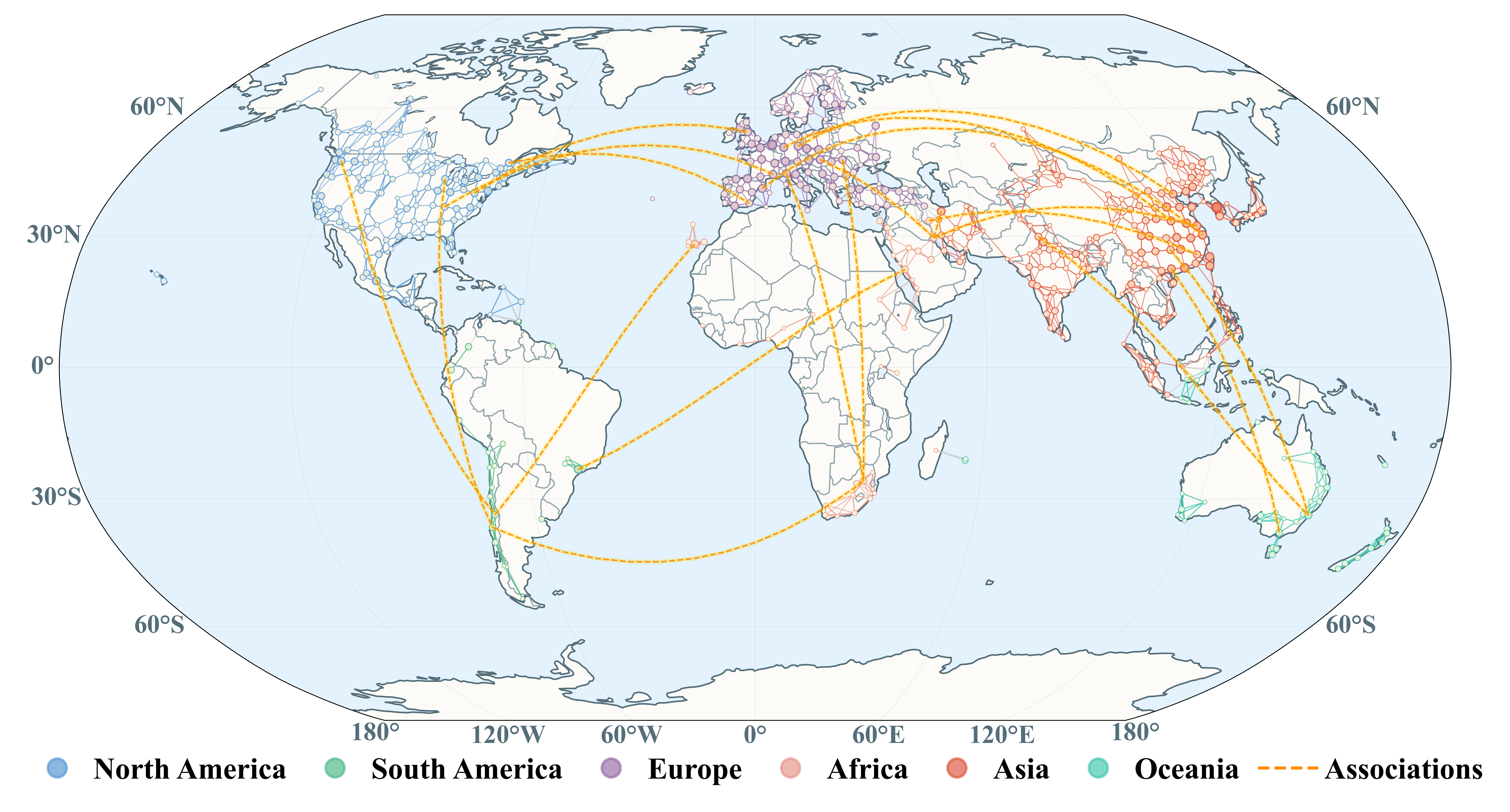}
    \caption{Visualization of learned global spatial associations across different continents.}
    \label{fig:connect}
\end{figure}

To address scalability and coverage limitations, we propose the Dynamic Sparse Topology Generator. Allowing the model to capture long range functional correlations between distant but ecologically similar regions and construct adaptive sparse graphs connecting semantically similar nodes across continents, as illustrated in Figure~\ref{fig:connect}. This approach transcends local geographic constraints, enabling effective knowledge transfer from data rich to data sparse regions.

\paragraph{Hybrid Neighbor Graph Construction. }

We construct a hybrid neighbor set fusing geographic proximity and semantic similarity. For geographic neighbors, we identify the $k_{geo}$ nearest stations via the Haversine distance function:
\begin{equation}
    d_{ij}^{geo} = \text{Haversine}(\mathbf{p}_i, \mathbf{p}_j)
\end{equation}
where $\mathbf{p}_i$ and $\mathbf{p}_j$ denote the coordinate vectors (latitude and longitude) of station $i$ and $j$.

For semantic neighbors, we leverage the identity embeddings to select stations with the highest semantic similarity:
\begin{equation}
    \mathcal{N}_i^{sem} = \operatorname{TopK}_{j \neq i}\left( -\|\mathbf{e}_i^{ID} - \mathbf{e}_j^{ID}\|_2 \right)
\end{equation}
The semantic search operates in the learned embedding space, clustering stations with similar profiles.

\paragraph{Dynamic Edge Weight Learning.} Distance-based static edge weights fail to capture the time-varying atmospheric transport, as wind patterns, seasonal circulation and synoptic weather systems constantly reshape inter-station effective connectivity. We thus learn dynamic edge weights adaptive to input-encoded current atmospheric conditions.

Given node features $\mathbf{H} \in \mathbb{R}^{B \times N \times D}$ at the current time step, we first project them through a linear layer and compute pairwise attention scores. For edge $(i, j) \in \mathcal{E}$, the dynamic attention weight is computed as:
\begin{equation}
    \alpha_{ij} = \tanh\left( \mathbf{a}^T \cdot \text{LeakyReLU}\left( \mathbf{W}_e \left[\mathbf{h}_i \| \mathbf{h}_j\right] \right) \right)
\end{equation}
where $\mathbf{W}_e \in \mathbb{R}^{D' \times 2D}$ projects the concatenated features, $\mathbf{a} \in \mathbb{R}^{D'}$ computes the scalar attention, and $\tanh$ activation allows both positive and negative attention values.

We initialize edge weights with a Gaussian kernel based on geographic distance: $w_{ij}^{static} = \exp(-d_{ij}^2 / 2\kappa^2)$, where $\kappa$ controls the decay rate. The final edge weight fuses static and dynamic components through a learned gate:
\begin{equation}
    g_{ij} = \text{sigmoid}\left( \mathbf{w}_g^T \left[\mathbf{h}_i \| \mathbf{h}_j \| w_{ij}^{static}\right] + b_g \right)
\end{equation}
\begin{equation}
    w_{ij}^{dyn} = g_{ij} \cdot w_{ij}^{static} + (1 - g_{ij}) \cdot \alpha_{ij}
\end{equation}

The gating mechanism allows the model to rely more on static distance weights when dynamic features are uninformative, while adaptively shifting to attention based weights when temporal patterns provide stronger signals.

\paragraph{Adaptive Sparse Edge Learning. }

To address the extreme density heterogeneity in global networks, we introduce a differentiable pruning mechanism that learns a node specific effective neighborhood size. Unlike fixed k approaches, we predict a continuous truncation threshold $\beta_i \in (0, k_{max})$ for each node via $\beta_i = k_{max} \cdot \sigma(\text{MLP}(\mathbf{h}_i))$. We then apply a soft mask $m_{ij}$ based on the attention rank $r_{ij}$ of each edge:
\begin{equation}
    m_{ij} = \text{sigmoid}\left( -\eta \cdot (r_{ij} - \beta_i) \right)
\end{equation}
where $\eta$ controls the cut off steepness. This mechanism acts as a learnable gate, retaining edges where the rank $r_{ij} < \beta_i$ while suppressing noise from lower ranked connections.

The final topology is obtained by re-normalizing the masked dynamic weights:
\begin{equation}
    \tilde{w}_{ij} = \frac{w_{ij}^{dyn} \cdot m_{ij}}{\sum_{k \in \mathcal{N}_i} w_{ik}^{dyn} \cdot m_{ik} + \epsilon}
\end{equation}
This formulation allows the model to end-to-end optimize the graph structure, automatically assigning wider receptive fields to data rich stations while imposing stricter sparsity on isolated nodes to maintain signal integrity.

\subsection{Air Aware Differential Propagation}
Standard Graph Neural Networks (GNNs) primarily perform low pass filtering (smoothing), which models the diffusion process but fails to capture the source generation process \cite{wu2020comprehensive,chen2025stable}. Stacking multiple layers to expand the receptive field often leads to over smoothing, where node representations become indistinguishable. Instead of simple stacking, we adopt an Adaptive Air Graph Diffusion mechanism. This process explicitly simulates the continuous diffusion of pollutant concentrations across the global network, allowing the model to capture dependencies ranging from local accumulation to long range transport.

\paragraph{Multi Step Diffusion Propagation. }

Given input features $\mathbf{H}^{(0)} = \mathbf{X} \in \mathbb{R}^{B \times N \times D}$ and the learned sparse graph $\mathcal{G}$, we model the spatial spreading of information as a recursive diffusion process. We compute the node states at the $l$th diffusion step as, where $\lambda$ is a restart probability that balances global diffusion with local information preservation.:
\begin{equation}
    \mathbf{h}_i^{(l)} = \sum_{j \in \mathcal{N}_i} \tilde{w}_{ij} \cdot \mathbf{h}_j^{(l-1)} + \lambda \cdot \mathbf{h}_i^{(0)}, \quad l = 1, \ldots, L
\end{equation}

The diffusion state collection $\mathbf{H} = [\mathbf{H}^{(0)}, \dots, \mathbf{H}^{(L)}]$ encodes station receptive fields across spatial scales: $\mathbf{H}^{(0)}$ captures immediate local context, and higher-order $\mathbf{H}^{(L)}$ aggregates information from broader regional clusters.

\paragraph{Air Aware Diffusion Aggregation. }

Different monitoring stations require different diffusion profiles. A background station might need to aggregate smooth regional trends, while a station near an emission source needs to highlight local gradients. We propose a spectral aware aggregation mechanism to adaptively fuse these diffusion states.

For each node $i$, we compute query, key, and value projections of its multi scale diffusion states, where $\mathbf{B}$ is a learnable prior matrix encoding step specific biases:
\begin{equation}
    \mathbf{Q}_i = \mathbf{H}_i \mathbf{W}_Q, \quad \mathbf{K}_i = \mathbf{H}_i \mathbf{W}_K, \quad \mathbf{V}_i = \mathbf{H}_i
\end{equation}
where $\mathbf{W}_Q, \mathbf{W}_K \in \mathbb{R}^{D \times D}$ are learnable projection matrices. We employ attention mechanism with signed coefficients to determine the importance of each diffusion step:
\begin{equation}
    \mathbf{A}_i = \tanh\left( \frac{\mathbf{Q}_i \mathbf{K}_i^T}{\sqrt{d_k}} \right) \odot \mathbf{B}
\end{equation}
The $\tanh$ activation differentiates our approach from standard smoothing aggregations: positive coefficients yield standard diffusion, while negative coefficients enable differential diffusion, capturing both smooth regional transport patterns and sharp boundaries from local emission sources.

\paragraph{Multi Scale Fusion and Inference.}
To model dynamics across spatial scales, we aggregate multi-scale diffusion states via a learnable weighted sum: $\mathbf{z}_i = \sum_{l=0}^{L} \text{Softmax}(\mathbf{w})_l \cdot \mathbf{H}^{(l)}$. Crucially, to ensure robustness in data sparse regions, we augment this representation with static identity priors $\mathbf{e}_i^{ID}$ via a gated residual connection:
\begin{equation}
    \hat{\mathbf{z}}_i = \mathbf{g}_i \odot \mathbf{z}_i + (1 - \mathbf{g}_i) \odot \mathbf{e}_i^{ID}
\end{equation}
where the gate $\mathbf{g}_i = \sigma(\mathbf{W}_g [\mathbf{z}_i \| \mathbf{e}_i^{ID}] + \mathbf{b}_g)$ adaptively balances dynamic state estimation with static environmental contexts. The final forecast $\hat{\mathbf{Y}}_i$ is then generated by projecting $\hat{\mathbf{z}}_i$ through an output MLP.

\section{Experiments}

\subsection{Datasets}

\begin{figure}[h]
    \centering
    \includegraphics[width=0.5\textwidth]{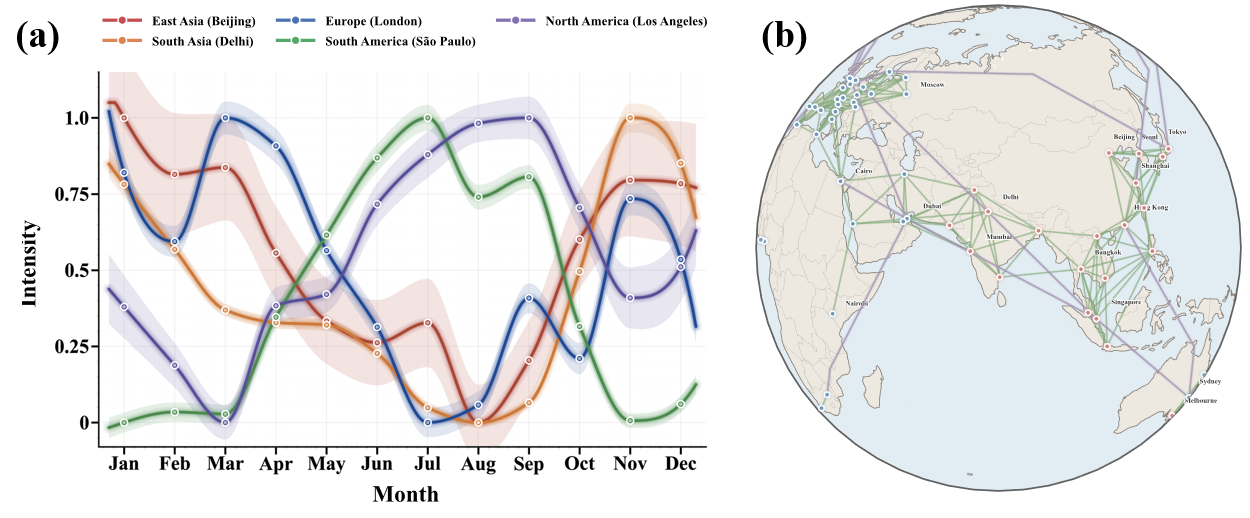}
    \caption{\textbf{Dataset overview.} (a) Monthly PM$_{2.5}$ profiles reveal distinct seasonal patterns across five representative regions. (b) Spatial distribution and learned connectivity of global monitoring stations.}
    \label{fig:dataset}
\end{figure}

We curate a large-scale global real-world dataset of 7,861 air quality monitoring stations across Africa, South America, Asia, Europe and North America. Spanning over a decade, it records daily concentrations of six key air pollutants ($\text{PM}_{2.5}$, $\text{PM}_{10}$, $\text{O}_3$, $\text{NO}_2$, $\text{SO}_2$, $\text{CO}$). We enrich it with spatially aligned meteorological and geographical features (prevailing wind direction, wind speed, elevation, coastal distance) retrieved via each station’s latitude and longitude. For rigorous evaluation, we partition the data into a primary Global set (all 7,861 stations, for cross-region generalization including sparse regions like Africa) and three regional subsets: China (1,692), USA (1,032) and Europe (1,497), with the Europe subset used to assess robustness in complex regional pollution scenarios. As illustrated in Fig.~\ref{fig:dataset}, our dataset exhibits significant temporal heterogeneity across different geographic regions. The open access components of this dataset will be publicly released upon paper acceptance.

\begin{table}[h]
    \centering
    \caption{Comparison of statistical details across air quality datasets. \textbf{Nodes}: Number of monitoring stations. \textbf{Features}: Number of target pollutants and auxiliary covariates.}
    \vspace{-0.5em}
    \label{tab:aq_datasets_refined}
    \resizebox{\linewidth}{!}{%
    \begin{tabular}{ccccc}
        \toprule
        \textbf{Dataset} & \textbf{Nodes ($N$)} & \textbf{Scale} & \textbf{Features} & \textbf{Time Range} \\
        \midrule
        London   & 13    & City        & 1 + 0  & 01/01/2017 $\sim$ 03/31/2018 \\
        KnowAir  & 184   & Nation      & 1 + 8  & 01/01/2015 $\sim$ 12/31/2018 \\
        LargeAQ  & 1,341 & Nation      & 1 + 5  & 01/01/2015 $\sim$ 12/31/2023 \\
        AirPCM-d & 156   & Nation      & 6 + 5  & 01/02/2015 $\sim$ 04/12/2025 \\
        AirPCM-h & 453   & CN, US, EU  & 6 + 5  & 01/17/2024 $\sim$ 04/17/2025 \\
        \midrule
        \rowcolor{white} 
        \textbf{\color{softpink}WorldAir} & \textbf{\color{softpink}7,861} & \textbf{\color{softpink}Global} & \textbf{\color{softpink}6 + $>$10$^\dagger$} & \textbf{\color{softpink}10/01/2012 $\sim$ 02/28/2025} \\
        \bottomrule
        \multicolumn{5}{l}{\small $^\dagger$ Due to the challenges of data acquisition, the covariates primarily consist of static attributes.}
    \end{tabular}
    }
\end{table}

\begin{table*}[t!]
    \centering
    \renewcommand{\arraystretch}{0.8}
    \caption{Performance comparison and Ablation study. \best{Best} and \secondbest{second best} results are highlighted. The ` -- ' marker indicates baselines incur out-of-memory issues even on minimum batch size.}
    \label{tab:overall_performance}
    
    \resizebox{\textwidth}{!}{%
        \begin{tabular}{lcccccccccccc}
            \toprule
            \multirow{2}{*}{\textbf{Model}} & \multicolumn{3}{c}{\textbf{China}} & \multicolumn{3}{c}{\textbf{Europe}} & \multicolumn{3}{c}{\textbf{USA}} & \multicolumn{3}{c}{\textbf{Global}} \\
            \cmidrule(lr){2-4} \cmidrule(lr){5-7} \cmidrule(lr){8-10} \cmidrule(lr){11-13}
             & MAE & RMSE & MAPE & MAE & RMSE & MAPE & MAE & RMSE & MAPE & MAE & RMSE & MAPE \\
            \midrule
            
            LV & 9.68 & 19.43 & 30.12\% & 14.32 & 23.98 & 51.64\% & 13.44 & 22.70 & 55.31\% & 14.18 & 25.22 & 42.98\% \\
            GM & 17.33 & 23.99 & 60.82\% & 23.50 & 57.53 & 112.97\% & 13.90 & 23.58 & 66.11\% & 23.06 & 45.16 & 83.67\% \\
            VAR & 11.14 & 17.78 & 38.11\% & 14.24 & 22.57 & 49.49\% & 12.20 & 20.38 & 53.58\% & 14.29 & 21.20 & 49.88\% \\
            \midrule 
            LightTS & 9.13 & 16.85 & 26.32\% & 13.30 & 22.63 & 46.69\% & 11.95 & 19.67 & 51.78\% & 12.57 & 27.58 & 40.65\% \\
            Informer & 8.58 & 16.02 & 26.42\% & 14.23 & 22.36 & 46.82\% & 11.69 & 19.26 & 43.85\% & 12.33 & 22.06 & \secondbest{35.50\%} \\
            Autoformer & 8.97 & 15.97 & 28.02\% & 13.70 & 22.37 & 50.05\% & 11.85 & 19.73 & 44.00\% & 13.39 & 22.61 & 37.60\% \\
            \midrule 
            STGCN & 9.67 & 16.14 & 30.19\% & 14.34 & 25.47 & 48.25\% & 10.75 & 19.01 & 43.71\% & 14.59 & 28.47 & 47.27\% \\
            STID & 8.10 & 15.14 & 25.86\% & \secondbest{11.90} & \secondbest{20.27} & 43.60\% & 10.45 & \secondbest{18.39} & 41.77\% & \secondbest{11.93} & \secondbest{21.56} & 35.82\% \\
            StemGNN & 9.55 & 16.24 & 30.04\% & 13.23 & 22.24 & 53.45\% & 10.91 & 18.96 & 44.72\% & 13.60 & 27.15 & 41.94\% \\
            AGCRN & 9.60 & 15.87 & 29.56\% & 12.45 & 21.44 & 47.69\% & 10.54 & 18.76 & \secondbest{41.28\%} & -- & -- & -- \\
            GWNet & \secondbest{7.91} & \secondbest{14.85} & 26.10\% & 12.21 & 21.33 & 45.53\% & 10.48 & \best{18.38} & 41.35\% & 12.89 & 26.48 & 39.67\% \\
            MTGNN & 8.83 & 15.68 & 28.22\% & 13.00 & 25.08 & 56.34\% & 10.56 & 18.50 & 42.67\% & 13.12 & 26.88 & 37.63\% \\
            \midrule 
            PM2.5GNN & 8.00 & 15.07 & \secondbest{25.29\%} & 12.54 & 22.05 & 47.76\% & \secondbest{10.43} & 18.68 & 42.35\% & 12.90 & 27.23 & 38.54\% \\
            GAGNN & 8.65 & 15.39 & 27.14\% & 12.61 & 21.83 & 46.65\% & 10.47 & 18.58 & 42.27\% & 12.81 & 27.17 & 38.44\% \\
            Airformer & 8.75 & 15.74 & 25.52\% & 11.93 & 21.31 & 43.53\% & 10.79 & 18.63 & 41.88\% & 12.52 & 23.36 & 36.62\% \\
            AirPhyNet & 8.13 & 15.10 & 25.59\% & 12.26 & 21.55 & 46.00\% & \secondbest{10.43} & 18.46 & 41.72\% & 12.97 & 27.31 & 39.05\% \\
            AirDualODE & 8.25 & 15.09 & 26.78\% & 12.30 & 21.65 & 45.55\% & 10.49 & 18.56 & 42.03\% & 12.78 & 27.32 & 37.11\% \\
            CauAir & 8.02 & 15.39 & 25.56\% & 12.17 & 20.51 & \secondbest{42.34\%} & 11.67 & 19.55 & 42.34\% & 13.22 & 24.38 & 36.86\% \\
            
            \midrule
            \textit{w/o Ext. Feat.} & 7.81 & 15.02 & 24.35\% & 11.89 & 20.05 & 41.28\% & 10.52 & 18.56 & 41.52\% & 11.56 & 20.89 & 34.42\% \\
            \textit{w/o Sem. Graph} & 7.88 & 15.18 & 24.68\% & 12.05 & 20.38 & 41.95\% & 10.68 & 18.82 & 42.15\% & 11.82 & 21.35 & 35.18\% \\
            \textit{w/o Adap. Sparse} & 7.96 & 15.35 & 24.89\% & 12.18 & 20.62 & 42.35\% & 10.79 & 18.95 & 42.56\% & 11.98 & 21.67 & 35.72\% \\
            \textit{w/o ISIE} & 8.05 & 15.52 & 25.18\% & 12.35 & 20.92 & 42.88\% & 10.95 & 19.18 & 43.12\% & 12.18 & 22.05 & 36.25\% \\
            \textit{w/o Dyn. Graph} & 8.28 & 15.95 & 25.85\% & 12.72 & 21.58 & 43.82\% & 11.22 & 19.75 & 44.05\% & 12.58 & 22.78 & 37.52\% \\
            
            \midrule
            \textbf{OmniAir} & \best{7.72} & \best{14.83} & \best{24.01\%} & \best{11.70} & \best{19.73} & \best{40.71\%} & \best{10.38} & \best{18.38} & \best{41.04\%} & \best{11.28} & \best{20.41} & \best{33.69\%} \\
            \bottomrule
        \end{tabular}%
    }
\end{table*}

\begin{table*}[h!]
    \centering
    \renewcommand{\arraystretch}{0.75} 
    \caption{Performance comparison on Global across different pollutants. \best{Best} and 
    \secondbest{second best} results are highlighted.}
    \vspace{-0.5em}
    \label{tab:pollutant_performance}
    
    \resizebox{\textwidth}{!}{%
        \begin{tabular}{lcccccccccccc}

            \toprule
            \multirow{2}{*}{\textbf{Model}} & \multicolumn{3}{c}{\textbf{CO}} & \multicolumn{3}{c}{\textbf{NO$_2$}} & \multicolumn{3}{c}{\textbf{O$_3$}} & \multicolumn{3}{c}{\textbf{PM$_{10}$}} \\
            \cmidrule(lr){2-4} \cmidrule(lr){5-7} \cmidrule(lr){8-10} \cmidrule(lr){11-13}
             & MAE & RMSE & MAPE & MAE & RMSE & MAPE & MAE & RMSE & MAPE & MAE & RMSE & MAPE \\
            \midrule
            
            AGCRN & 1.26 & 2.47 & 34.50\% & 3.29 & 4.69 & 48.01\% & 7.12 & 10.70 & 28.65\% & 9.71 & 20.95 & \secondbest{41.22\%} \\
            AirFormer & 1.58 & 2.84 & 43.30\% & 3.64 & 5.21 & 49.95\% & 8.02 & 11.75 & 31.93\% & 12.66 & 24.13 & 55.22\% \\
            AirPhyNet & 1.22 & 2.44 & 32.74\% & 3.26 & 4.71 & 45.91\% & 7.19 & 10.77 & 28.44\% & 9.62 & \secondbest{20.61} & 42.57\% \\
            CauAir & 1.24 & 2.44 & 33.82\% & 3.24 & 4.69 & \secondbest{44.80\%} & 7.21 & 10.77 & 28.61\% & 9.89 & 20.84 & 43.81\% \\
            AirDualODE & \secondbest{1.20} & \secondbest{2.40} & \secondbest{32.25\%} & \secondbest{3.22} & \secondbest{4.67} & 45.06\% & \best{7.06} & \best{10.58} & \secondbest{27.80\%} & \secondbest{9.60} & 20.68 & 41.81\% \\
            \midrule
            \textbf{OmniAir} & \best{1.04} & \best{2.20} & \best{29.47\%} & \best{3.05} & \best{4.53} & \best{42.26\%} & \secondbest{7.11} & \secondbest{10.64} & \best{27.76\%} & \best{9.17} & \best{20.58} & \best{38.56\%} \\
            
            \bottomrule
        \end{tabular}%
        \vspace{-1em}
    }
\end{table*}

\subsection{Baselines}

For performance evaluation, we compare OmniAir with statistical baselines LV, GM and VAR: LV predicts from the last observation \cite{burtscher1999exploring}, GM models system uncertainty via differential equations \cite{ando2004geometric}, VAR captures linear interdependencies in multivariate time series \cite{stock2001vector}. We also include three categories of SOTA deep learning models: (1) General Time Series: LightTS~\cite{campos2023lightts}, Informer~\cite{zhou2021informer}, Autoformer~\cite{wu2021autoformer}; (2) Spatio-Temporal: STGCN~\cite{yu2017stgcn}, STID~\cite{shao2022stid}, StemGNN~\cite{cao2020stemgnn}, AGCRN~\cite{bai2020agcrn}, GWNet~\cite{wu2019gwnet}, MTGNN~\cite{wu2020mtgnn}; (3) Air Quality: PM2.5-GNN~\cite{wang2020pm2}, GAGNN~\cite{chen2023gagnn}, Airformer~\cite{liang2023airformer}, AirPhyNet~\cite{hettige2024airphynet}, AirDualODE~\cite{tian2024air}, CauAir~\cite{ma2025causal}.

\subsection{Experimental Setup}
All experiments were conducted on an NVIDIA A800 (80 GB memory), with code implemented in Python 3.11/PyTorch based on BasicTS \cite{liang2022basicts}. The dataset was split chronologically into training/validation/test sets at a 6:2:2 ratio. All models were trained with consistent hyperparameters: input/output sequence lengths of 30/14, batch size 32, learning rate 0.001, weight decay $10^{-5}$, max epochs 300, and early stopping patience 20. Fourier coordinate encoding and identity embedding dimensions were set to 32 and 64, respectively; the topology generator used 10 geographic and 5 semantic neighbors uniformly across datasets for fair comparison. We optimized with Adam and evaluated via MAE, RMSE, and MAPE; all experiments were repeated five times with different random seeds (average results reported). More details refer to Appendix \ref{sec:implementation_details}.

\begin{table*}[t!]
\centering
\caption{Efficiency comparison with baselines. \textbf{Mem}: The maximum memory usage (GB) during training. \textbf{Train}: Average training time (s/epoch). \textbf{Infer}: Average inference time. \textbf{BS}: The maximum allowable batch size (up to 64).}
\label{tab:efficiency_final_corrected}

\resizebox{\textwidth}{!}{%
\begin{tabular}{l ccc ccc ccc ccc cc}
\toprule
\multirow{2}{*}{\textbf{Model}} & \multicolumn{3}{c}{\textbf{China}} & \multicolumn{3}{c}{\textbf{Europe} } & \multicolumn{3}{c}{\textbf{USA}} & \multicolumn{3}{c}{\textbf{Global} } & \multicolumn{2}{c}{} \\
\cmidrule(lr){2-4} \cmidrule(lr){5-7} \cmidrule(lr){8-10} \cmidrule(lr){11-13}
 & Mem & Train & Infer & Mem & Train & Infer & Mem & Train & Infer & Mem & Train & Infer & BS & MAE \\
\midrule

PM2.5GNN & 
55.84\inc{498\%} & 45.0\inc{291\%} & 11.4\inc{443\%} & 
49.58\inc{500\%} & 45.9\inc{388\%} & 11.4\inc{533\%} & 
33.85\inc{503\%} & 32.1\inc{379\%} & 7.7\inc{542\%} & 
47.24\inc{45\%} & 195.8\inc{538\%} & 47.3\inc{586\%} & 
18\inc{72\%} & 12.90\inc{14\%} \\

AirFormer & 
50.51\inc{441\%} & 98.0\inc{752\%} & 18.3\inc{771\%} & 
44.69\inc{440\%} & 86.4\inc{819\%} & 16.3\inc{806\%} & 
30.83\inc{450\%} & 60.8\inc{807\%} & 10.9\inc{808\%} & 
43.22\inc{33\%} & 347.5\inc{1032\%} & 62.0\inc{799\%} & 
16\inc{75\%} & 12.52\inc{11\%} \\

StemGNN & 
21.42\inc{129\%} & 40.3\inc{250\%} & 7.2\inc{243\%} & 
17.89\inc{116\%} & 39.5\inc{320\%} & 7.5\inc{317\%} & 
10.67\inc{90\%} & 20.3\inc{203\%} & 3.7\inc{208\%} & 
43.22\inc{33\%} & 1343.4\inc{4276\%} & 285.7\inc{4041\%} & 
8\inc{88\%} & 13.60\inc{21\%} \\

STGCN & 
22.76\inc{144\%} & 57.3\inc{398\%} & 11.4\inc{443\%} & 
20.13\inc{143\%} & 43.9\inc{367\%} & 8.6\inc{378\%} & 
13.87\inc{147\%} & 24.3\inc{263\%} & 4.6\inc{283\%} & 
38.35\inc{18\%} & 363.0\inc{1082\%} & 68.2\inc{888\%} & 
16\inc{75\%} & 14.59\inc{29\%} \\

GWNet & 
42.39\inc{354\%} & 123.5\inc{974\%} & 22.5\inc{971\%} & 
37.50\inc{353\%} & 102.9\inc{995\%} & 18.1\inc{906\%} & 
25.85\inc{361\%} & 57.8\inc{763\%} & 10.0\inc{733\%} & 
36.43\inc{12\%} & 521.3\inc{1598\%} & 98.4\inc{1326\%} & 
5\inc{92\%}  & 12.89\inc{14\%} \\

AGCRN & 
43.65\inc{367\%} & 162.0\inc{1309\%} & 34.0\inc{1519\%} & 
38.15\inc{361\%} & 150.3\inc{1499\%} & 38.7\inc{2050\%} & 
25.42\inc{353\%} & 93.1\inc{1290\%} & 26.2\inc{2083\%} & 
-- & -- & -- & 
OOM & -- \\

\midrule

\textbf{OmniAir} & 
\textbf{\color{softpink}9.34} &  \textbf{\color{softpink}11.5} & \textbf{\color{softpink}2.1} & 
\textbf{\color{softpink}8.27} & \textbf{\color{softpink}9.4} & \textbf{\color{softpink}1.8} & 
\textbf{\color{softpink}5.61} & \textbf{\color{softpink}6.7} & \textbf{\color{softpink}1.2} & 
\textbf{\color{softpink}32.50} & \textbf{\color{softpink}30.7} & \textbf{\color{softpink}6.9} & 
\textbf{\color{softpink}64} & \textbf{\color{softpink}11.28} \\

\bottomrule
\end{tabular}%
\vspace{-1em}
}
\end{table*}

\subsection{Overall Performance Comparison}
Table \ref{tab:overall_performance} compares OmniAir with SOTA baselines on three regional subsets and the Global dataset, where OmniAir consistently outperforms all baselines across all metrics regionally and globally. Traditional statistical methods underperform for their inability to model complex non-linear spatio-temporal dependencies; deep learning baselines perform competitively in data-rich regions (e.g., China, Europe) but fail to generalize globally, with several advanced models suffering scalability bottlenecks and out-of-memory errors on large global graphs. By contrast, OmniAir achieves a 5.45\% global MAE reduction, with its advantage expanding with network size. While GNNs exhibit ~47\% average error growth when scaling from China to the Global dataset, OmniAir restricts this degradation to 31\%. Pollutant-specific evaluation (Table \ref{tab:pollutant_performance}) confirms OmniAir’s robust adaptability to diverse physicochemical properties, yielding the lowest error rates for primary pollutants and $\text{PM}_{10}$ (outperforming AirDualODE by 4.48\%, demonstrating superior source-generation duality capture). Even for secondary pollutants such as $\text{O}_3$ (complex photochemical reactions), OmniAir retains superior accuracy, validating its effective disentanglement of diffusion and source-generation mechanisms for distinct atmospheric components.

\subsection{Ablation Study}
To rigorously validate the contribution of each OmniAir component, we conduct a comprehensive ablation study on the Global dataset, constructing five variants by systematically ablating key modules: 1) \textit{w/o Ext. Feat.}: excludes auxiliary environmental attributes to test multi-source context value; 2) \textit{w/o Sem. Graph}: removes semantic connections, confining topology to geographic neighbors only; 3) \textit{w/o Adap. Sparse}: replaces learnable pruning with a fixed neighborhood size; 4) \textit{w/o ISIE}: replaces the inductive encoder with standard transductive node embeddings; 5) \textit{w/o Dyn. Graph}: disables the time-varying edge weight mechanism. Results are summarized in the bottom of Table \ref{tab:overall_performance}, confirming the necessity of all components, \textit{w/o Dyn. Graph} shows the most significant performance degradation, with a MAE increase of over 10\% on the Global dataset. Detailed variant definitions are provided in Appendix \ref{ablation}.

\subsection{Efficiency Analysis}

Beyond predictive accuracy, we evaluate the computational efficiency of our framework in Table \ref{tab:efficiency_final_corrected}. Complex models face high time/space complexity, limiting their deployment on large global networks; by contrast, OmniAir drastically reduces memory consumption and training time (over 90\% faster than several SOTA baselines). Specifically, it achieves an order-of-magnitude speedup: training latency is reduced by over 10$\times$ vs. AirFormer and 43$\times$ vs. StemGNN globally. Its lightweight architecture enables larger batch sizes and faster inference than heavy Transformer variants. This efficiency confirms OmniAir is both accurate and highly scalable, making it practical for real-time global air quality monitoring under constrained computational resources. Theoretical proof of linear complexity is in Appendix \ref{sec:complexity}.

\begin{figure}[t!]
    \centering
    \includegraphics[width=\linewidth]{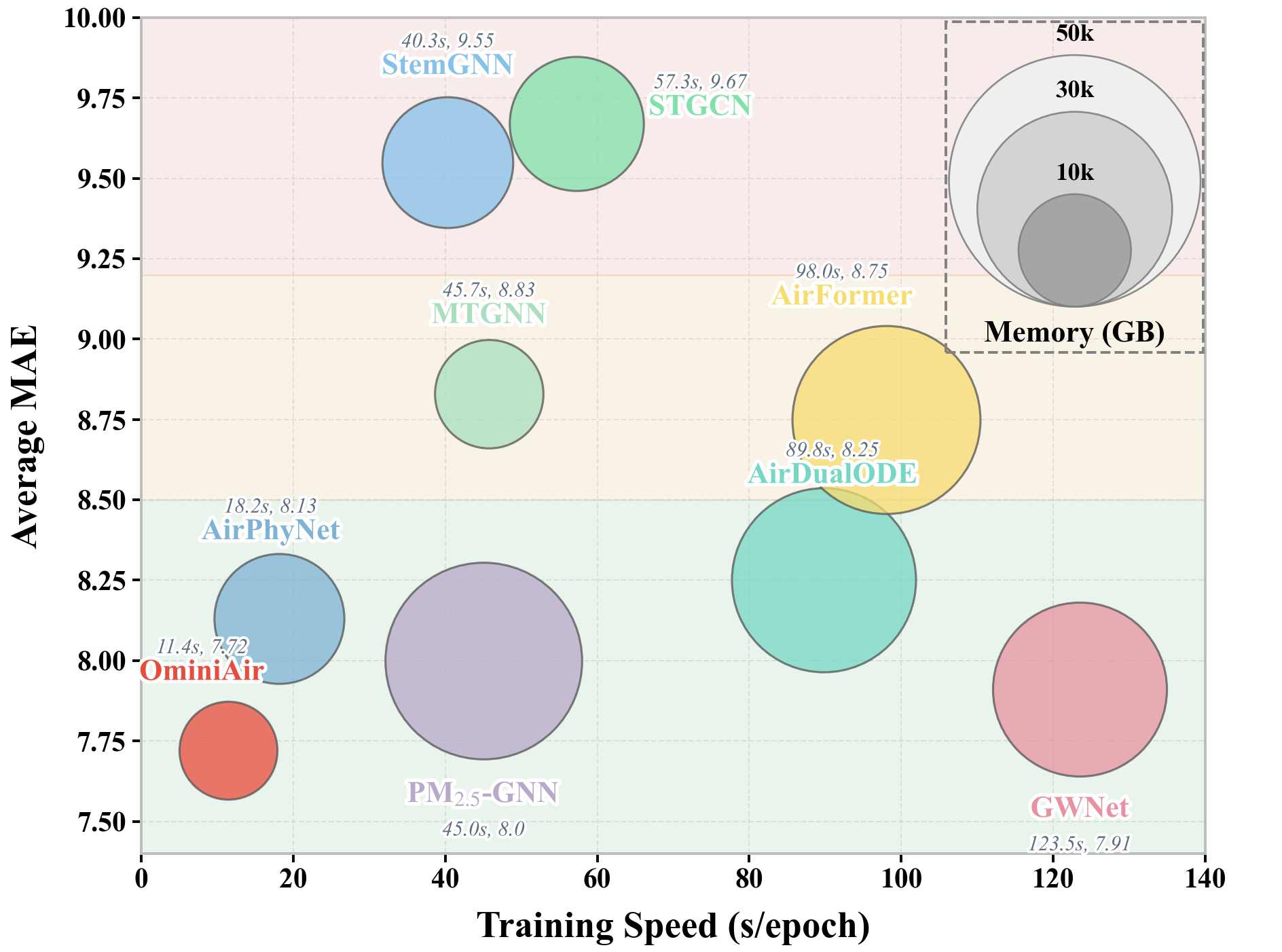}
    \caption{Efficiency trade-off comparison regarding MAE, training speed, and memory usage.}
    \label{fig:efficiency}
    \vspace{-0.5em}
\end{figure}

\begin{table}[h!]
    \centering
    \caption{Average performance comparison of KnowAir and LargeAQ. \best{Best} and \secondbest{second best} results are highlighted.}
    \vspace{-0.5em}
    \label{tab:overall_performance_knowair}
    \resizebox{\linewidth}{!}{%
        \renewcommand{\arraystretch}{0.90} 
        \begin{tabular}{lcccccc}
            \toprule
            \multirow{2}{*}{\textbf{Model}} & \multicolumn{3}{c}{\textbf{KnowAir}} & \multicolumn{3}{c}{\textbf{LargeAQ}} \\
            \cmidrule(lr){2-4} \cmidrule(lr){5-7}
             & MAE & RMSE & MAPE(\%) & MAE & RMSE & MAPE(\%) \\
            \midrule
            STGCN      & 9.18  & 14.02 & 53.86 & 17.13 & 26.15 & 62.96 \\
            STID       & 9.33  & 14.06 & 58.98 & 17.65 & 27.05 & 67.70 \\
            AGCRN      & 9.56  & 14.45 & 57.87 & 17.55 & 26.87 & 65.48 \\
            GWNet      & 10.36 & 15.08 & 63.45 & 17.76 & 26.77 & 69.06 \\
            PM2.5GNN   & \secondbest{9.06} & 14.05 & 54.25 & 16.86 & 25.75 & 61.63 \\
            GAGNN      & 11.25 & 16.81 & 67.49 & 22.16 & 39.22 & 74.86 \\
            AirFormer  & 9.20  & \secondbest{13.76} & \secondbest{52.96} & \secondbest{16.27} & \secondbest{24.07} & \secondbest{59.36} \\
            AirPhyNet  & 10.52 & 15.74 & 66.27 & 18.92 & 28.72 & 69.62 \\
            \midrule
            \textbf{Ours} & \best{8.43} & \best{12.77} & \best{48.31} & \best{14.14} & \best{22.09} & \best{46.19} \\
            \bottomrule
        \end{tabular}%
        \vspace{-0.5em}
    }
\end{table}

\subsection{Generalization on Small scale Datasets}
To further verify the robustness and generalization capability of our approach beyond our collected dataset, we conducted extensive experiments on two widely used public benchmark datasets, KnowAir \cite{wang2020pm2} and LargeAQ \cite{ma2025causal}. The results in Table \ref{tab:overall_performance_knowair} indicate that OmniAir consistently outperforms baseline models across varying prediction horizons, ranging from short-term to long-term forecasting. Unlike baseline models that often exhibit performance degradation as the forecasting horizon extends, our method maintains stable error rates. This success can be attributed to our model's ability to effectively disentangle dynamic spatial correlations from temporal dependencies, confirming its adaptability to different geographical environments and data distributions. Furthermore, the consistent superiority across all metrics demonstrates the universal efficiency of OmniAir in both multi horizon forecasting and the effective integration of diverse covariates, proving its reliability as a general purpose framework.

\subsection{Hyperparameter Sensitivity Analysis}
\label{sec:hyperparameter}

We conduct a sensitivity analysis of two key hyperparameters for WorldAir in Fig.~\ref{fig:sensitivity}. For Fourier encoding dimension, performance peaks at 32: lower values fail to capture high-frequency spatial variations, while excessive dimensions introduce artifacts. For semantic neighbors, the optimal value is 5. Notably, models with semantic connections significantly outperform the baseline without a semantic graph, validating the necessity of modeling non-Euclidean correlations for global air quality forecasting.

\begin{figure}[h!]
    \centering
    \includegraphics[width=\linewidth]{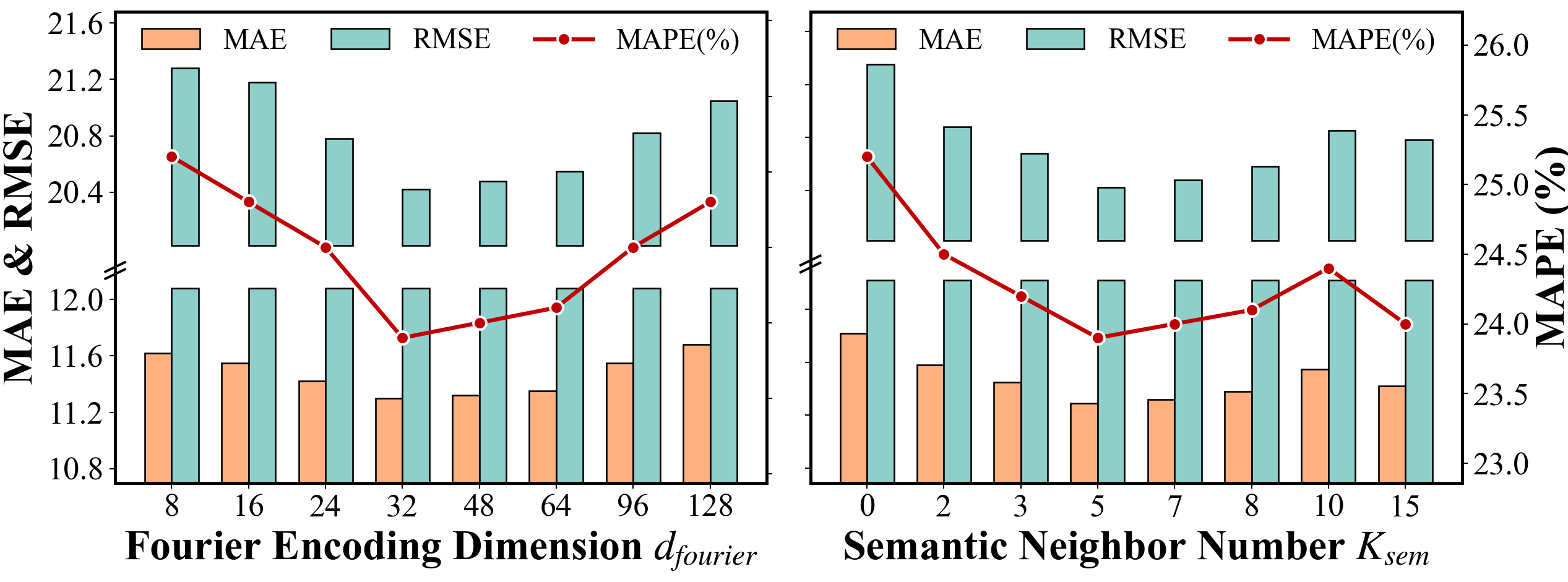}
    \caption{Sensitivity analysis of OmniAir}
    \label{fig:sensitivity}
    \vspace{-1em}
\end{figure}

\subsection{Embedding Space Visualization}
\begin{figure}[h!]
    \centering
    \includegraphics[width=0.5\textwidth]{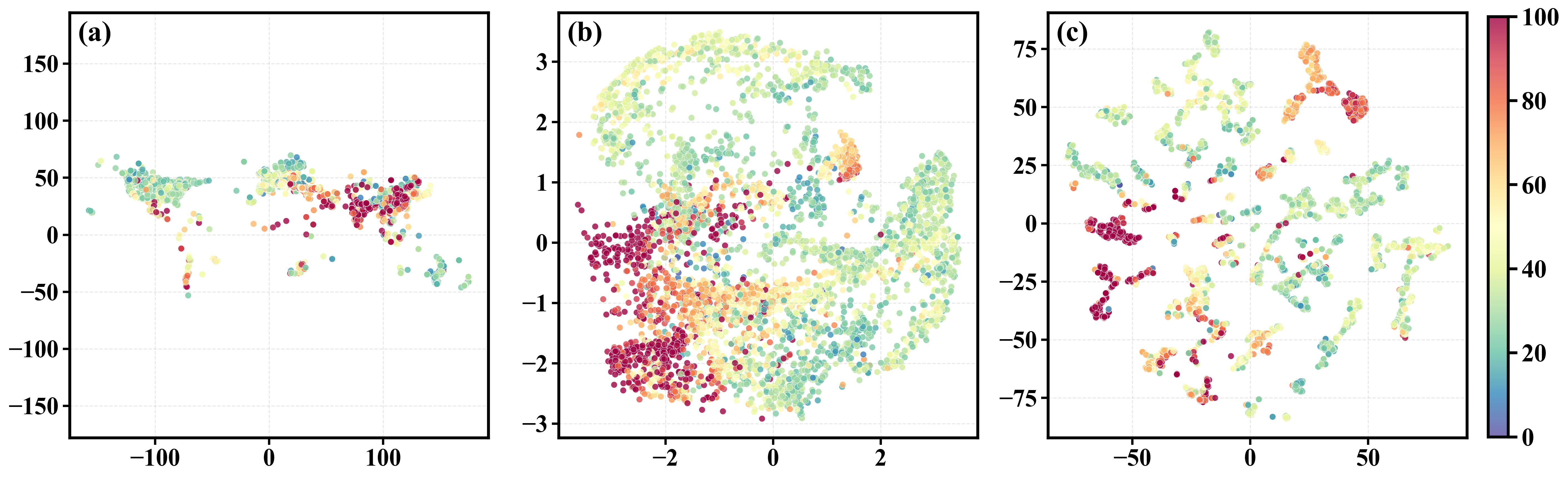}
    \caption{Geographic coordinates (a) vs. PCA (b) and t-SNE (c) projections of learned embeddings (colored by mean PM$_{2.5}$ severity). Learned high-dimensional embeddings cluster stations with similar PM$_{2.5}$ levels in 2D projections.}
    \label{fig:embedding_space}
    \vspace{-1em}
\end{figure}

Figure~\ref{fig:embedding_space} compares the physical distribution of stations with their learned semantic representations. While the geographic coordinates in Panel (a) show clusters based on continental location, the high dimensional embeddings are reduced to 2D via PCA (Panel b) and t-SNE (Panel c) demonstrate a semantic reorganization driven by pollution severity (indicated by color). The mapping of geographically distant but environmentally similar stations into proximate latent clusters validates the effectiveness of OminiAir in capturing global non-Euclidean dependencies.

\subsection{Training Dynamics and Regional Adaptation}
\label{sub:training_dynamics}
\begin{figure}[H]
    \centering
    \includegraphics[width=0.8\linewidth]{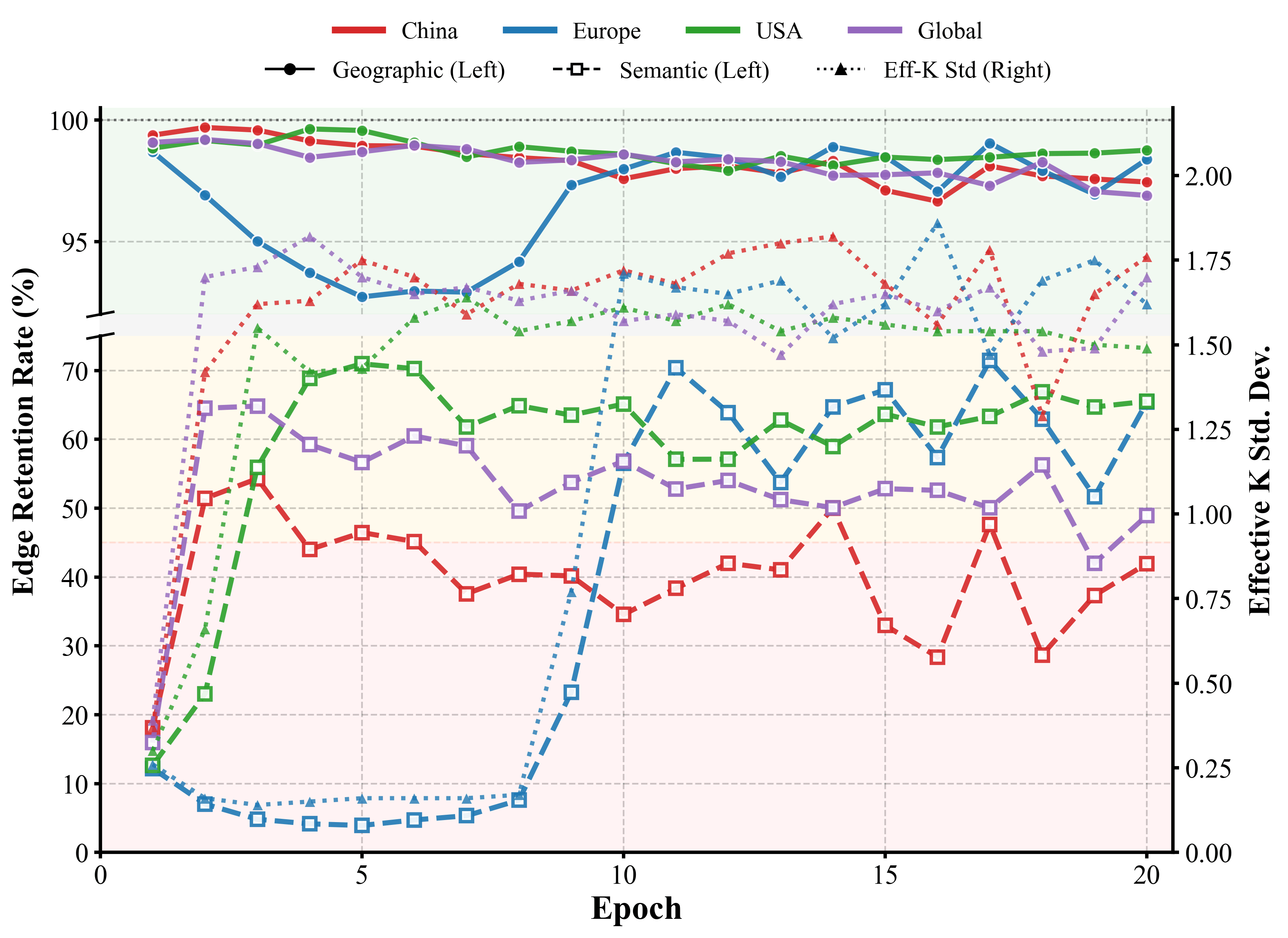}
    \caption{Training dynamics of the adaptive sparse mechanism. Solid/dashed lines denote geographic/semantic edge retention (left axis); the dotted line tracks the standard deviation of effective neighborhood size $K$ (right axis).}
    \label{fig:adaptive_sparse}
\end{figure}

To verify adaptive behavior, we visualize edge retention rate evolution in Fig.~\ref{fig:adaptive_sparse}, which reveals distinct topology evolution trajectories: geographic edges maintain a stable retention rate above 95\%, while semantic edges undergo aggressive pruning, dropping from an initial 60\%–70\% to a refined 5\%–40\% subset. This differential pruning shows notable regional heterogeneity; semantic retention plummets to $\approx$5\% in station-dense Europe, yet remains at $\approx$30\% for the global set to enable cross-continental knowledge transfer in data-sparse regions. A concurrent rise in the standard deviation of neighborhood size $K$ further confirms the model’s transition from a uniform $k$-neighbor structure to a variance-adaptive one, which dynamically optimizes receptive fields according to local node density.

\section{Conclusion}
We present OmniAir, the first inductive framework for global air quality prediction. Leveraging our curated WorldAir dataset (7,800+ stations, the largest to date), OmniAir overcomes regional barriers of traditional transductive models and bridges monitoring gaps in data-sparse regions via cross-continental knowledge transfer. 

\section*{Impact Statement}
This research advances global environmental intelligence by establishing a scalable, inductive framework for high precision air quality forecasting. The Global subset  evaluates performance on under monitored regions in the Global South, where OmniAir's inductive capability provides critical coverage. Our work contributes to the democratization of health critical data and supports evidence based policy making in data sparse areas. However, we emphasize that model predictions serve as a complement to, rather than a substitute for, ground truth infrastructure. Stakeholders must remain aware of inherent forecasting uncertainties and avoid over reliance on algorithmic outputs for critical health or emergency decisions without corroboration from official local authorities.
\bibliography{OminiAir/main}

@article{campos2023lightts,
  title={Lightts: Lightweight time series classification with adaptive ensemble distillation},
  author={Campos, David and Zhang, Miao and Yang, Bin and Kieu, Tung and Guo, Chenjuan and Jensen, Christian S},
  journal={Proceedings of the ACM on Management of Data},
  volume={1},
  number={2},
  pages={1--27},
  year={2023},
  publisher={ACM New York, NY, USA}
}

@inproceedings{zhou2021informer,
  title={Informer: Beyond efficient transformer for long sequence time-series forecasting},
  author={Zhou, Haoyi and Zhang, Shanghang and Peng, Jieqi and Zhang, Shuai and Li, Jianxin and Xiong, Hui and Zhang, Wancai},
  booktitle={Proceedings of the AAAI conference on artificial intelligence},
  volume={35},
  number={12},
  pages={11106--11115},
  year={2021}
}

@article{wu2021autoformer,
  title={Autoformer: Decomposition transformers with auto-correlation for long-term series forecasting},
  author={Wu, Haixu and Xu, Jiehui and Wang, Jianmin and Long, Mingsheng},
  journal={Advances in neural information processing systems},
  volume={34},
  pages={22419--22430},
  year={2021}
}

@article{yu2017stgcn,
  title={Spatio-temporal graph convolutional networks: A deep learning framework for traffic forecasting},
  author={Yu, Bing and Yin, Haoteng and Zhu, Zhanxing},
  journal={arXiv preprint arXiv:1709.04875},
  year={2017}
}

@inproceedings{shao2022stid,
  title={Spatial-temporal identity: A simple yet effective baseline for multivariate time series forecasting},
  author={Shao, Zezhi and Zhang, Zhao and Wang, Fei and Wei, Wei and Xu, Yongjun},
  booktitle={Proceedings of the 31st ACM international conference on information \& knowledge management},
  pages={4454--4458},
  year={2022}
}

@article{cao2020stemgnn,
  title={Spectral temporal graph neural network for multivariate time-series forecasting},
  author={Cao, Defu and Wang, Yujing and Duan, Juanyong and Zhang, Ce and Zhu, Xia and Huang, Congrui and Tong, Yunhai and Xu, Bixiong and Bai, Jing and Tong, Jie and others},
  journal={Advances in neural information processing systems},
  volume={33},
  pages={17766--17778},
  year={2020}
}

@article{bai2020agcrn,
  title={Adaptive graph convolutional recurrent network for traffic forecasting},
  author={Bai, Lei and Yao, Lina and Li, Can and Wang, Xianzhi and Wang, Can},
  journal={Advances in neural information processing systems},
  volume={33},
  pages={17804--17815},
  year={2020}
}

@article{wu2019gwnet,
  title={Graph wavenet for deep spatial-temporal graph modeling},
  author={Wu, Zonghan and Pan, Shirui and Long, Guodong and Jiang, Jing and Zhang, Chengqi},
  journal={arXiv preprint arXiv:1906.00121},
  year={2019}
}

@inproceedings{wu2020mtgnn,
  title={Connecting the dots: Multivariate time series forecasting with graph neural networks},
  author={Wu, Zonghan and Pan, Shirui and Long, Guodong and Jiang, Jing and Chang, Xiaojun and Zhang, Chengqi},
  booktitle={Proceedings of the 26th ACM SIGKDD international conference on knowledge discovery \& data mining},
  pages={753--763},
  year={2020}
}

@inproceedings{wang2020pm2,
  title={Pm2. 5-gnn: A domain knowledge enhanced graph neural network for pm2. 5 forecasting},
  author={Wang, Shuo and Li, Yanran and Zhang, Jiang and Meng, Qingye and Meng, Lingwei and Gao, Fei},
  booktitle={Proceedings of the 28th international conference on advances in geographic information systems},
  pages={163--166},
  year={2020}
}

@article{chen2023gagnn,
  title={Group-aware graph neural network for nationwide city air quality forecasting},
  author={Chen, Ling and Xu, Jiahui and Wu, Binqing and Huang, Jianlong},
  journal={ACM Transactions on Knowledge Discovery from Data},
  volume={18},
  number={3},
  pages={1--20},
  year={2023},
  publisher={ACM New York, NY}
}

@inproceedings{liang2023airformer,
  title={Airformer: Predicting nationwide air quality in china with transformers},
  author={Liang, Yuxuan and Xia, Yutong and Ke, Songyu and Wang, Yiwei and Wen, Qingsong and Zhang, Junbo and Zheng, Yu and Zimmermann, Roger},
  booktitle={Proceedings of the AAAI conference on artificial intelligence},
  volume={37},
  number={12},
  pages={14329--14337},
  year={2023}
}

@article{hettige2024airphynet,
  title={Airphynet: Harnessing physics-guided neural networks for air quality prediction},
  author={Hettige, Kethmi Hirushini and Ji, Jiahao and Xiang, Shili and Long, Cheng and Cong, Gao and Wang, Jingyuan},
  journal={arXiv preprint arXiv:2402.03784},
  year={2024}
}

@article{tian2024air,
  title={Air quality prediction with physics-guided dual neural odes in open systems},
  author={Tian, Jindong and Liang, Yuxuan and Xu, Ronghui and Chen, Peng and Guo, Chenjuan and Zhou, Aoying and Pan, Lujia and Rao, Zhongwen and Yang, Bin},
  journal={arXiv preprint arXiv:2410.19892},
  year={2024}
}

@inproceedings{ma2025causal,
  title={Causal learning meet covariates: Empowering lightweight and effective nationwide air quality forecasting},
  author={Ma, Jiaming and Cui, Zhiqing and Wang, Binwu and Wang, Pengkun and Zhou, Zhengyang and Zhao, Zhe and Wang, Yang},
  booktitle={International Joint Conference on Artificial Intelligence},
  year={2025}
}

@inproceedings{liang2022basicts,
  title={Basicts: An open source fair multivariate time series prediction benchmark},
  author={Liang, Yubo and Shao, Zezhi and Wang, Fei and Zhang, Zhao and Sun, Tao and Xu, Yongjun},
  booktitle={International symposium on benchmarking, measuring and optimization},
  pages={87--101},
  year={2022},
  organization={Springer}
}

@misc{who2022billions,
  title = {Billions of people still breathe unhealthy air: new {WHO} data},
  author = {{World Health Organization}},
  year = {2022},
  month = {April},
  howpublished = {\url{https://www.who.int/news/item/04-04-2022-billions-of-people-still-breathe-unhealthy-air-new-who-data}},
}

@article{horn2024air,
  title={The Air Quality Index (AQI) in historical and analytical perspective a tutorial review},
  author={Horn, Seth A and Dasgupta, Purnendu K},
  journal={Talanta},
  volume={267},
  pages={125260},
  year={2024},
  publisher={Elsevier}
}

@article{kumar2023critical,
  title={Critical review on emerging health effects associated with the indoor air quality and its sustainable management},
  author={Kumar, Pradeep and Singh, AB and Arora, Taruna and Singh, Sevaram and Singh, Rajeev},
  journal={Science of The Total Environment},
  volume={872},
  pages={162163},
  year={2023},
  publisher={Elsevier}
}

@article{asudani2023impact,
  title={Impact of word embedding models on text analytics in deep learning environment: a review},
  author={Asudani, Deepak Suresh and Nagwani, Naresh Kumar and Singh, Pradeep},
  journal={Artificial intelligence review},
  volume={56},
  number={9},
  pages={10345--10425},
  year={2023},
  publisher={Springer}
}

@article{han2023survey,
  title={A survey of machine learning and deep learning in remote sensing of geological environment: Challenges, advances, and opportunities},
  author={Han, Wei and Zhang, Xiaohan and Wang, Yi and Wang, Lizhe and Huang, Xiaohui and Li, Jun and Wang, Sheng and Chen, Weitao and Li, Xianju and Feng, Ruyi and others},
  journal={ISPRS Journal of Photogrammetry and Remote Sensing},
  volume={202},
  pages={87--113},
  year={2023},
  publisher={Elsevier}
}

@article{zhang2022deep,
  title={Deep learning for air pollutant concentration prediction: A review},
  author={Zhang, Bo and Rong, Yi and Yong, Ruihan and Qin, Dongming and Li, Maozhen and Zou, Guojian and Pan, Jianguo},
  journal={Atmospheric Environment},
  volume={290},
  pages={119347},
  year={2022},
  publisher={Elsevier}
}

@article{kumar2022critical,
  title={A critical evaluation of air quality index models (1960--2021)},
  author={Kumar, Prashant and others},
  journal={Environmental Monitoring and Assessment},
  volume={194},
  number={5},
  pages={1--45},
  year={2022},
  publisher={Springer}
}

@article{wang2023review,
  title={A review of the technology and applications of methods for evaluating the transport of air pollutants},
  author={Wang, Xiaoqi and Cheng, Shuiyuan and Zhou, Ying and Zhang, Hanyu and Guan, Panbo and Zhang, Zhida and Bai, Weichao and Dai, Wujun},
  journal={Journal of Environmental Sciences},
  volume={123},
  pages={341--349},
  year={2023},
  publisher={Elsevier}
}

@article{he2022spatial,
  title={Spatial heterogeneity of air pollution statistics in Europe},
  author={He, Hankun and Sch{\"a}fer, Benjamin and Beck, Christian},
  journal={Scientific Reports},
  volume={12},
  number={1},
  pages={12215},
  year={2022},
  publisher={Nature Publishing Group UK London}
}

@article{uno2009asian,
  title={Asian dust transported one full circuit around the globe},
  author={Uno, Itsushi and Eguchi, Kenta and Yumimoto, Keiya and Takemura, Toshihiko and Shimizu, Atsushi and Uematsu, Mitsuo and Liu, Zhaoyan and Wang, Zifa and Hara, Yukari and Sugimoto, Nobuo},
  journal={Nature Geoscience},
  volume={2},
  number={8},
  pages={557--560},
  year={2009},
  publisher={Nature Publishing Group UK London}
}

@book{seinfeld2016atmospheric,
  title={Atmospheric chemistry and physics: from air pollution to climate change},
  author={Seinfeld, John H and Pandis, Spyros N},
  year={2016},
  publisher={John Wiley \& Sons}
}

@inproceedings{marobust,
  title={Robust Spatio-Temporal Centralized Interaction for OOD Learning},
  author={Ma, Jiaming and Wang, Binwu and Wang, Pengkun and Zhou, Zhengyang and Wang, Xu and Wang, Yang},
  booktitle={Forty-second International Conference on Machine Learning}
}

@inproceedings{wang2023pattern,
  title={Pattern expansion and consolidation on evolving graphs for continual traffic prediction},
  author={Wang, Binwu and Zhang, Yudong and Wang, Xu and Wang, Pengkun and Zhou, Zhengyang and Bai, Lei and Wang, Yang},
  booktitle={Proceedings of the 29th ACM SIGKDD Conference on Knowledge Discovery and Data Mining},
  pages={2223--2232},
  year={2023}
}

@article{yao2022wild,
  title={Wild-time: A benchmark of in-the-wild distribution shift over time},
  author={Yao, Huaxiu and Choi, Caroline and Cao, Bochuan and Lee, Yoonho and Koh, Pang Wei W and Finn, Chelsea},
  journal={Advances in Neural Information Processing Systems},
  volume={35},
  pages={10309--10324},
  year={2022}
}

@article{karniadakis2021physics,
  title={Physics-informed machine learning},
  author={Karniadakis, George Em and Kevrekidis, Ioannis G and Lu, Lu and Perdikaris, Paris and Wang, Sifan and Yang, Liu},
  journal={Nature Reviews Physics},
  volume={3},
  number={6},
  pages={422--440},
  year={2021},
  publisher={Nature Publishing Group UK London}
}

@article{lu2025causality,
  title={A Causality-Aware Spatiotemporal Model for Multi-Region and Multi-Pollutant Air Quality Forecasting},
  author={Lu, Junxin and Sun, Shiliang},
  journal={arXiv preprint arXiv:2509.21260},
  year={2025}
}

@article{feng2024spatio,
  title={Spatio-temporal field neural networks for air quality inference},
  author={Feng, Yutong and Wang, Qiongyan and Xia, Yutong and Huang, Junlin and Zhong, Siru and Liang, Yuxuan},
  journal={arXiv preprint arXiv:2403.02354},
  year={2024}
}

@inproceedings{wang2025airradar,
  title={Airradar: Inferring nationwide air quality in china with deep neural networks},
  author={Wang, Qiongyan and Xia, Yutong and ZHong, Siru and Li, Weichuang and Wu, Yuankai and Cheng, Shifen and Zhang, Junbo and Zheng, Yu and Liang, Yuxuan},
  booktitle={Proceedings of the AAAI Conference on Artificial Intelligence},
  volume={39},
  number={27},
  pages={28467--28475},
  year={2025}
}

@article{daly2007air,
  title={Air pollution modeling--An overview},
  author={Daly, Aaron and Zannetti, Paolo},
  journal={Ambient air pollution},
  pages={15--28},
  year={2007},
  publisher={The Arab School for Science and Technology}
}

@article{li2023physics,
  title={Physics-informed deep learning to reduce the bias in joint prediction of nitrogen oxides},
  author={Li, Lianfa and Khalili, Roxana and Lurmann, Frederick and Pavlovic, Nathan and Wu, Jun and Xu, Yan and Liu, Yisi and O'Sharkey, Karl and Ritz, Beate and Oman, Luke and others},
  journal={arXiv preprint arXiv:2308.07441},
  year={2023}
}

@inproceedings{zhangeulerian,
  title={Eulerian Neural Network Informed by Chemical Transport for Air Quality Forecasting},
  author={Zhang, Xukai and Wang, Shuliang and Jin, Guangyin and Yuan, Ziqiang and Yuan, Hanning and Ruan, Sijie},
  booktitle={The Thirty-ninth Annual Conference on Neural Information Processing Systems}
}

@article{stock2001vector,
  title={Vector autoregressions},
  author={Stock, James H and Watson, Mark W},
  journal={Journal of Economic perspectives},
  volume={15},
  number={4},
  pages={101--115},
  year={2001},
  publisher={American Economic Association}
}

@inproceedings{burtscher1999exploring,
  title={Exploring last n value prediction},
  author={Burtscher, Martin and Zorn, Benjamin G},
  booktitle={1999 International Conference on Parallel Architectures and Compilation Techniques (Cat. No. PR00425)},
  pages={66--76},
  year={1999},
  organization={IEEE}
}

@article{ando2004geometric,
  title={Geometric means},
  author={Ando, Tsuyoshi and Li, Chi-Kwong and Mathias, Roy},
  journal={Linear algebra and its applications},
  volume={385},
  pages={305--334},
  year={2004},
  publisher={Elsevier}
}

@inproceedings{huang2025std,
  title={Std-plm: Understanding both spatial and temporal properties of spatial-temporal data with plm},
  author={Huang, Yiheng and Mao, Xiaowei and Guo, Shengnan and Chen, Yubin and Shen, Junfeng and Li, Tiankuo and Lin, Youfang and Wan, Huaiyu},
  booktitle={Proceedings of the AAAI Conference on Artificial Intelligence},
  volume={39},
  number={11},
  pages={11817--11825},
  year={2025}
}

@article{uddinscnode,
  title={SCNode: Spatial and Contextual Coordinates for Graph Representation Learning},
  author={Uddin, Md Joshem and Tola, Astrit and Sikand, Varin Singh and Akcora, Cuneyt Gurcan and Coskunuzer, Baris},
  journal={Transactions on Machine Learning Research}
}

@inproceedings{shao2022spatial,
  title={Spatial-temporal identity: A simple yet effective baseline for multivariate time series forecasting},
  author={Shao, Zezhi and Zhang, Zhao and Wang, Fei and Wei, Wei and Xu, Yongjun},
  booktitle={Proceedings of the 31st ACM international conference on information \& knowledge management},
  pages={4454--4458},
  year={2022}
}

@article{wu2020comprehensive,
  title={A comprehensive survey on graph neural networks},
  author={Wu, Zonghan and Pan, Shirui and Chen, Fengwen and Long, Guodong and Zhang, Chengqi and Yu, Philip S},
  journal={IEEE transactions on neural networks and learning systems},
  volume={32},
  number={1},
  pages={4--24},
  year={2020},
  publisher={IEEE}
}

@inproceedings{chen2025stable,
  title={Stable Fair Graph Representation Learning with Lipschitz Constraint},
  author={Chen, Qiang and Wu, Zhongze and Su, Xiu and Lin, Xi and Qu, Zhe and You, Shan and Yang, Shuo and Xu, Chang},
  booktitle={Forty-second International Conference on Machine Learning}
}

@article{bi2023accurate,
  title={Accurate medium-range global weather forecasting with 3D neural networks},
  author={Bi, Kaifeng and Xie, Lingxi and Zhang, Hengheng and Chen, Xin and Gu, Xiaotao and Tian, Qi},
  journal={Nature},
  volume={619},
  number={7970},
  pages={533--538},
  year={2023},
  publisher={Nature Publishing Group UK London}
}

@article{lam2023learning,
  title={Learning skillful medium-range global weather forecasting},
  author={Lam, Remi and Sanchez-Gonzalez, Alvaro and Willson, Matthew and Wirnsberger, Peter and Fortunato, Meire and Alet, Ferran and Ravuri, Suman and Ewalds, Timo and Eaton-Rosen, Zach and Hu, Weihua and others},
  journal={Science},
  volume={382},
  number={6677},
  pages={1416--1421},
  year={2023},
  publisher={American Association for the Advancement of Science}
}

@article{xu2023dynamic,
  title={Dynamic graph neural network with adaptive edge attributes for air quality prediction: A case study in China},
  author={Xu, Jing and Wang, Shuo and Ying, Na and Xiao, Xiao and Zhang, Jiang and Jin, Zhiling and Cheng, Yun and Zhang, Gangfeng},
  journal={Heliyon},
  volume={9},
  number={7},
  year={2023},
  publisher={Elsevier}
}

@article{gao2025oneforecast,
  title={OneForecast: a universal framework for global and regional weather forecasting},
  author={Gao, Yuan and Wu, Hao and Shu, Ruiqi and Dong, Huanshuo and Xu, Fan and Chen, Rui Ray and Yan, Yibo and Wen, Qingsong and Hu, Xuming and Wang, Kun and others},
  journal={arXiv preprint arXiv:2502.00338},
  year={2025}
}

@article{brown2025alphaearth,
  title={Alphaearth foundations: An embedding field model for accurate and efficient global mapping from sparse label data},
  author={Brown, Christopher F and Kazmierski, Michal R and Pasquarella, Valerie J and Rucklidge, William J and Samsikova, Masha and Zhang, Chenhui and Shelhamer, Evan and Lahera, Estefania and Wiles, Olivia and Ilyushchenko, Simon and others},
  journal={arXiv preprint arXiv:2507.22291},
  year={2025}
}

@article{yang2025local,
  title={Local off-grid weather forecasting with multi-modal earth observation data},
  author={Yang, Qidong and Giezendanner, Jonathan and Civitarese, Daniel Salles and Jakubik, Johannes and Schmitt, Eric and Chandra, Anirban and Vila, Jeremy and Hohl, Detlef and Hill, Chris and Watson, Campbell and others},
  journal={Journal of Advances in Modeling Earth Systems},
  volume={17},
  number={12},
  pages={e2025MS005207},
  year={2025},
  publisher={Wiley Online Library}
}

@article{sun2025can,
  title={Can AI weather models predict out-of-distribution gray swan tropical cyclones?},
  author={Sun, Y Qiang and Hassanzadeh, Pedram and Zand, Mohsen and Chattopadhyay, Ashesh and Weare, Jonathan and Abbot, Dorian S},
  journal={Proceedings of the National Academy of Sciences},
  volume={122},
  number={21},
  pages={e2420914122},
  year={2025},
  publisher={National Academy of Sciences}
}
\bibliographystyle{OminiAir/main}

\newpage
\appendix
\onecolumn
\begin{center}
    \vspace*{0.5cm}
    
    {\Large \bfseries Breaking the Regional Barrier: Scalable and Efficient Spatio-Temporal Modeling for Worldwide Air Quality}

    \vspace{2em}
    {\Large \textmd{Appendix}}
    \vspace{1.5em}
\end{center}

\section*{Table of Contents}
\noindent\rule{\linewidth}{1.0pt} 

\vspace{1em}

\renewcommand{\arraystretch}{1.2} 

\begin{tabularx}{\linewidth}{@{}l X r@{}}

    \textbf{A} & \textbf{Theoretical Justifications} \dotfill & \textbf{\pageref{app:theory}} \\
    \addlinespace[0.3em]
       & \hspace{1.5em} A.1 \quad Spectral Analysis of Fourier Positional Encoding \dotfill & \pageref{subsec:theory_fourier_detailed} \\
       & \hspace{1.5em} A.2 \quad Representation Stability via Lipschitz Continuity \dotfill & \pageref{thm:lipschitz} \\
       & \hspace{1.5em} A.3 \quad Deriving AAGD from Reaction-Diffusion Dynamics \dotfill & \pageref{sub:deriving_aagd} \\ 
    \addlinespace[0.8em]

    \textbf{B} & \textbf{Computational Complexity} \dotfill & \textbf{\pageref{sec:complexity}} \\
    \addlinespace[0.8em]

    \textbf{C} & \textbf{Air Quality Prediction Overview} \dotfill & \textbf{\pageref{sec:aqi_definition}} \\
    \addlinespace[0.3em]
       & \hspace{1.5em} C.1 \quad Definition and Motivation \dotfill & \pageref{sub:def_motiv} \\
       & \hspace{1.5em} C.2 \quad Characterization of Major Pollutants \dotfill & \pageref{sub:char_pollutants} \\

    \addlinespace[0.8em]

    \textbf{D} & \textbf{World Air Quality Analysis} \dotfill & \textbf{\pageref{sec:world_analysis}} \\
    \addlinespace[0.3em]
       & \hspace{1.5em} D.1 \quad Inequality and The Digital Divide \dotfill & \pageref{subsec:inequality} \\
       & \hspace{1.5em} D.2 \quad Regional Station Analysis \dotfill & \pageref{app:regional_analysis} \\
    \addlinespace[0.8em]

    \textbf{E} & \textbf{Dataset Details} \dotfill & \textbf{\pageref{app:dataset_details}} \\
    \addlinespace[0.8em]

    \textbf{F} & \textbf{Static Geospatial Feature} \dotfill & \textbf{\pageref{app:static_features}} \\
    \addlinespace[0.3em]
       & \hspace{1.5em} F.1 \quad Rationale for Feature Selection \dotfill & \pageref{sub:feature_rationale} \\
       & \hspace{1.5em} F.2 \quad Contextual Neighborhood Aggregation \dotfill & \pageref{sub:context_agg} \\
       & \hspace{1.5em} F.3 \quad Stratified Feature Encoding \dotfill & \pageref{sub:stratified_enc} \\
       & \hspace{1.5em} F.4 \quad Calculation of Topographic Features \dotfill & \pageref{sub:calc_topo} \\
    \addlinespace[0.8em]

    \textbf{G} & \textbf{OmniAir Framework Implementation (Algorithm)} \dotfill & \textbf{\pageref{app:framework_impl}} \\
    \addlinespace[0.8em]
    
    \textbf{H} & \textbf{Extended Analysis of Adaptive Sparse Topology} \dotfill & \textbf{\pageref{app:adaptive_topology_analysis}} \\
    \addlinespace[0.3em]
       & \hspace{1.5em} H.1 \quad Mathematical Derivation of Differentiable Pruning \dotfill & \pageref{sub:math_pruning} \\
    \addlinespace[0.8em]

    \textbf{I} & \textbf{Global Spatio-Temporal Analysis} \dotfill & \textbf{\pageref{app:global_aqi_overview}} \\
    \addlinespace[0.8em]

    \textbf{J} & \textbf{Implementation and Training Details} \dotfill & \textbf{\pageref{sec:implementation_details}} \\
    \addlinespace[0.3em]
       & \hspace{1.5em} J.1 \quad Experimental Settings \dotfill & \pageref{subsec:experimental_settings} \\
       & \hspace{1.5em} J.2 \quad Evaluation Metrics \dotfill & \pageref{sec:metrics} \\
    \addlinespace[0.8em]

    \textbf{K} & \textbf{Detailed Experimental Results} \dotfill & \textbf{\pageref{sec:experimental_results}} \\
    \addlinespace[0.3em]
       & \hspace{1.5em} K.1 \quad Performance on KnowAir Dataset \dotfill & \pageref{sub:exp_knowair} \\
       & \hspace{1.5em} K.2 \quad Performance on LargeAQ Dataset \dotfill & \pageref{sub:exp_largeaq} \\
       & \hspace{1.5em} K.3 \quad Ablation Study \dotfill & \pageref{ablation} \\
       & \hspace{1.5em} K.4 \quad Statistical Significance Test \dotfill & \pageref{sec:statistics} \\
       & \hspace{1.5em} K.5 \quad Hyperparameter Sensitivity Analysis \dotfill & \pageref{sec:hyperparameter} \\
    \addlinespace[0.8em]

    \textbf{L} & \textbf{Station Level Feature Visualization} \dotfill & \textbf{\pageref{app:station_features_vis}} \\
    \addlinespace[0.3em]
       & \hspace{1.5em} .1 \quad Interpretation of Station Level Features \dotfill & \pageref{sub:interp_station} \\
    \addlinespace[0.8em]

    \textbf{M} & \textbf{Broader Impact Statement} \dotfill & \textbf{\pageref{sec:impact}} \\
    \addlinespace[0.8em]

    \textbf{N} & \textbf{Code and Data Availability} \dotfill & \textbf{\pageref{sec:availability}} \\
    \addlinespace[0.8em]

    \textbf{O} & \textbf{Case Study} \dotfill & \textbf{\pageref{sec:case_study}} \\
    \addlinespace[0.8em]

    \textbf{P} & \textbf{Discussion} \dotfill & \textbf{\pageref{sec:discussion}} \\

\end{tabularx}

\newpage

\section{Theoretical Justifications}
\label{app:theory}
In this section, we provide mathematical justifications for the OmniAir framework. We first present a spectral analysis of the Fourier Feature Mapping, demonstrating via the kernel method that it modulates the spectral bias to capture high-frequency spatial variations. Next, we establish the physical stability of the Inductive Semantic Identity Encoder (ISIE) via Lipschitz continuity analysis. Finally, we derive the spectral filtering mechanism of the Adaptive Air Graph Diffusion directly from the continuous Reaction Diffusion equation, proving that signed aggregation coefficients are mathematically necessary for approximating the differential operators required to identify pollution sources.

\subsection{Spectral Analysis of Fourier Positional Encoding}
\label{subsec:theory_fourier_detailed}

Standard Multi-Layer Perceptrons (MLPs) operating on low dimensional coordinates suffer from \textit{Spectral Bias}, effectively acting as low pass filters that converge extremely slowly on high frequency spatial variations. In this section, we provide a step-by-step derivation proving that our Fourier mapping transforms the effective kernel of the network, enabling the efficient learning of high frequency signals.

\subsubsection{Derivation of the Induced Stationary Kernel}
\label{sub:deriving_aagd}
\begin{definition}[Fourier Feature Mapping]
Let $\mathbf{v} \in \mathbb{R}^d$ be the input coordinate. The mapping $\gamma: \mathbb{R}^d \to \mathbb{R}^{2M}$ is defined as:
\begin{equation}
    \gamma(\mathbf{v}) = \frac{1}{\sqrt{M}} \left[ \cos(2\pi \mathbf{b}_1^\top \mathbf{v}), \dots, \cos(2\pi \mathbf{b}_M^\top \mathbf{v}), \sin(2\pi \mathbf{b}_1^\top \mathbf{v}), \dots, \sin(2\pi \mathbf{b}_M^\top \mathbf{v}) \right]^\top
\end{equation}
where $\mathbf{b}_j \stackrel{i.i.d.}{\sim} p(\mathbf{b})$ (e.g., Gaussian $\mathcal{N}(0, \sigma^2 \mathbf{I})$).
\end{definition}

\begin{theorem}[Convergence to Shift Invariant Kernel]
As the embedding dimension $M \to \infty$, the kernel $K_{\gamma}(\mathbf{x}, \mathbf{y}) = \langle \gamma(\mathbf{x}), \gamma(\mathbf{y}) \rangle$ converges almost surely to a shift invariant kernel $k(\mathbf{x}-\mathbf{y})$ whose Fourier transform is the spectral density $p(\mathbf{b})$.
\end{theorem}

\begin{proof}

\textbf{Expansion of the Dot Product.}
The inner product of the mapped feature vectors for two locations $\mathbf{x}$ and $\mathbf{y}$ is:
\begin{align}
    K_{\gamma}(\mathbf{x}, \mathbf{y}) &= \gamma(\mathbf{x})^\top \gamma(\mathbf{y}) \\
    &= \frac{1}{M} \sum_{j=1}^M \left[ \cos(2\pi \mathbf{b}_j^\top \mathbf{x}) \cos(2\pi \mathbf{b}_j^\top \mathbf{y}) + \sin(2\pi \mathbf{b}_j^\top \mathbf{x}) \sin(2\pi \mathbf{b}_j^\top \mathbf{y}) \right]
\end{align}
We apply the cosine difference identity $\cos(A - B) = \cos A \cos B + \sin A \sin B$. Letting $A = 2\pi \mathbf{b}_j^\top \mathbf{x}$ and $B = 2\pi \mathbf{b}_j^\top \mathbf{y}$, the summation simplifies to:
\begin{equation}
    K_{\gamma}(\mathbf{x}, \mathbf{y}) = \frac{1}{M} \sum_{j=1}^M \cos\left(2\pi \mathbf{b}_j^\top (\mathbf{x} - \mathbf{y})\right)
\end{equation}
Crucially, the kernel now depends only on the difference vector $\Delta = \mathbf{x} - \mathbf{y}$.

\textbf{Convergence to Expectation.}
Since $\mathbf{b}_j \sim p(\mathbf{b})$ are i.i.d. samples, by the Strong Law of Large Numbers, as $M \to \infty$, the Monte Carlo average converges to the expectation:
\begin{equation}
    K_{\infty}(\Delta) = \lim_{M \to \infty} K_{\gamma}(\mathbf{x}, \mathbf{y}) = \mathbb{E}_{\mathbf{b} \sim p(\mathbf{b})} \left[ \cos(2\pi \mathbf{b}^\top \Delta) \right]
\end{equation}
Expressing the expectation as an integral over the density $p(\mathbf{b})$:
\begin{equation}
    K_{\infty}(\Delta) = \int_{\mathbb{R}^d} p(\mathbf{b}) \cos(2\pi \mathbf{b}^\top \Delta) d\mathbf{b}
\end{equation}

\textbf{Recovery of the Fourier Transform.}
Using Euler's formula $e^{i\theta} = \cos \theta + i \sin \theta$, we write $\cos \theta = \text{Re}(e^{i\theta})$. The integral becomes:
\begin{align}
    K_{\infty}(\Delta) &= \text{Re} \left( \int_{\mathbb{R}^d} p(\mathbf{b}) e^{i 2\pi \mathbf{b}^\top \Delta} d\mathbf{b} \right) \\
    &= \int_{\mathbb{R}^d} p(\mathbf{b}) \cos(2\pi \mathbf{b}^\top \Delta) d\mathbf{b} + i \underbrace{\int_{\mathbb{R}^d} p(\mathbf{b}) \sin(2\pi \mathbf{b}^\top \Delta) d\mathbf{b}}_{=0}
\end{align}
The imaginary term integrates to zero because $p(\mathbf{b})$ is chosen to be an even function (symmetric about the origin, e.g., Gaussian) while $\sin(\cdot)$ is an odd function. Thus, $K_{\infty}(\Delta)$ is exactly the Fourier transform of the spectral density $p(\mathbf{b})$.

\end{proof}

\subsubsection{Implication for High Frequency Learning}

This derivation leads to a direct control mechanism for the spectral bias via the bandwidth parameter $\sigma$:

\begin{corollary}[Uncertainty Principle and Bandwidth Control]
Let $p(\mathbf{b}) = \mathcal{N}(0, \sigma^2 \mathbf{I})$. The induced kernel is Gaussian:
\begin{equation}
    K_{\infty}(\Delta) = \exp\left(-\frac{1}{2} (2\pi \sigma \|\Delta\|)^2\right) = \exp\left(-2\pi^2 \sigma^2 \|\Delta\|^2\right)
\end{equation}
According to the uncertainty principle of Fourier analysis, the variance in the spatial domain is inversely proportional to the variance in the frequency domain.
\begin{itemize}
    \item \textbf{Low $\sigma$ (Standard Coordinates):} $p(\mathbf{b})$ is narrow $\implies$ Kernel $K(\Delta)$ is wide. This results in rapid eigenvalue decay, causing the network to ignore high frequency details.
    \item \textbf{High $\sigma$ (Our Method):} By increasing $\sigma$, we widen $p(\mathbf{b})$, which sharpens the Kernel $K(\Delta)$ towards a Dirac delta. A sharper kernel possesses a flatter eigenvalue spectrum, ensuring that high frequency eigenvalues $\lambda_k$ remain large enough to be learned efficiently by Gradient Descent.
\end{itemize}
\end{corollary}

\textbf{Conclusion:} In our framework, the input $\mathbf{v}$ represents latitude and longitude. Since raw coordinates are numerically dense, a standard MLP operates in the Low $\sigma$ regime and over-smooths the signal. By utilizing the Fourier mapping with a tuned $\sigma$, we project these coordinates into a high dimensional manifold where spatially adjacent locations become distinguishable, enabling the model to capture high frequency spatial heterogeneity effectively.

\subsection{Representation Stability via Lipschitz Continuity}
\label{thm:lipschitz}
The ISIE module maps physical features $\mathbf{x}$ to an identity embedding $\Phi(\mathbf{x})$. To ensure physical robustness, we require that the encoder is stable: bounded perturbations in the input should not yield divergent embeddings.

\begin{definition}[$K$-Lipschitz Continuity]
A function $f: \mathbb{R}^n \to \mathbb{R}^m$ is called $K$-Lipschitz continuous if there exists a real constant $K \ge 0$ such that for all $\mathbf{u}, \mathbf{v} \in \mathbb{R}^n$:
\begin{equation}
    \|f(\mathbf{u}) - f(\mathbf{v})\|_2 \le K \|\mathbf{u} - \mathbf{v}\|_2
\end{equation}
\end{definition}

\begin{theorem}[Stability Bound for Inductive Encoder]
\label{thm:lipschitz}
Let the encoder $\Phi(\mathbf{x})$ be an $L$-layer perceptron defined recursively as:
\begin{equation}
    \mathbf{h}^{(0)} = \mathbf{x}, \quad \mathbf{h}^{(l)} = \sigma(\mathbf{W}_l \mathbf{h}^{(l-1)}), \quad \Phi(\mathbf{x}) = \mathbf{W}_L \mathbf{h}^{(L-1)}
\end{equation}
where $\sigma(\cdot)$ is a 1-Lipschitz activation function (e.g., ReLU, Tanh) and $\mathbf{W}_l$ are learnable weight matrices. For any two inputs $\mathbf{x}$ and $\mathbf{y}$, the distance in the embedding space is strictly bounded by:
\begin{equation}
    \|\Phi(\mathbf{x}) - \Phi(\mathbf{y})\|_2 \le \left( \prod_{l=1}^L \|\mathbf{W}_l\|_2 \right) \|\mathbf{x} - \mathbf{y}\|_2
\end{equation}
where $\|\mathbf{W}_l\|_2$ denotes the spectral norm (largest singular value) of the weight matrix at layer $l$.
\end{theorem}

\begin{proof}
We proceed by induction on the depth $L$.

\textbf{Base Case ($l=1$):} Let $\mathbf{h}^{(1)} = \sigma(\mathbf{W}_1 \mathbf{x})$. Using the 1-Lipschitz property of $\sigma$ ($ \|\sigma(\mathbf{a}) - \sigma(\mathbf{b})\| \le \|\mathbf{a} - \mathbf{b}\| $) and the definition of the spectral norm $\|\mathbf{W}\|_2 = \sup_{\mathbf{v} \neq 0} \frac{\|\mathbf{W}\mathbf{v}\|_2}{\|\mathbf{v}\|_2}$:
\begin{align}
    \|\mathbf{h}^{(1)}_{x} - \mathbf{h}^{(1)}_{y}\|_2 &= \|\sigma(\mathbf{W}_1 \mathbf{x}) - \sigma(\mathbf{W}_1 \mathbf{y})\|_2 \\
    &\le \|\mathbf{W}_1 (\mathbf{x} - \mathbf{y})\|_2 \\
    &\le \|\mathbf{W}_1\|_2 \|\mathbf{x} - \mathbf{y}\|_2
\end{align}

\textbf{Inductive Step:} Assume the bound holds for layer $l-1$ with constant $K_{l-1} = \prod_{k=1}^{l-1} \|\mathbf{W}_k\|_2$, i.e., $\|\mathbf{h}^{(l-1)}_x - \mathbf{h}^{(l-1)}_y\| \le K_{l-1} \|\mathbf{x} - \mathbf{y}\|$.
For layer $l$, we have $\mathbf{h}^{(l)} = \sigma(\mathbf{W}_l \mathbf{h}^{(l-1)})$. Applying the properties again:
\begin{align}
    \|\mathbf{h}^{(l)}_{x} - \mathbf{h}^{(l)}_{y}\|_2 &\le \|\mathbf{W}_l (\mathbf{h}^{(l-1)}_{x} - \mathbf{h}^{(l-1)}_{y})\|_2 \\
    &\le \|\mathbf{W}_l\|_2 \|\mathbf{h}^{(l-1)}_{x} - \mathbf{h}^{(l-1)}_{y}\|_2 \\
    &\le \|\mathbf{W}_l\|_2 \cdot \left( K_{l-1} \|\mathbf{x} - \mathbf{y}\|_2 \right)
\end{align}
Substituting $K_{l-1}$ yields the product of norms: $\prod_{k=1}^{l} \|\mathbf{W}_k\|_2$.
\end{proof}

\textbf{Conclusion:} This theorem proves that the learned manifold of station identities is \textit{smooth} with respect to physical attributes. Moreover, considering the domain of physical attributes $\mathcal{X}$ forms a \textbf{compact metric space}, the continuity of $\Phi$ theoretically guarantees that the resulting identity manifold remains bounded and well behaved, preventing representation divergence irrespective of extreme parameter configurations.

\subsection{Deriving AAGD from Reaction Diffusion Dynamics}
\label{sub:deriving_aagd}
We now demonstrate that standard GNNs are physically insufficient for air quality modeling because they inherently suppress the "source term" in atmospheric dynamics. We prove that the AAGD module, through its signed coefficients, provides the necessary mathematical operators to recover these sources.

\subsubsection{Discrete PDE and Source Identification}

The continuous spatio-temporal dynamics of air pollutants are governed by the Reaction Diffusion equation:
\begin{equation}
    \frac{\partial C}{\partial t} = \underbrace{\mathcal{D} \nabla^2 C}_{\text{Diffusion}} + \underbrace{S(x, t)}_{\text{Source}} - \underbrace{\gamma C}_{\text{Decay}}
    \label{eq:pde}
\end{equation}
where $C$ is concentration, $\mathcal{D}$ is the diffusion coefficient, $\nabla^2$ is the Laplacian operator, and $S$ represents emission sources.

To solve this on a graph $\mathcal{G}=(\mathcal{V}, \mathcal{E})$, we discretize the spatial domain. The continuous Laplacian $\nabla^2 C$ is approximated by the Graph Laplacian operator $-\mathbf{L}\mathbf{C}$:
\begin{equation}
    (\nabla^2 C)_i \approx \sum_{j \in \mathcal{N}_i} A_{ij}(C_j - C_i) = -[\mathbf{L}\mathbf{C}]_i
\end{equation}
Using forward Euler discretization for time with step $\Delta t$, Eq. \eqref{eq:pde} becomes:
\begin{equation}
    \frac{\mathbf{C}_{t+1} - \mathbf{C}_t}{\Delta t} = -\mathcal{D} \mathbf{L} \mathbf{C}_t + \mathbf{S}_t - \gamma \mathbf{C}_t
\end{equation}
To forecast air quality, the model must implicitly infer the source term $\mathbf{S}_t$ (e.g., sudden emissions) from the observed concentration history. Rearranging the discrete equation reveals the exact structure of $\mathbf{S}_t$:
\begin{equation}
    \mathbf{S}_t = \underbrace{\frac{\mathbf{C}_{t+1} - \mathbf{C}_t}{\Delta t}}_{\text{Temporal Change}} + \underbrace{\mathcal{D} \mathbf{L} \mathbf{C}_t}_{\text{Spatial Laplacian}} + \underbrace{\gamma \mathbf{C}_t}_{\text{Decay}}
    \label{eq:source_term}
\end{equation}
Crucially, recovering the source term $\mathbf{S}_t$ requires the model to represent the spatial Laplacian $\mathbf{L} \mathbf{C}_t$.

\subsubsection{Limitation of Positive Aggregation (Standard GNNs)}

Standard Graph Attention Networks (GATs) compute node updates via a convex combination of neighbors:
\begin{equation}
    \mathbf{h}_i' = \sum_{j \in \mathcal{N}_i \cup \{i\}} \alpha_{ij} \mathbf{h}_j, \quad \text{with } \alpha_{ij} \ge 0, \sum \alpha_{ij} = 1
\end{equation}
Geometrically, the output $\mathbf{h}_i'$ is constrained to lie within the \textbf{Convex Hull} of the input features $\{\mathbf{h}_j\}$. This is strictly a smoothing operation (low-pass filter). It is mathematically impossible for a convex combination to generate a value outside the range of the inputs, which is necessary to represent a "source" or a sharp gradient. Thus, standard GNNs structurally equate to the diffusion term but cannot model the generation term $\mathbf{S}_t$.

\subsubsection{Necessity of Signed Coefficients for Source Recovery}

Recall the definition of the graph Laplacian operator on a signal $\mathbf{x}$:
\begin{equation}
    [\mathbf{L}\mathbf{x}]_i = \mathbf{x}_i - \sum_{j \in \mathcal{N}_i} \frac{1}{\sqrt{d_i d_j}} \mathbf{x}_j
\end{equation}
This operator is an affine combination that inherently requires mixed signs: a \textbf{positive coefficient} ($+1$) for the central node and \textbf{negative coefficients} for the neighbors to compute the spatial difference (gradient).

\begin{theorem}[Necessity of Signed Weights]
Let the aggregation function be linear: $\mathbf{h}_i' = \sum_{j} \theta_{ij} \mathbf{h}_j$.
\begin{enumerate}
    \item \textbf{Positive Constraint (Softmax):} If $\theta_{ij} \ge 0$, the aggregation is limited to smoothing. It cannot approximate the Laplacian $\mathbf{L}$ because it cannot compute differences.
    \item \textbf{Signed Capability (Our Method):} By utilizing Tanh attention, AAGD allows $\theta_{ij} \in [-1, 1]$. This enables the model to assign $\theta_{ii} > 0$ and $\theta_{ij} < 0$ (for $j \neq i$), effectively approximating the discrete Laplacian $\mathbf{L}$.
\end{enumerate}
\end{theorem}
\textbf{Conclusion:} The signed coefficients provided by AAGD are not merely an architectural choice but a mathematical necessity to satisfy Eq. \eqref{eq:source_term}, allowing the network to disentangle emission sources ($\mathbf{S}_t$) from natural diffusion processes.

\section{Computational Complexity}
\label{sec:complexity}

Scalability is a prerequisite for global-scale air quality forecasting, where the number of monitoring stations ($N$) reaches the order of $10^4$, significantly exceeding the scale of typical urban datasets. In this section, we provide a rigorous comparative analysis of the time complexity between OmniAir and existing state-of-the-art architectures. Let $D$ denote the feature dimension, $L$ the number of network layers, $T$ the input sequence length, and $K$ the effective neighborhood size (where $K \ll N$).

\paragraph{Complexity Bottlenecks in Existing Models.}
State-of-the-art spatio-temporal forecasting models often encounter prohibitive computational costs when scaling to global networks due to their dense connectivity assumptions. Spectral GNNs (e.g., StemGNN, GWNet) typically rely on the Graph Fourier Transform or learnable adaptive adjacency matrices to capture latent spatial correlations. These approaches necessitate constructing and operating on dense $N \times N$ dependency matrices, resulting in a quadratic complexity of $\mathcal{O}(N^2 \cdot D)$ that exhausts GPU memory rapidly as $N$ increases. Similarly, Standard Transformers (e.g., AirFormer) employ global self attention mechanisms. Calculating the attention scores requires computing the pairwise dot product between all station queries and keys, which inherently incurs an $\mathcal{O}(T \cdot N^2 \cdot D)$ computational bottleneck. While Linear Transformers (e.g., Informer) utilize ProbSparse attention mechanisms to reduce this cost to $\mathcal{O}(T \cdot N \log N \cdot D)$, they achieve this by sampling a subset of queries, which inevitably leads to information loss and reduced precision in station level pollutant inference.

\paragraph{Linear Scalability of OmniAir.}
In contrast, OmniAir is engineered with a strict linear scalability constraint suitable for global deployment. We decouple the computational cost into a one time initialization phase and a recurring runtime phase.
During the Initialization Phase, we employ the Identity Encoder to map static attributes to identity embeddings. Based on these embeddings and geographic coordinates, we construct the hybrid graph topology using efficient spatial indexing structures. This step incurs a complexity of $\mathcal{O}(N \log N \cdot D)$, but since it is a one off pre-computation process, it does not impact the training or inference speed.

Crucially, during the Runtime Phase (Training and Inference), OmniAir adopts a static structure, dynamic weight strategy to ensure efficiency. The graph topology remains fixed, restricting the receptive field of each node to its top $K$ neighbors. Consequently, the dynamic weight modulation in the DSTG module and the message passing in the AAGD module are implemented using Sparse Matrix Multiplication. Since the number of edges is $|\mathcal{E}| \approx N \times K$, the complexity for propagating information over $L$ layers is strictly linear with respect to the number of stations, i.e., $\mathcal{O}(T \cdot L \cdot N \cdot K \cdot D)$. This design avoids any $N \times N$ dense matrix operations, ensuring that both memory usage and computation time grow linearly with the network size.

\paragraph{Summary.}
As summarized in Table \ref{tab:complexity_comparison}, OmniAir achieves significant efficiency gains over quadratic baselines. By amortizing the structure learning cost during initialization and strictly enforcing sparse operations during runtime, OmniAir represents the only viable solution among the compared methods for real time global-scale forecasting with limited computational resources.

\begin{table}[h]
    \centering
    \caption{Theoretical Time Complexity Comparison regarding the number of nodes $N$. ($K$: Neighbor size where $K \ll N$)}
    \label{tab:complexity_comparison}
    \resizebox{0.6\linewidth}{!}{
    \begin{tabular}{l l l}
        \toprule
        \textbf{Model Category} & \textbf{Representative Models} & \textbf{Time Complexity} \\
        \midrule
        Spectral GNN & StemGNN / GWNet & $\mathcal{O}(N^2)$ \\
        Standard Transformer & AirFormer & $\mathcal{O}(N^2)$ \\
        Linear Transformer & Informer & $\mathcal{O}(N \log N)$ \\
        \midrule
        \textbf{OmniAir (Ours)} & \textbf{Initialization} & $\mathcal{O}(N \log N)$ \\
        & \textbf{Training / Inference} & $\mathbf{\mathcal{O}(N \cdot K)}$ \\
        \bottomrule
    \end{tabular}
    }
\end{table}

\section{Air Quality Prediction}
\label{sec:aqi_definition}

\subsection{Definition and Motivation}
\label{sub:def_motiv}
The Air Quality Index (AQI) is a dimensionless, composite indicator used by government agencies to communicate the current and forecasted degree of air pollution to the public. It transforms complex concentrations of multiple pollutants into a single numerical value, categorizing air quality levels from "Good" to "Hazardous." Formally, the AQI is determined by the maximum sub index of major criterion pollutants, calculated as:
\begin{equation}
    AQI = \max(I_{PM2.5}, I_{PM10}, I_{O_3}, I_{NO_2}, I_{SO_2}, I_{CO})
\end{equation}
where $I_p$ represents the individual sub index for pollutant $p$.

The imperative for accurate AQI prediction stems from its profound implications for public health and urban management. Air pollution is a leading environmental risk factor, contributing to respiratory infections, cardiovascular diseases, and lung cancer. A precise, station level prediction system acts as a critical early warning mechanism, enabling sensitive populations to take preventive measures and empowering policymakers to implement dynamic emission control strategies before pollution episodes escalate.

\subsection{Characterization of Major Pollutants}
\label{sub:char_pollutants}

\begin{figure*}[htbp]
    \centering
    \begin{subfigure}[b]{0.32\linewidth}
        \includegraphics[width=\linewidth]{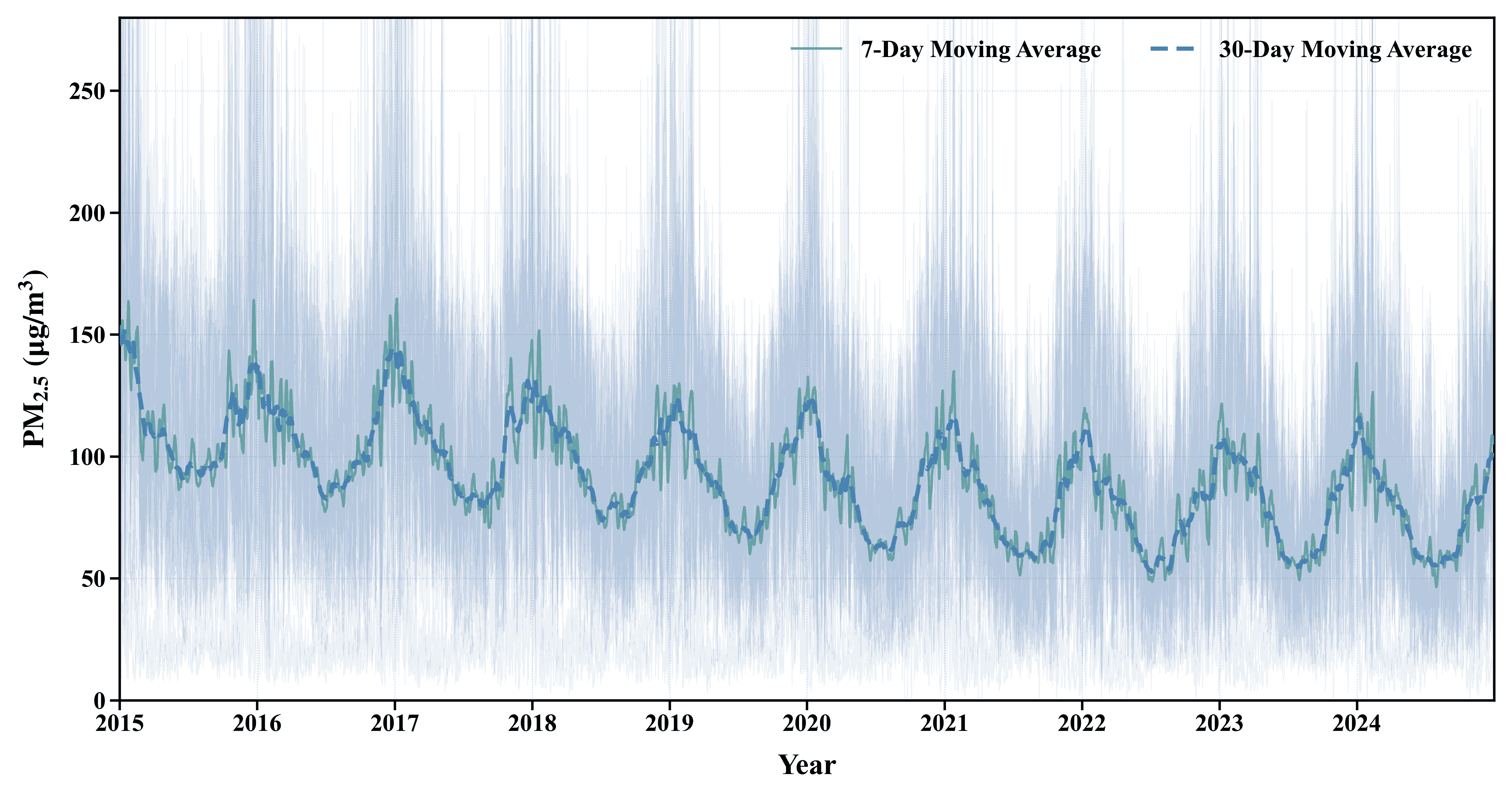}
        \caption{PM$_{2.5}$}
        \label{fig:pm25}
    \end{subfigure}
    \hfill
    \begin{subfigure}[b]{0.32\linewidth}
        \includegraphics[width=\linewidth]{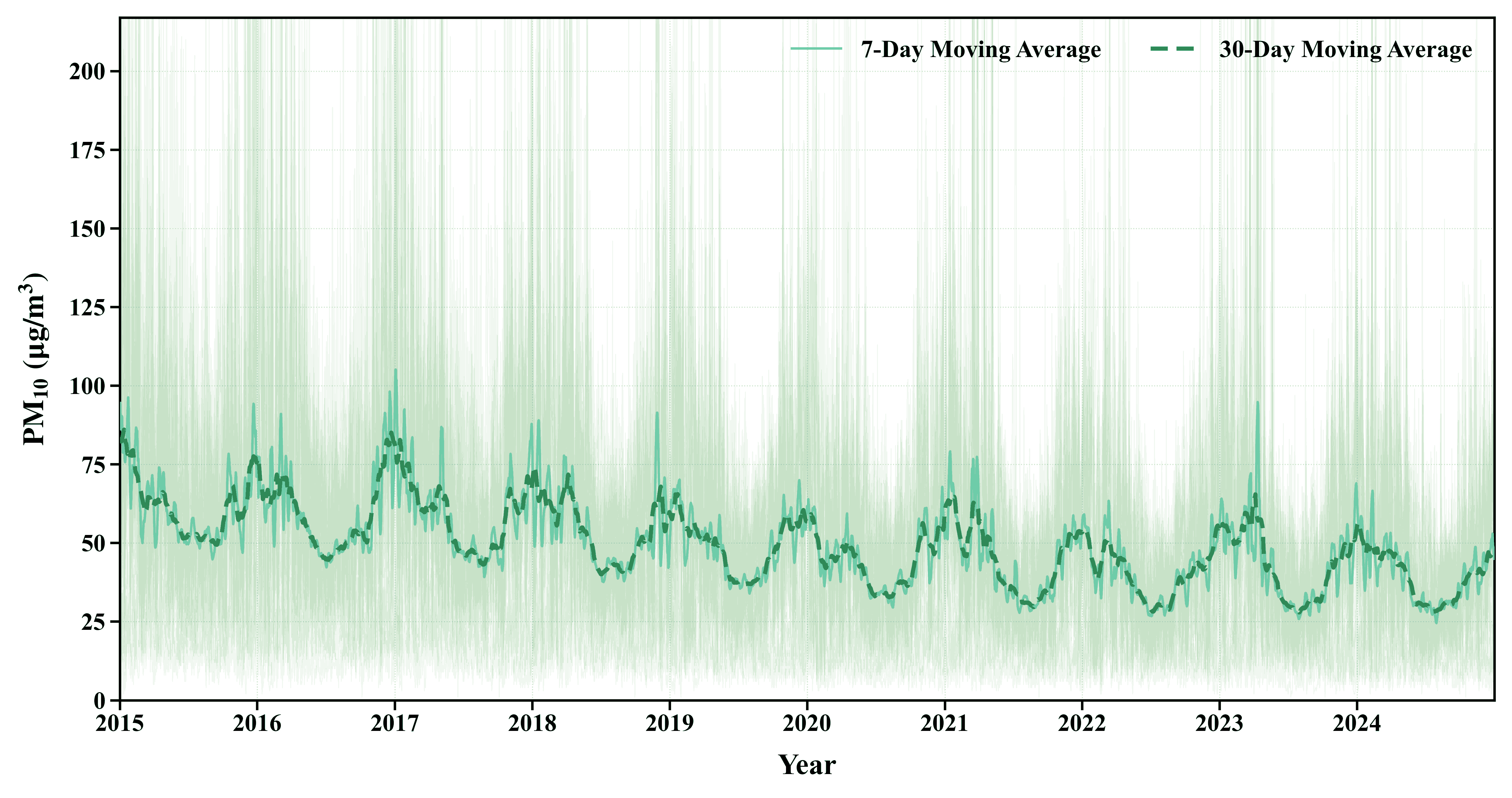}
        \caption{PM$_{10}$}
        \label{fig:pm10}
    \end{subfigure}
    \hfill
    \begin{subfigure}[b]{0.32\linewidth}
        \includegraphics[width=\linewidth]{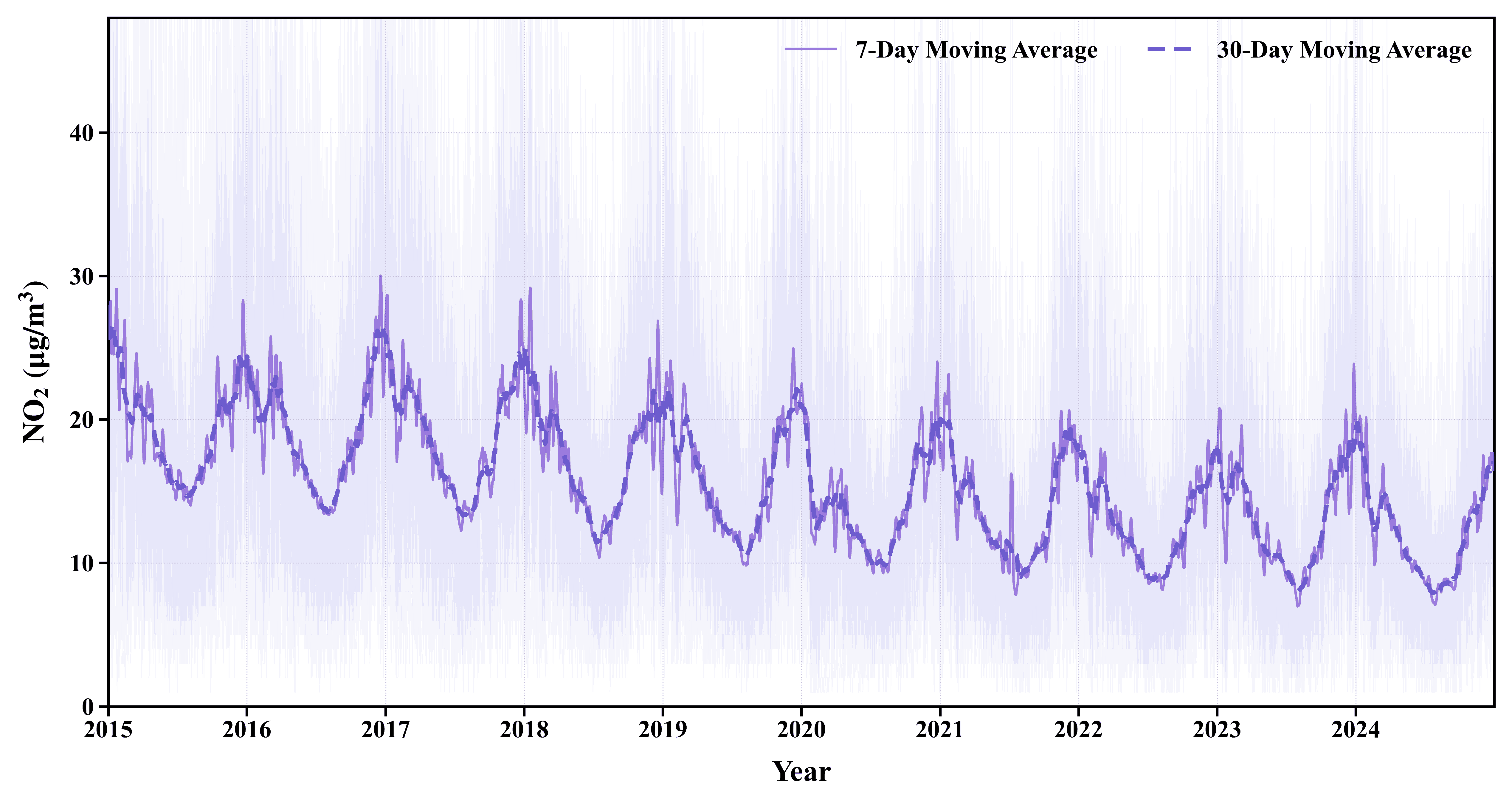}
        \caption{NO$_2$}
        \label{fig:no2}
    \end{subfigure}
    
    \vspace{0.3cm} 
    \begin{subfigure}[b]{0.32\linewidth}
        \includegraphics[width=\linewidth]{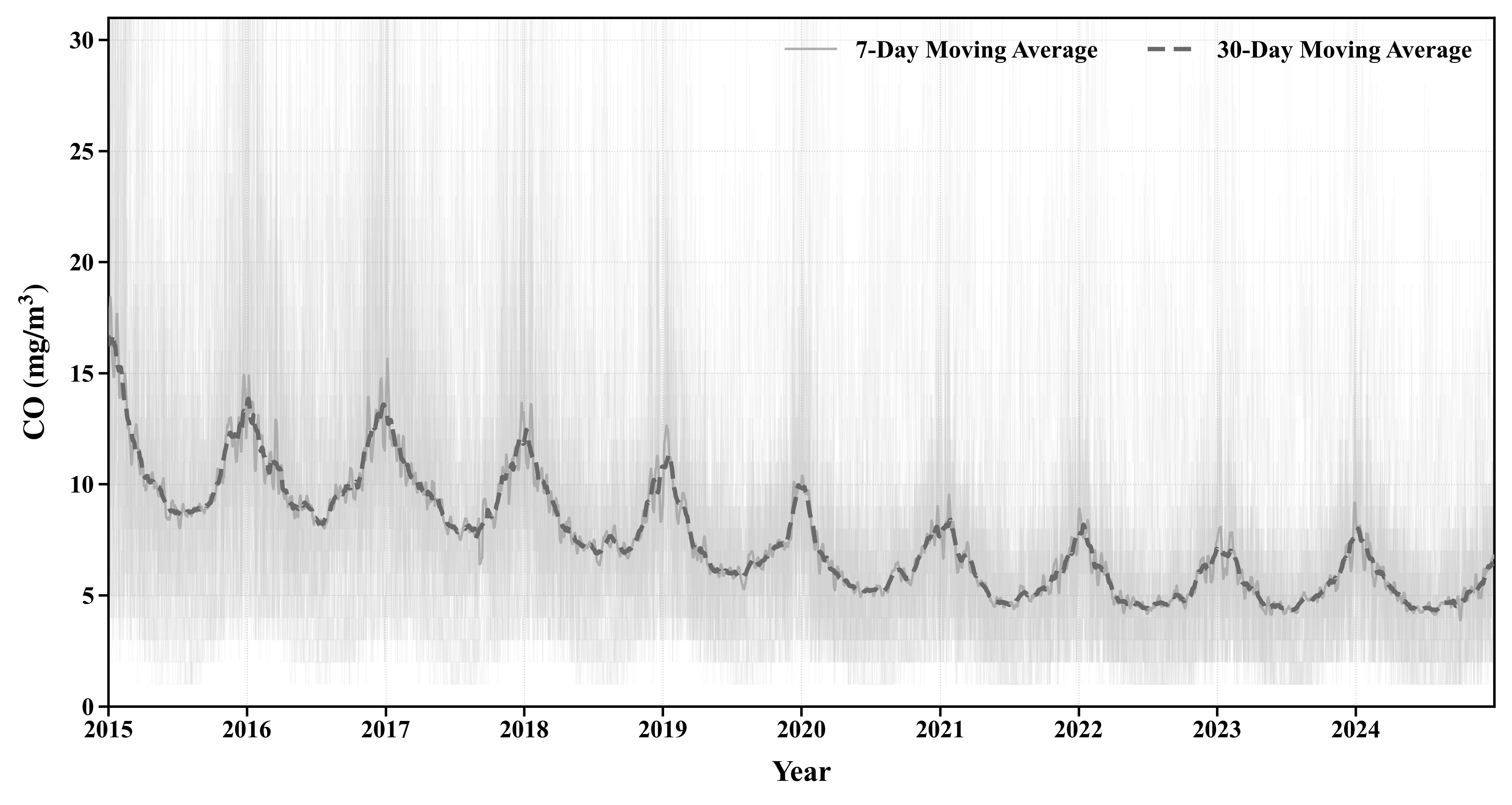}
        \caption{CO}
        \label{fig:co}
    \end{subfigure}
    \hfill
    \begin{subfigure}[b]{0.32\linewidth}
        \includegraphics[width=\linewidth]{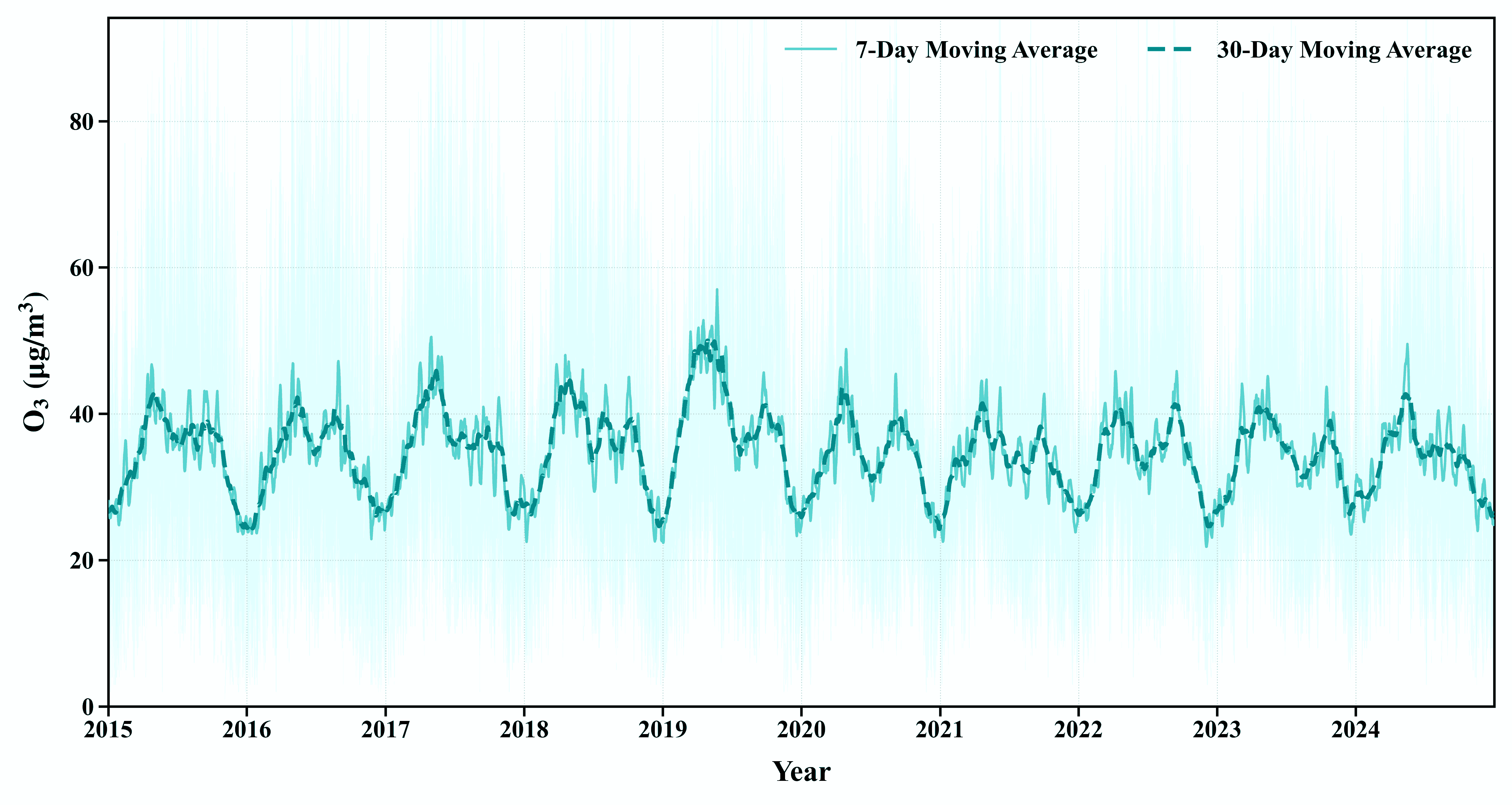}
        \caption{O$_3$}
        \label{fig:o3}
    \end{subfigure}
    \hfill
    \begin{subfigure}[b]{0.32\linewidth}
        \includegraphics[width=\linewidth]{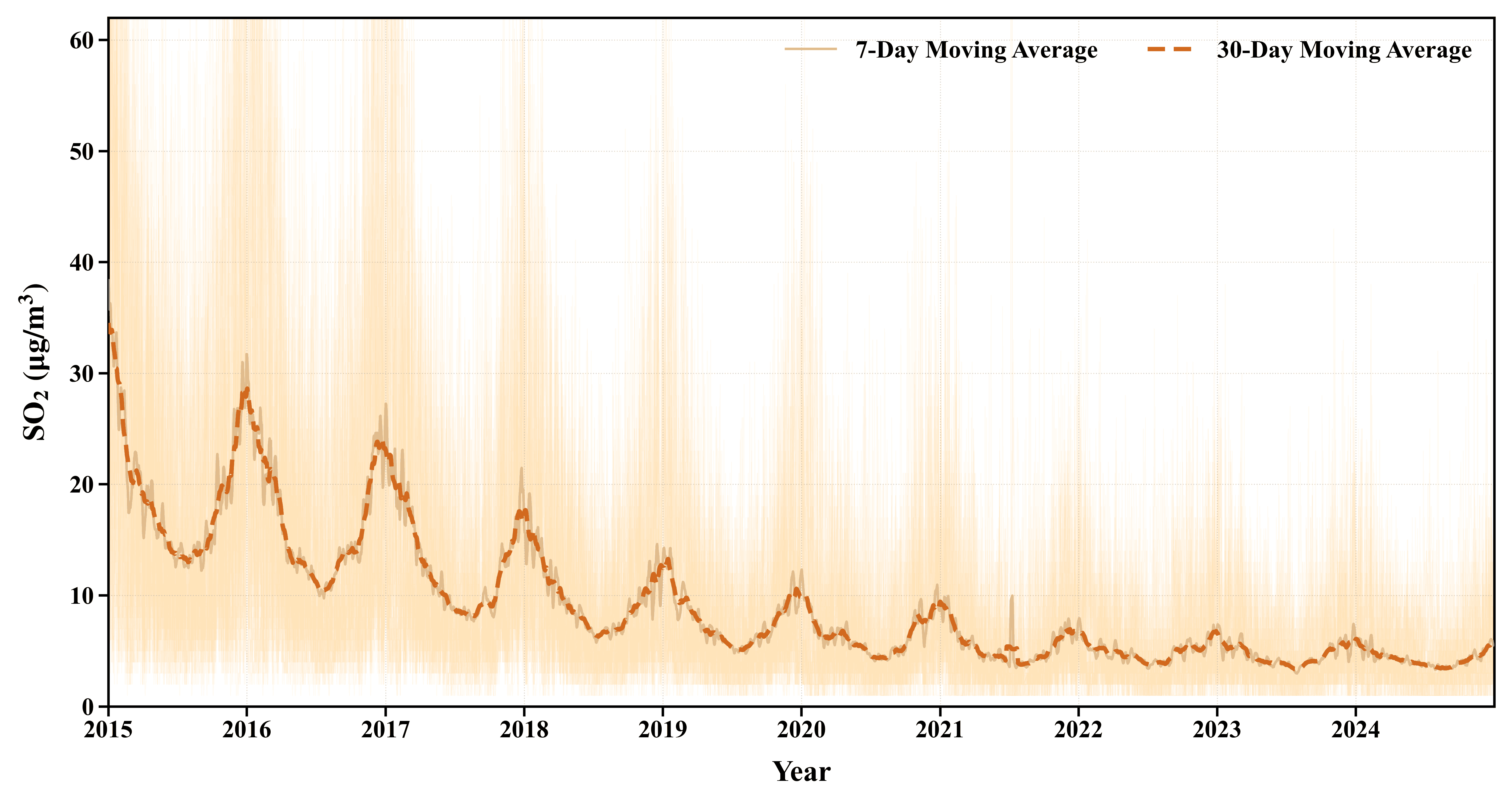}
        \caption{SO$_2$}
        \label{fig:so2}
    \end{subfigure}
    
    \caption{\textbf{Temporal Evolution of Major Pollutants.} The time series reveal distinct physicochemical patterns: (a-b) Particulates show winter seasonality; (c-d) NO$_2$ and CO align with anthropogenic cycles; (e) O$_3$ peaks in summer due to photochemistry; and (f) SO$_2$ reflects industrial trends.}
    \label{fig:pollutants_overview}
\end{figure*}

Understanding the distinct physicochemical dynamics of pollutants is crucial for effective modeling. As illustrated in Figure \ref{fig:pollutants_overview}, we analyze the temporal evolution of six major pollutants:

\textbf{Particulate Matter (PM$_{2.5}$ \& PM$_{10}$):} 
PM$_{2.5}$ and PM$_{10}$ (Fig. \ref{fig:pm25} and \ref{fig:pm10}) exhibit strong seasonal fluctuations driven by winter heating emissions and atmospheric stability. While PM$_{2.5}$ is prone to transboundary transmission, PM$_{10}$ is more localized, originating from mechanical processes like dust resuspension, resulting in higher volatility.

\textbf{Primary Gaseous Pollutants (NO$_2$, CO, SO$_2$):} 
NO$_2$ (Fig. \ref{fig:no2}) and CO (Fig. \ref{fig:co}) serve as key tracers for anthropogenic activity, with concentrations correlating strongly with traffic density and urban combustion. SO$_2$ (Fig. \ref{fig:so2}) shows a long-term decline due to desulfurization policies but retains seasonality in industrial regions.

\textbf{Secondary Pollutant (O$_3$):} 
In contrast to primary pollutants, Ozone (Fig. \ref{fig:o3}) is formed via photochemical reactions. Consequently, it displays an inverse seasonal trend, peaking in summer due to intense solar radiation.

\section{World Air Quality Analysis}
\label{sec:world_analysis}
\begin{figure}[htbp]
    \centering
    \includegraphics[width=0.8\linewidth]{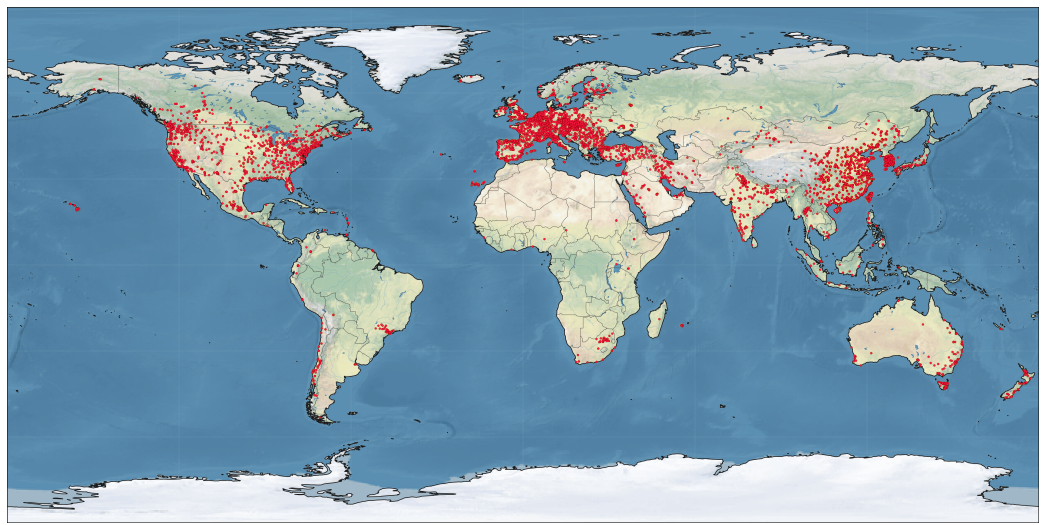}
    \caption{Geographical distribution of the air quality.}
    \label{fig:stations}
\end{figure}

Station distribution across diverse, uneven regions is shown in Fig.~\ref{fig:stations}.

To provide a comprehensive understanding of the global air quality landscape, we conducted a spatial and compositional analysis of the monitoring data. This analysis focuses on the geographical distribution of monitoring stations and the statistical distribution of major pollutants.

\subsection{Inequality and The Digital Divide}
\label{subsec:inequality}

While our dataset represents one of the most comprehensive archives of ground based air quality monitoring—encompassing major human settlements across all inhabited continents—a critical analysis reveals a profound spatial inequality in global monitoring infrastructure.
\begin{figure}[htbp]
    \centering
    \includegraphics[width=0.55\linewidth]{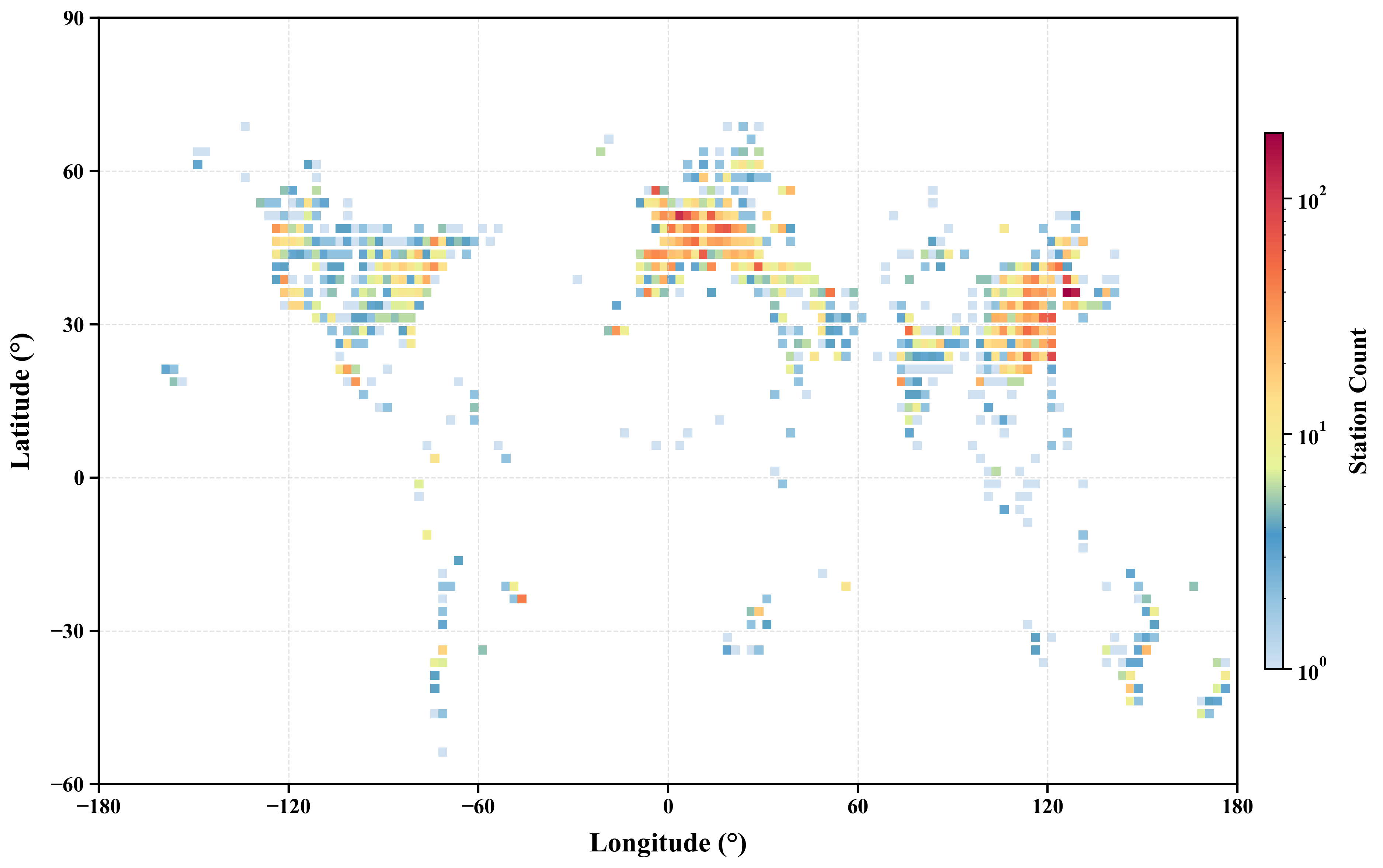}
    \caption{\textbf{Global Station Density Heatmap.} The visualization highlights the severe imbalance in monitoring infrastructure, with high density in the Northern Hemisphere (dark red) and significant data vacuums in the Global South.}
    \label{fig:density_red}
\end{figure}

\textbf{The Monitoring Gap.} As visualized in the station density heatmap (Figure \ref{fig:density_red}), the distribution of monitoring resources is heavily skewed towards the Global North. Developed regions, such as North America and Western Europe, along with rapidly industrializing economies like Eastern China, benefit from dense, redundant sensor networks. In stark contrast, vast regions in the Global South, particularly across Africa, South America, and Southeast Asia, suffer from severe data sparsity. This "digital divide" creates significant observational blind spots, leaving millions of people in developing nations without access to reliable, real time air quality information.

\begin{figure}[htbp]
    \centering
    \includegraphics[width=0.35\linewidth]{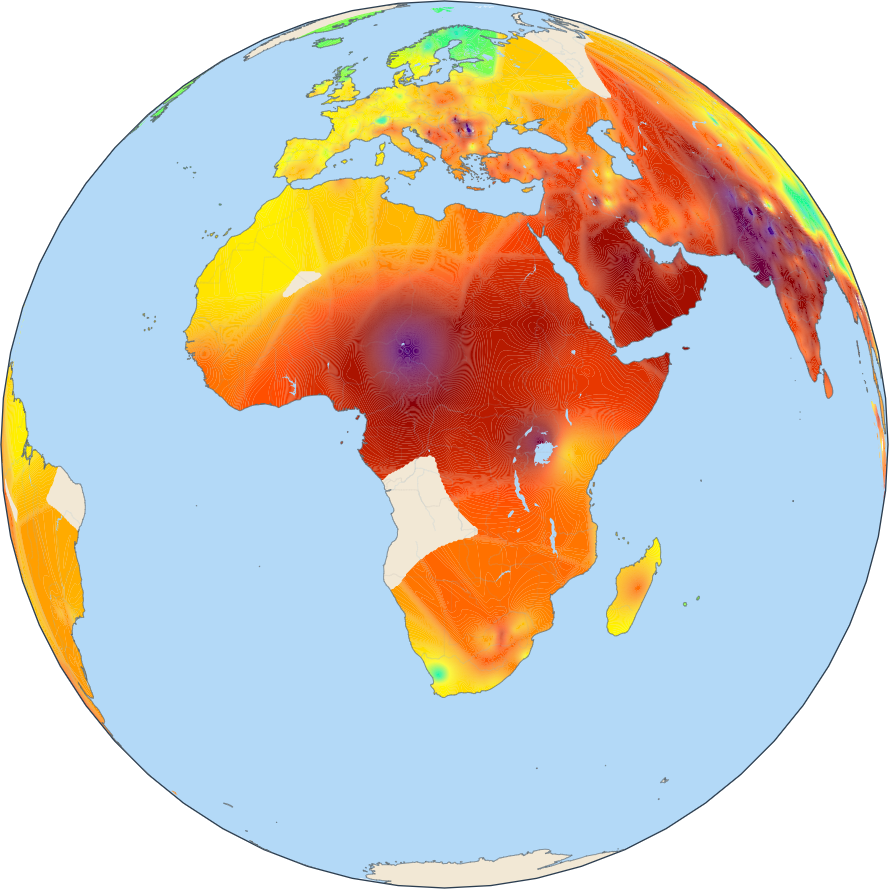}
    \caption{\textbf{Global PM$_{2.5}$ Distribution (Africa centered View).} The 3D orthographic projection reveals the stark contrast between data-rich regions (Europe, East Asia) and the severe monitoring gaps across the African continent and the Middle East, where elevated pollution levels (red-purple) coincide with minimal ground based observation capacity.}
    \label{fig:globe_africa}
\end{figure}

\subsection{Regional Station Analysis}
\label{app:regional_analysis}

\begin{figure}[htbp]
    \centering
    \includegraphics[width=0.6\linewidth]{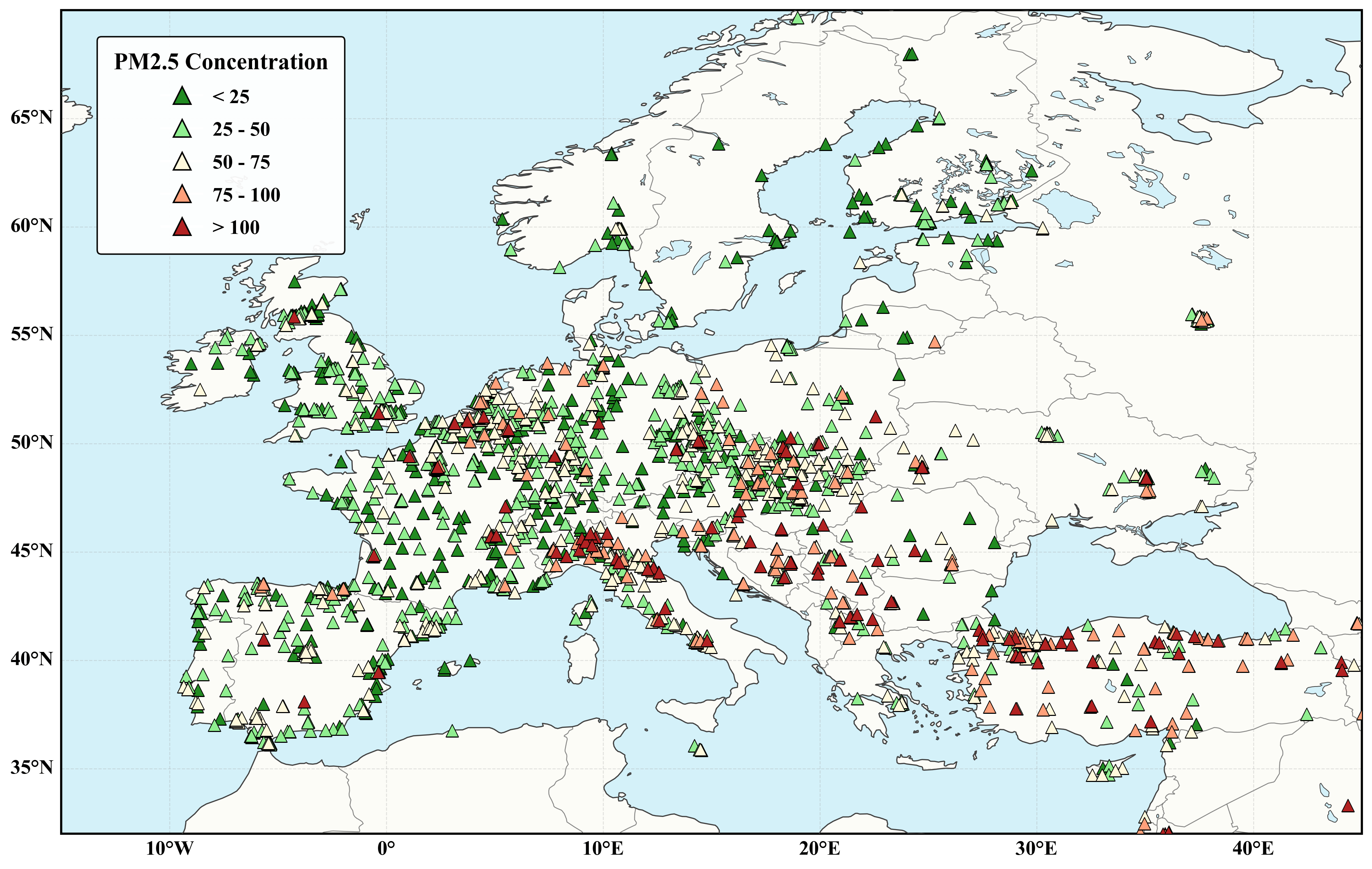}
    \caption{Visualization of the Europe stations highlighting dense local clustering.}
    \label{fig:flow}
\end{figure}

To further elucidate the challenges posed by regional density heterogeneity, we present a detailed visualization of the European monitoring subnet in Figure~\ref{fig:flow}. This region represents a quintessential data-rich environment, contrasting sharply with data-sparse regions like Africa or South America.

The visualization also highlights significant spatial heterogeneity in PM$_{2.5}$ concentrations. We observe distinct pollution hotspots (indicated by red/orange markers) in regions such as the \textbf{Po Valley in Northern Italy} and parts of \textbf{Eastern Europe} (e.g., Poland and the Balkans).  These hotspots are often sharply delineated from nearby clean areas (green markers) by topographic barriers like the Alps. For instance, the transition from the high pollution basin to the clean Alpine slopes occurs over a short geographic distance, creating steep pollution gradients. This necessitates the \textbf{Contextual Neighborhood Aggregation} mechanism in our ISIE module, specifically the pollution weighted centroid features, to capture these directional biases and anisotropic diffusion patterns that simple coordinate based encodings fail to resolve.

\section{Dataset Details}
\label{app:dataset_details}

\paragraph{\textbf{Fine grained Validation.}}
To further verify the prediction performance of our model and the effectiveness of the supplementary weather covariates at finer temporal granularities, we incorporate two national level datasets:
\begin{itemize}
    \item \textbf{LargeAQ (1-Hour Granularity):} A station level dataset covering \textbf{1,341 stations} across 333 major Chinese cities (2016--2023). Its high-frequency \textbf{hourly} resolution allows us to test the model's sensitivity to rapid meteorological changes.
    \item \textbf{KnowAir (3-Hour Granularity):} A city level benchmark covering \textbf{184 cities} (2015--2018). With a \textbf{3-hour} resolution, it serves to validate the model's capability in capturing meso-scale pollution transmission patterns supported by meteorological dynamics.
\end{itemize}

\begin{table}[htbp]
    \centering
    \caption{\textbf{Dataset Features.} (Left) KnowAir features including meteorological node attributes and spatial edge attributes. (Right) LargeAQ dynamic meteorological variables.}
    \label{tab:dataset_features}
    \small
    \renewcommand{\arraystretch}{1.15}
    
    \begin{minipage}[t]{0.52\linewidth}
        \centering
        \textbf{(a) KnowAir Dataset} \\[0.5em]
        \begin{tabular}{@{}lcc@{}}
        \toprule
        \textbf{Variable} & \textbf{Symbol} & \textbf{Unit} \\
        \midrule
        \multicolumn{3}{l}{\textit{Node Attributes (Meteorological)}} \\
        Boundary Layer Height & $H_{PBL}$ & $m$ \\
        K Index & $KI$ & $K$ \\
        U-component of Wind & $u$ & $m/s$ \\
        V-component of Wind & $v$ & $m/s$ \\
        2m Temperature & $T_{2m}$ & $K$ \\
        Relative Humidity & $RH$ & $\%$ \\
        Total Precipitation & $TP$ & $m$ \\
        Surface Pressure & $P_{surf}$ & $Pa$ \\
        \midrule
        \multicolumn{3}{l}{\textit{Edge Attributes (Spatial)}} \\
        Wind Speed (Source) & $|v|$ & $km/h$ \\
        Distance (Src-Sink) & $d$ & $km$ \\
        Wind Direction (Source) & $\beta$ & $^\circ$ \\
        Direction (Src to Sink) & $\gamma$ & $^\circ$ \\
        Advection Coefficient & $S$ & $\%$ \\
        \bottomrule
        \end{tabular}
    \end{minipage}%
    \hfill
    \begin{minipage}[t]{0.45\linewidth}
        \centering
        \textbf{(b) LargeAQ Dataset} \\[0.5em]
        \begin{tabular}{@{}lcc@{}}
        \toprule
        \textbf{Variable} & \textbf{Symbol} & \textbf{Unit} \\
        \midrule
        2m Temperature & $T_{2m}$ & $^{\circ}$C \\
        Relative Humidity & $RH$ & $\%$ \\
        Total Precipitation & $TP$ & $mm$ \\
        U-component Wind & $u_{10}$ & $m/s$ \\
        V-component Wind & $v_{10}$ & $m/s$ \\
        Wind Speed & $WS$ & $m/s$ \\
        Wind Direction & $WD$ & $^\circ$ \\
        \bottomrule
        \end{tabular}
    \end{minipage}
\end{table}

\section{Static Geospatial Feature}
\label{app:static_features}

Table \ref{tab:static_features} details the 32 static geospatial features retrieved based on station coordinates.

\subsection{Rationale for Feature Selection}
\label{sub:feature_rationale}

The selection of these 32 geospatial features is grounded in the physicochemical mechanisms governing air quality, effectively capturing the duality of emission intensity and environmental capacity. Anthropogenic indicators, such as road network density and industrial facility counts, serve as direct proxies for primary emission sources, exhibiting strong correlations with traffic related NO$_2$/CO and combustion related SO$_2$. Complementarily, environmental context dictates pollutant fate: topographic factors (e.g., roughness, TPI) and meteorological history determine the physical dispersion or stagnation of particulates (PM$_{2.5}$, PM$_{10}$), while climatic variables (temperature, precipitation) and land use (vegetation) modulate the photochemical formation of secondary pollutants like O$_3$ and the removal processes via wet/dry deposition.
To capture the local geomorphological characteristics, we compute the Topographic Position Index (TPI) and Terrain Roughness based on the elevation data within a defined window.

\subsection{Contextual Neighborhood Aggregation}
\label{sub:context_agg}

To rigorously capture the local spatial heterogeneity invariant to global coordinates, we construct a geographic $K$-nearest neighbor set $\mathcal{N}_i$ and compute high order statistical features. We first calculate the first and second moments (mean $\mu_i^{nbr}$ and standard deviation $\sigma_i^{nbr}$) of the local concentration field. To further quantify the directional bias of the pollution distribution, we derive the pollution weighted centroid $\mathbf{p}_i^{*}$ and the station's geometric offset $\delta_i^{c}$ relative to this centroid:

\begin{equation}
    \mu_i^{nbr} = \frac{1}{|\mathcal{N}_i|} \sum_{j \in \mathcal{N}_i} c_j, \quad \mathbf{p}_i^{*} = \frac{\sum_{j \in \mathcal{N}_i} c_j \cdot \mathbf{p}_j}{\sum_{j \in \mathcal{N}_i} c_j}, \quad \delta_i^{c} = \|\mathbf{p}_i - \mathbf{p}_i^{*}\|_2
\end{equation}
where $c_j$ represents the historical mean concentration. Additionally, we define the local anomaly term $\Delta_i^{self} = c_i - \mu_i^{nbr}$ to explicitly differentiate between local emission sources (positive anomaly) and background regions (negative anomaly). For newly deployed stations without historical logs, neighborhood statistics are estimated using the nearest pre-existing anchors.

We concatenate these statistical descriptors into a unified context vector $\mathbf{f}_i^{nbr} = \left[\mu_i^{nbr}, \sigma_i^{nbr}, \delta_i^{c}, \Delta_i^{self}, \mathbf{p}_i^{level}\right]$, where $\mathbf{p}_i^{level}$ denotes the discrete distribution of pollution categories within the neighborhood. The final inductive identity embedding $\mathbf{e}_i^{ID}$ is generated by projecting the spectral position $\gamma(\mathbf{p}_i)$ jointly with these environmental priors and auxiliary geographic attributes $\mathbf{f}_i^{geo}$:
\begin{equation}
    \mathbf{e}_i^{ID} = \text{MLP}\left( \gamma(\mathbf{p}_i) \oplus \mathbf{f}_i^{nbr} \oplus \mathbf{f}_i^{geo} \oplus \psi(l_i) \right)
\end{equation}
Here, $\psi(l_i)$ is a learnable embedding of the station's own pollution grade. This formulation ensures that the learned representation is driven by observable environmental contexts rather than transductive indices, supporting direct inference on unseen nodes.

\subsection{Stratified Feature Encoding}
\label{sub:stratified_enc}

Acknowledging the inherent heterogeneity of global monitoring networks, we recognize that not all fine-grained attributes are uniformly available or equally predictive across diverse geographical regions. To address this, our framework adopts a stratified encoding strategy that prioritizes a universally accessible \textbf{core set} (e.g., coordinates, elevation) for baseline stability, while treating dense environmental metadata as a flexible auxiliary set. This decoupling supports \textbf{flexible spatial extrapolation}, allowing missing attributes in data sparse zones to be approximated via semantic neighbors. Crucially, the inductive nature of the ISIE module enables real-time supplementation of dynamic covariates—during inference without architectural modification. This design ensures the model effectively leverages rich metadata in data-rich environments while maintaining robust generalization in resource constrained areas. 

To ensure a rigorous benchmarking environment and maintain consistency with prior studies, we utilized the complete set of available variables as covariates for the LargeAQ and KnowAir datasets, strictly adhering to the graph construction and feature engineering paradigm proposed in PM2.5GNN. In contrast, for our developed WorldAir dataset, we specifically integrated a curated set of static geospatial features to explicitly model the influence of topography and long-term climatic patterns. These selected features include \texttt{elevation}, \texttt{climate\_avg\_wind}, \texttt{climate\_avg\_wind\_dir}, \texttt{terrain\_tpi}, \texttt{terrain\_roughness}, and \texttt{distance\_to\_coast\_km}, which collectively characterize the physical monitoring environment.

\begin{table}[htbp]
    \centering
    \caption{List of Collected Environmental Variables}
    \label{tab:static_features}
    \small
    \renewcommand{\arraystretch}{1.1}
    
    \begin{minipage}[t]{0.48\linewidth}
        \centering
        \begin{tabular}{@{}clcl@{}}
        \toprule
        \textbf{\#} & \textbf{Variable} & \textbf{Unit} & \textbf{Source} \\
        \midrule
        1 & Distance to coastline & km & Nat. Earth \\
        2 & Climate zone & - & Wiki \\
        3 & City name & - & Nominatim \\
        4 & Country code & - & Nominatim \\
        5 & Place type & - & Nominatim \\
        6 & Is urban area & Bool & Nominatim \\
        7 & Annual avg. temp. & $^{\circ}$C & Open-Meteo \\
        8 & Annual temp. range & $^{\circ}$C & Open-Meteo \\
        9 & Annual precip. & mm & Open-Meteo \\
        10 & Avg. wind speed & km/h & Open-Meteo \\
        11 & Dominant wind dir. & $^{\circ}$ & Open-Meteo \\
        12 & Elevation & m & Open-Meteo \\
        13 & Topographic Pos. Idx. & - & Open-Meteo \\
        14 & Terrain roughness & - & Open-Meteo \\
        15 & Building cnt. (200m) & Cnt & Overpass \\
        16 & Road cnt. (500m) & Cnt & Overpass \\
        \bottomrule
        \end{tabular}
    \end{minipage}%
    \hfill
    \begin{minipage}[t]{0.48\linewidth}
        \centering
        \begin{tabular}{@{}clcl@{}}
        \toprule
        \textbf{\#} & \textbf{Variable} & \textbf{Unit} & \textbf{Source} \\
        \midrule
        17 & Primary road cnt. & Cnt & Overpass \\
        18 & Industrial area cnt. & Cnt & Overpass \\
        19 & Factory cnt. (2km) & Cnt & Overpass \\
        20 & Power plant cnt. & Cnt & Overpass \\
        21 & Chimney cnt. (2km) & Cnt & Overpass \\
        22 & Forest cnt. (1km) & Cnt & Overpass \\
        23 & Park cnt. (500m) & Cnt & Overpass \\
        24 & Airport cnt. (5km) & Cnt & Overpass \\
        25 & Farmland cnt. (1km) & Cnt & Overpass \\
        26 & Primary land use & - & Overpass \\
        27 & Dist. to water body & km & Overpass \\
        28 & Nearest water type & - & Overpass \\
        29 & Nearest water name & - & Overpass \\
        30 & Dist. to major road & km & Overpass \\
        31 & Dist. to industrial & km & Overpass \\
        32 & Population density & /$km^2$ & WorldPop \\
        \bottomrule
        \end{tabular}
    \end{minipage}
\end{table}
\subsection{Calculation of Topographic Features}
\label{sub:calc_topo}

\paragraph{Topographic Position Index (TPI)}
TPI compares the elevation of a central point ($z_0$) to the mean elevation ($\bar{z}$) of its surrounding neighborhood. It identifies whether the location is a valley, ridge, or flat slope.
\begin{equation}
    TPI = z_0 - \bar{z} = z_0 - \frac{1}{n} \sum_{i \in W} z_i
\end{equation}
where $z_0$ is the elevation of the specific monitoring station, $W$ represents the neighborhood window, $n$ is the total number of pixels in the window, and $z_i$ is the elevation of the $i$-th pixel in the neighborhood.

\paragraph{Terrain Roughness}
Terrain roughness represents the variability of elevation in the local area. It is calculated as the standard deviation of elevation values within the neighborhood window:
\begin{equation}
    Roughness = \sqrt{\frac{1}{n} \sum_{i=1}^{n} (z_i - \bar{z})^2}
\end{equation}
Higher roughness values indicate complex terrain, which can obstruct wind flow and trap pollutants, while lower values indicate flat terrain favorable for dispersion.

\paragraph{Spatial Distance Calculation}
We employ the Haversine formula for great circle distance on Earth's surface:
\begin{equation}
    d = 2r \arcsin\left(\sqrt{h(\Delta\theta) + \cos\theta_1 \cos\theta_2 \cdot h(\Delta\varphi)}\right)
\end{equation}
where $h(x) = \sin^2(x/2)$, $r \approx 6371$ km is Earth's radius, $\theta$ and $\varphi$ denote latitude and longitude in radians, and $\Delta$ indicates the difference between two points.

\section{OmniAir Framework Implementation}
\label{app:framework_impl}

\begin{algorithm}[H]
\caption{Forward Propagation of OmniAir Framework}
\label{alg:OmniAir_forward}
\begin{algorithmic}[1]
\REQUIRE 
    Historical observations $\mathbf{X}_{hist} : \textcolor{magenta}{(B, T, N, C)}$, 
    Static attributes $\mathbf{f}^{geo} : \textcolor{magenta}{(N, D_{geo})}$, 
    Coordinates $\mathbf{p} : \textcolor{magenta}{(N, 2)}$
\ENSURE 
    Forecasted air quality $\hat{\mathbf{Y}} : \textcolor{magenta}{(B, \tau, N, C)}$

\vspace{0.5em}
\STATE \textit{\# --- Inductive Semantic Identity Encoder (ISIE) ---}
\STATE \textit{\# multi-scale fourier mapping for coordinates}
\STATE $\gamma(\mathbf{p}) : \textcolor{magenta}{(N, D_{pos})} \leftarrow \textbf{FourierMap}(\mathbf{p})$ 
\STATE \textit{\# aggregate neighborhood context (mean, std, centroid)}
\STATE $\mathbf{f}^{nbr} : \textcolor{magenta}{(N, D_{nbr})} \leftarrow \textbf{ContextAgg}(\mathbf{X}_{train}, \mathcal{N}_{geo})$
\STATE \textit{\# generate inductive identity embedding}
\STATE $\mathbf{E}^{ID} : \textcolor{magenta}{(N, D)} \leftarrow \textbf{MLP}_{ID}(\gamma(\mathbf{p}) \oplus \mathbf{f}^{nbr} \oplus \mathbf{f}^{geo})$

\vspace{0.5em}
\STATE \textit{\# --- Dynamic Sparse Topology Generator (DSTG) ---}
\STATE $\mathbf{H} : \textcolor{magenta}{(B, T, N, D)} \leftarrow \textbf{Linear}(\mathbf{X}_{hist})$
\STATE \textit{\# construct hybrid neighbors (geographic + semantic)}
\STATE $\mathcal{N}^{sem} \leftarrow \textbf{TopK}(-\|\mathbf{E}_i^{ID} - \mathbf{E}_j^{ID}\|_2)$
\STATE \textit{\# compute dynamic attention and fusion gate}
\STATE $\boldsymbol{\alpha} \leftarrow \tanh(\textbf{Attn}(\mathbf{H})), \quad \mathbf{g}_{edge} \leftarrow \sigma(\textbf{Gate}(\mathbf{H}, \mathbf{W}^{static}))$
\STATE \textit{\# adaptive pruning with learnable threshold}
\STATE $\boldsymbol{\beta} : \textcolor{magenta}{(B, N)} \leftarrow k_{max} \cdot \sigma(\textbf{MLP}_{\beta}(\mathbf{H}))$
\STATE $\mathbf{M} : \textcolor{magenta}{(B, N, N)} \leftarrow \sigma(-\eta \cdot (\textbf{Rank}(\boldsymbol{\alpha}) - \boldsymbol{\beta}))$
\STATE \textit{\# normalize sparse adjacency matrix}
\STATE $\tilde{\mathcal{A}} \leftarrow \textbf{Normalize}( ( \mathbf{g}_{edge} \mathbf{W}^{static} + (1-\mathbf{g}_{edge})\boldsymbol{\alpha} ) \odot \mathbf{M} )$

\vspace{0.5em}
\STATE \textit{\# --- Air Aware Differential Propagation (AAGD) ---}
\STATE $\mathbf{H}^{(0)} \leftarrow \mathbf{H}$
\FOR{$l = 1$ \textbf{to} $L$}
    \STATE \textit{\# multi-step diffusion with sparse graph}
    \STATE $\mathbf{H}^{(l)} : \textcolor{magenta}{(B, T, N, D)} \leftarrow \tilde{\mathcal{A}} \mathbf{H}^{(l-1)} + \lambda \mathbf{H}^{(0)}$
\ENDFOR
\STATE \textit{\# spectral aggregation with signed coefficients}
\STATE $\mathbf{Q}, \mathbf{K} \leftarrow \textbf{Linear}_Q([\mathbf{H}^{(0)}...\mathbf{H}^{(L)}]), \textbf{Linear}_K([\mathbf{H}^{(0)}...\mathbf{H}^{(L)}])$
\STATE $\mathbf{A}_{spec} : \textcolor{magenta}{(B, N, L)} \leftarrow \tanh(\mathbf{Q}\mathbf{K}^T / \sqrt{d}) \odot \mathbf{B}_{prior}$
\STATE $\mathbf{Z} : \textcolor{magenta}{(B, T, N, D)} \leftarrow \sum_{l=0}^{L} \mathbf{A}_{spec}[:, :, l] \cdot \mathbf{H}^{(l)}$

\vspace{0.5em}
\STATE \textit{\# --- Module 4: Latent Identity Augmented Inference ---}
\STATE \textit{\# fusion gate for static and dynamic features}
\STATE $\mathbf{g}_{id} : \textcolor{magenta}{(B, T, N, 1)} \leftarrow \sigma(\textbf{Linear}_g([\mathbf{Z} \| \mathbf{E}^{ID}]))$
\STATE $\hat{\mathbf{Z}} \leftarrow \mathbf{g}_{id} \odot \mathbf{Z} + (1-\mathbf{g}_{id}) \odot \mathbf{E}^{ID}$
\STATE \textit{\# final projection to output horizon}
\STATE $\hat{\mathbf{Y}} : \textcolor{magenta}{(B, \tau, N, C)} \leftarrow \textbf{MLP}_{out}(\hat{\mathbf{Z}})$

\STATE \textbf{Return} $\hat{\mathbf{Y}}$
\end{algorithmic}
\end{algorithm}

\section{Extended Analysis of Adaptive Sparse Topology}
\label{app:adaptive_topology_analysis}

In this section, we provide the detailed mathematical formulation of the differentiable edge pruning mechanism used in the Dynamic Sparse Topology Generator (DSTG) and analyze its training dynamics to validate the model's ability to handle global density heterogeneity.

\subsection{Mathematical Derivation of Differentiable Pruning}
\label{sub:math_pruning}

To address the extreme variation in station density (e.g., dense clusters in Europe vs. sparse coverage in Africa), we introduce a mechanism that learns a node specific effective neighborhood size. Unlike fixed $k$ approaches, we predict a continuous truncation threshold $\beta_i$ for each node based on its hidden state $\mathbf{h}_i$:
\begin{equation}
    \beta_i = k_{max} \cdot \text{sigmoid}\left( \text{MLP}_\beta(\mathbf{h}_i) \right)
\end{equation}
This scales the threshold to the continuous range $(0, k_{max})$. To make the top $k$ selection differentiable, we employ a relaxation of the ranking operation. For each edge $(i, j)$, we compute its rank $r_{ij}$ based on attention scores and generate a soft mask $m_{ij}$:
\begin{equation}
    m_{ij} = \text{sigmoid}\left( -\eta \cdot (r_{ij} - \beta_i) \right)
\end{equation}
Here, $\eta > 0$ is a temperature parameter controlling the steepness of the cut-off. The mask acts as a soft gate: when the rank $r_{ij} < \beta_i$, $m_{ij} \to 1$ (edge retained); otherwise $m_{ij} \to 0$ (edge pruned). 

Finally, the dynamic edge weights are re-normalized to ensure signal integrity during message passing:
\begin{equation}
    \tilde{w}_{ij} = \frac{w_{ij}^{dyn} \cdot m_{ij}}{\sum_{k \in \mathcal{N}_i} w_{ik}^{dyn} \cdot m_{ik} + \epsilon}
\end{equation}
This formulation allows gradients to flow back into the threshold predictor $\text{MLP}_\beta$, enabling the model to end-to-end optimize the graph structure.

\paragraph{Differentiable Pruning Mechanism.}
To address the non-differentiability of the discrete ranking operation, we employ a Continuous Relaxation strategy that approximates the hard top $k$ selection with a differentiable Soft Heaviside function. Specifically, we define the soft mask $m_{ij}$ using a tempered sigmoid function centered at the learned threshold $\beta_i$: $m_{ij} = \sigma\left( \eta \cdot (\beta_i - r_{ij}) \right)$, where $\eta$ controls the transition sharpness. By treating the rank $r_{ij}$ as a fixed input, this formulation creates a smooth decision boundary that allows the gradient from the task loss $\mathcal{L}$ to propagate back to the threshold predictor $\beta_i$ via the chain rule:
\begin{equation}
    \frac{\partial \mathcal{L}}{\partial \beta_i} = \sum_{j \in \mathcal{N}_i} \frac{\partial \mathcal{L}}{\partial \tilde{w}_{ij}} \cdot \tilde{w}_{ij} (1 - m_{ij}) \cdot \eta
\end{equation}
This non-zero gradient enables the model to end-to-end optimize the effective neighborhood size, dynamically balancing the trade-off between information gain and noise suppression.

\paragraph{Inductive Extrapolation Capability.}
Unlike transductive approaches that learn fixed embeddings for a closed set of nodes, our framework is inherently \textit{inductive}. The ISIE module functions as a parameter shared operator rather than a lookup table. Consequently, for any unseen station $v_{\text{new}}$ during inference, provided its spatial coordinates $\mathbf{x}_{\text{new}}$ and dynamic inputs, the model can instantaneously generate its semantic embedding $\mathbf{e}_{\text{new}} = \Phi(\mathbf{x}_{\text{new}})$. This allows the new station to seamlessly integrate into the AAGD message-passing mechanism using pre-trained weights, enabling zero-shot reasoning on expanded spatial graphs without retraining.

\newpage

\section{Global Spatio-Temporal Analysis}
\label{app:global_aqi_overview}

\begin{figure*}[h!]
    \centering
    \begin{subfigure}{0.32\linewidth}
        \includegraphics[width=\linewidth]{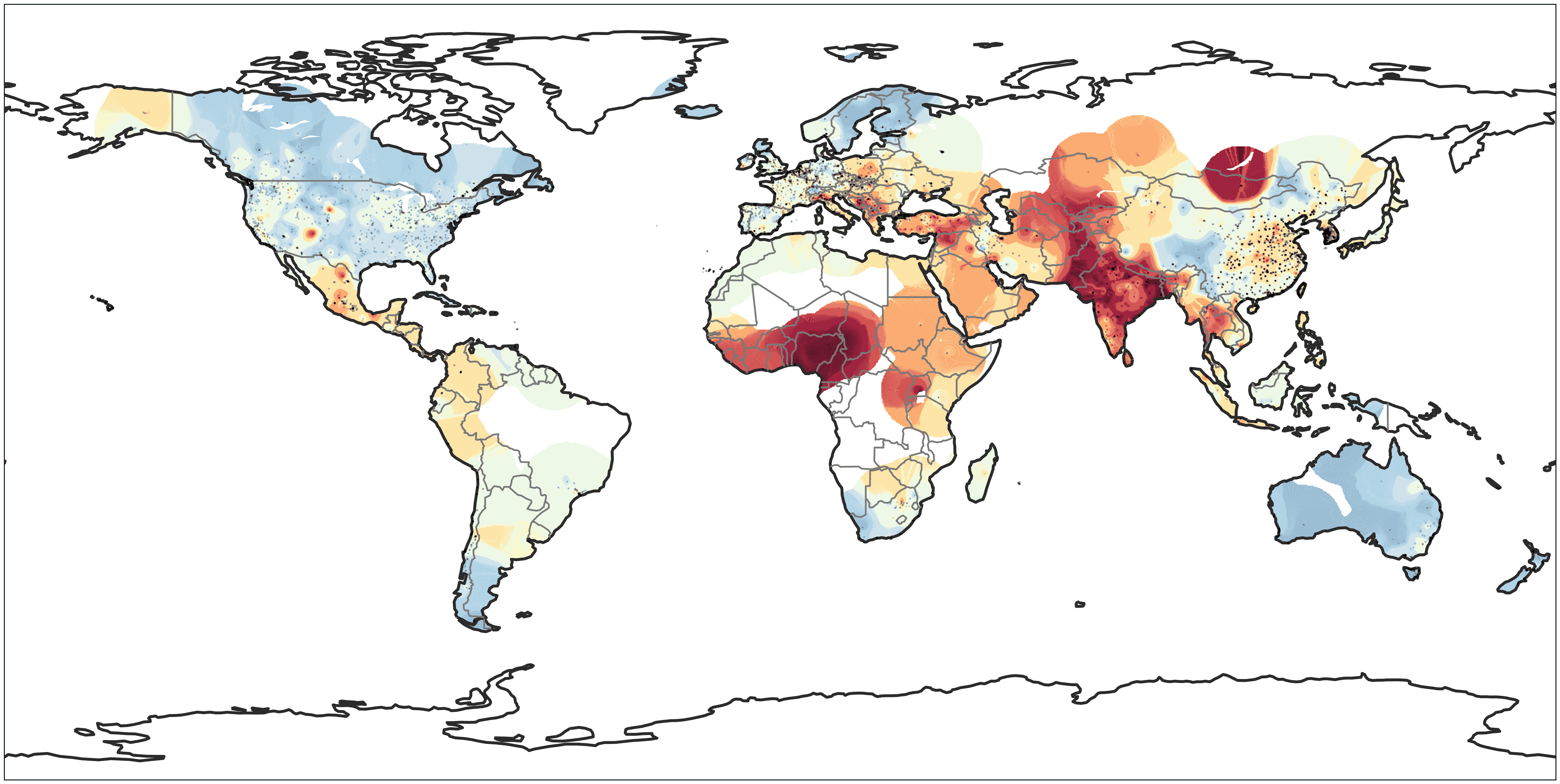}
        \caption{January}
    \end{subfigure}
    \hfill
    \begin{subfigure}{0.32\linewidth}
        \includegraphics[width=\linewidth]{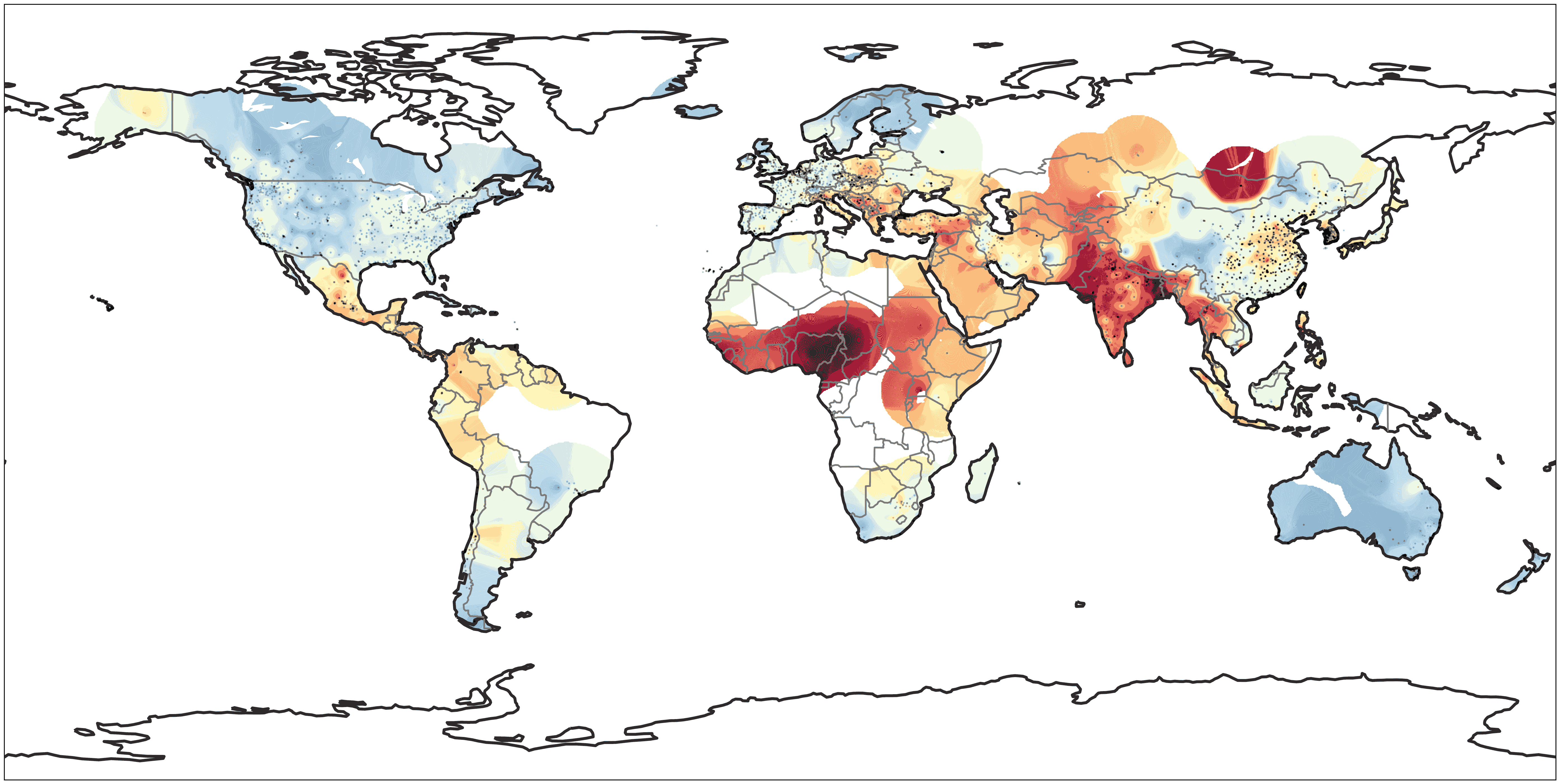}
        \caption{February}
    \end{subfigure}
    \hfill
    \begin{subfigure}{0.32\linewidth}
        \includegraphics[width=\linewidth]{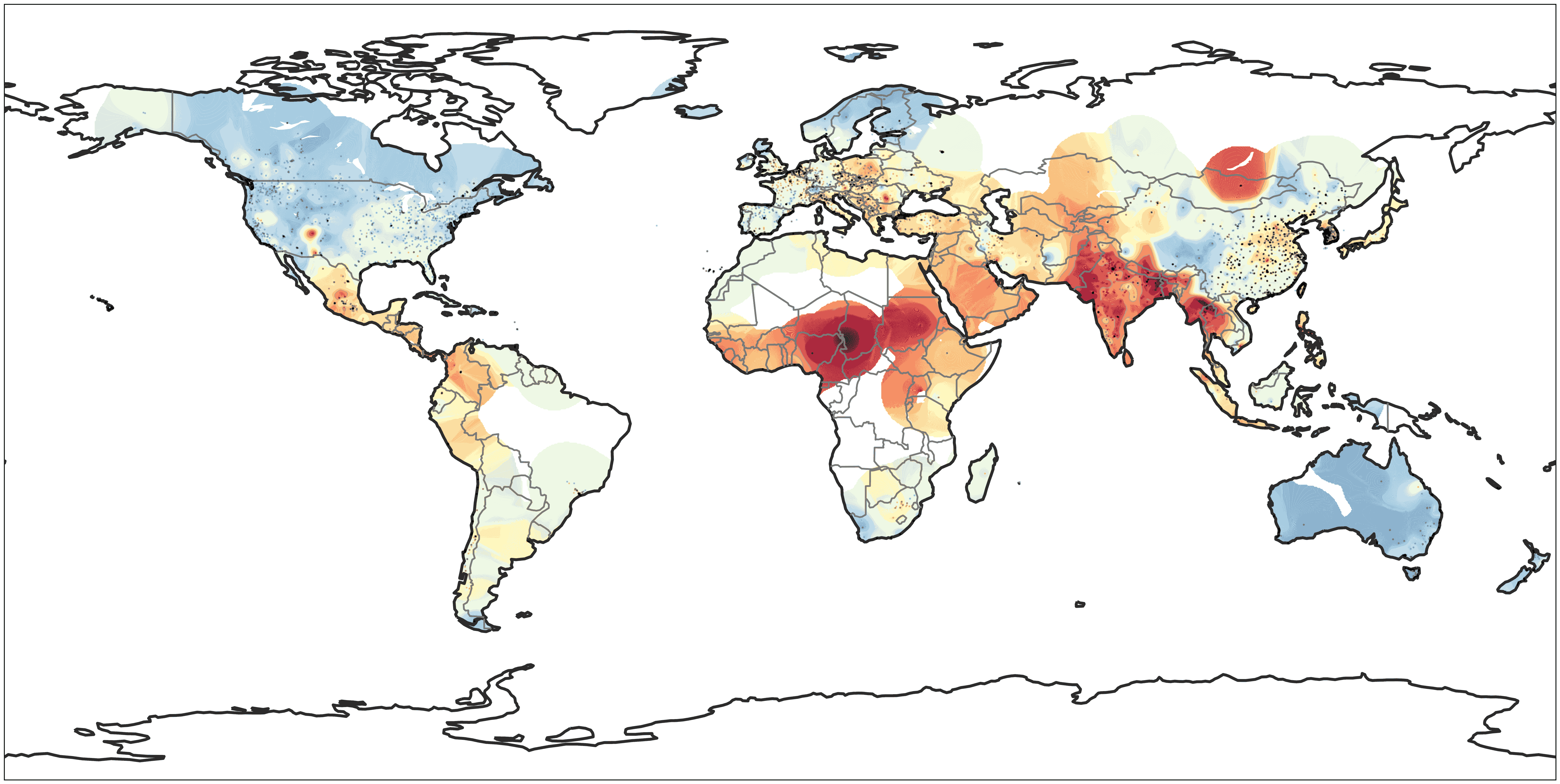}
        \caption{March}
    \end{subfigure}
    
    \vspace{0.5em}

    \begin{subfigure}{0.32\linewidth}
        \includegraphics[width=\linewidth]{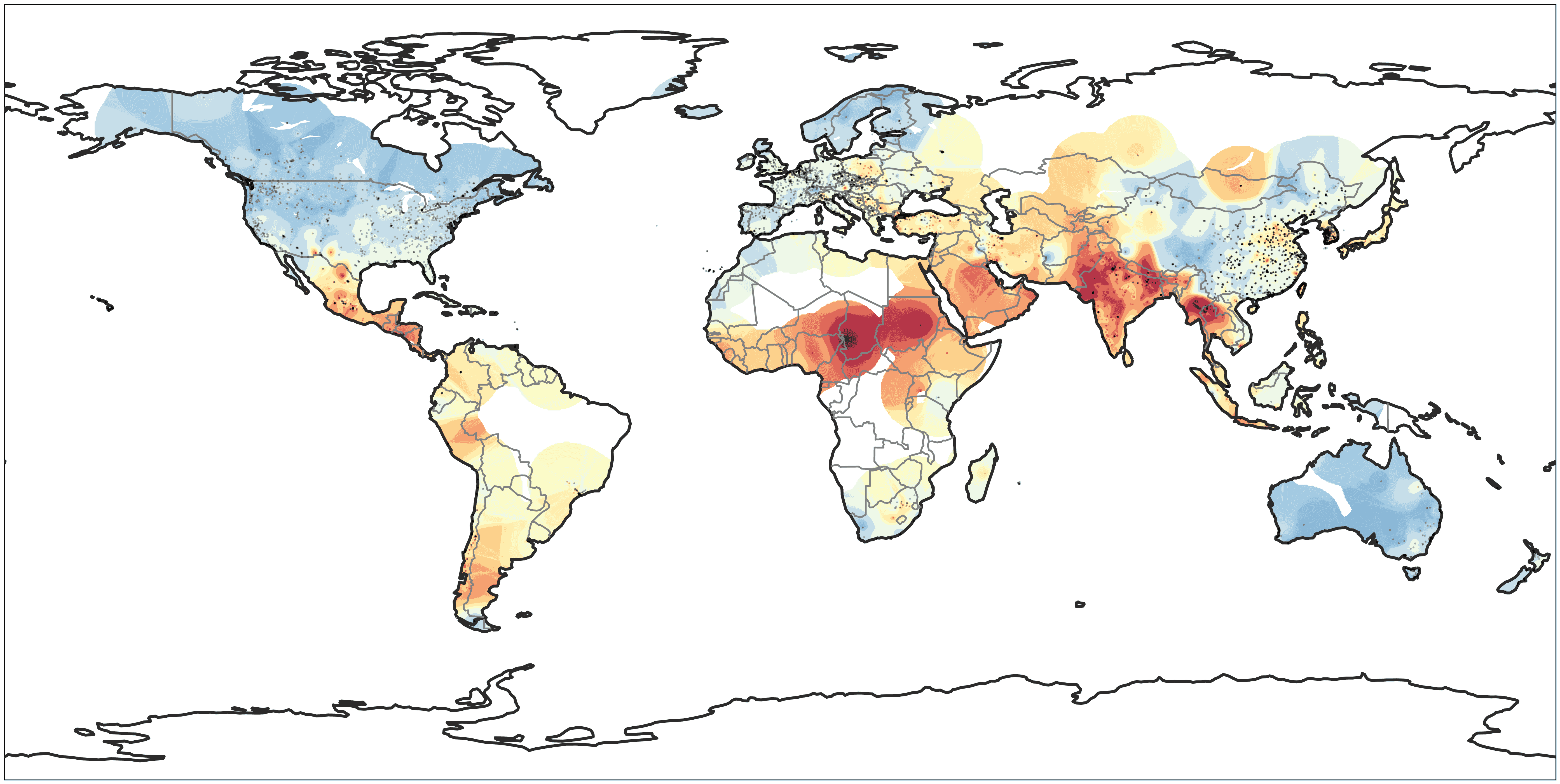}
        \caption{April}
    \end{subfigure}
    \hfill
    \begin{subfigure}{0.32\linewidth}
        \includegraphics[width=\linewidth]{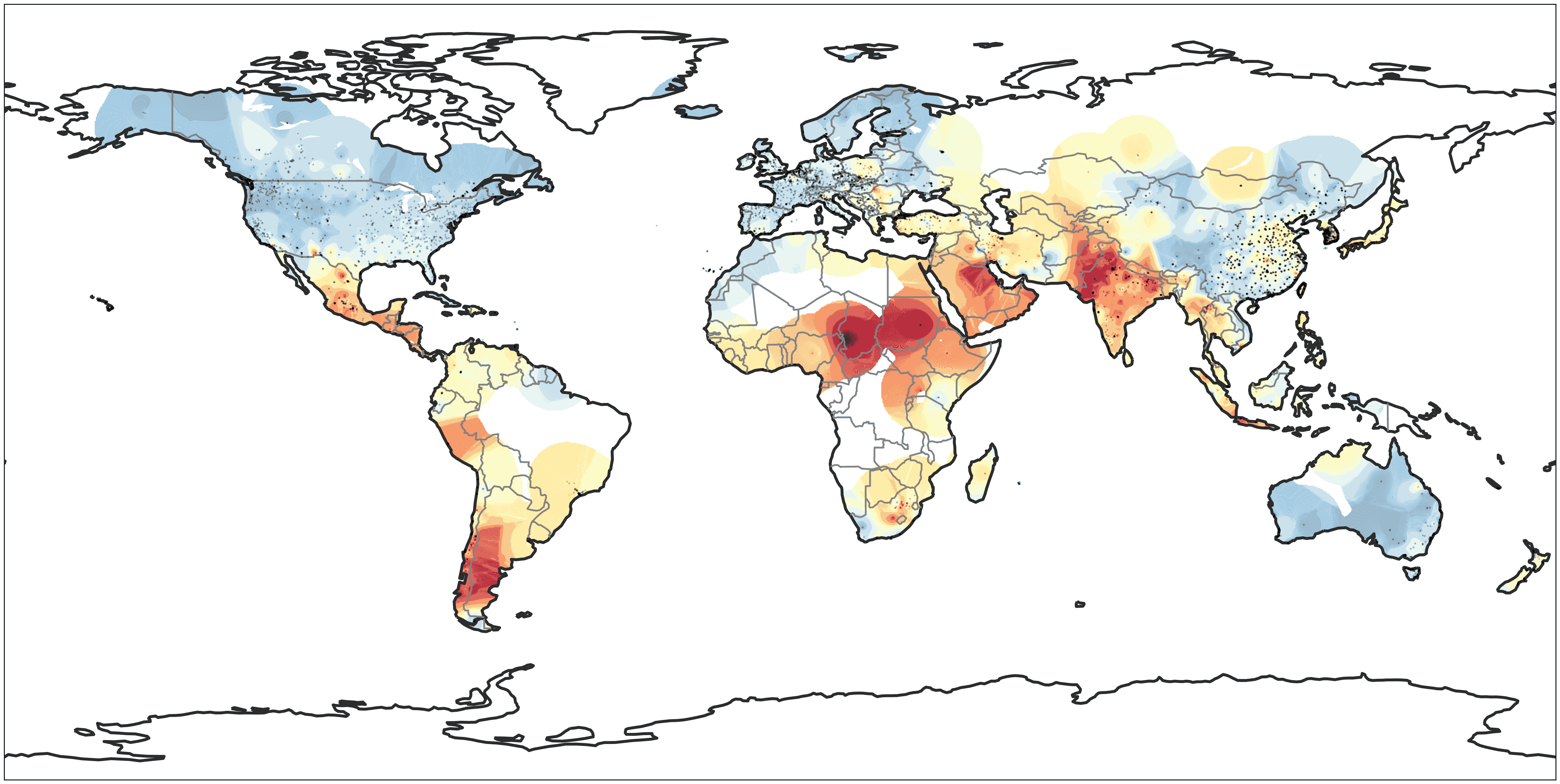}
        \caption{May}
    \end{subfigure}
    \hfill
    \begin{subfigure}{0.32\linewidth}
        \includegraphics[width=\linewidth]{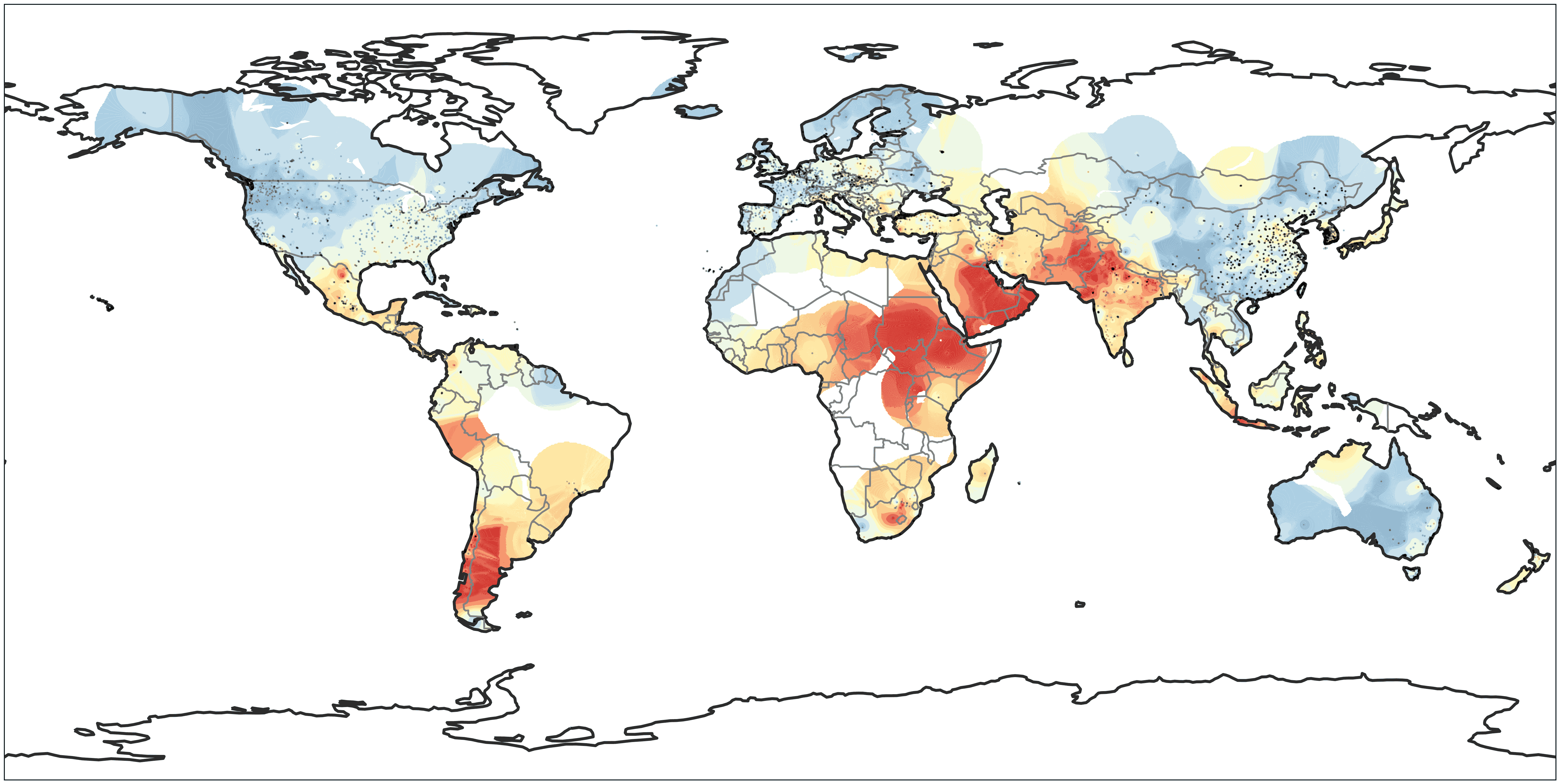}
        \caption{June}
    \end{subfigure}

    \vspace{0.5em}

    \begin{subfigure}{0.32\linewidth}
        \includegraphics[width=\linewidth]{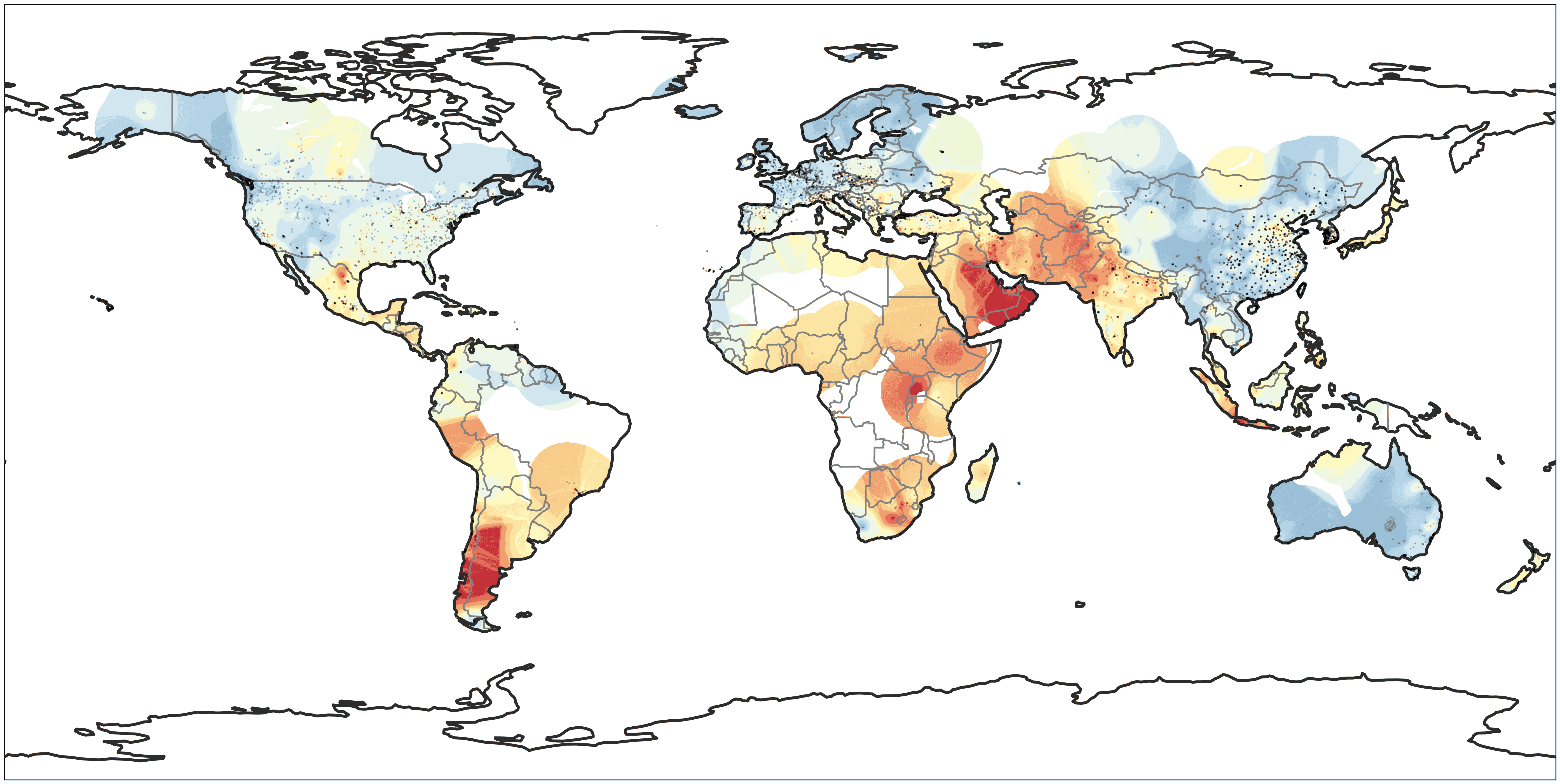}
        \caption{July}
    \end{subfigure}
    \hfill
    \begin{subfigure}{0.32\linewidth}
        \includegraphics[width=\linewidth]{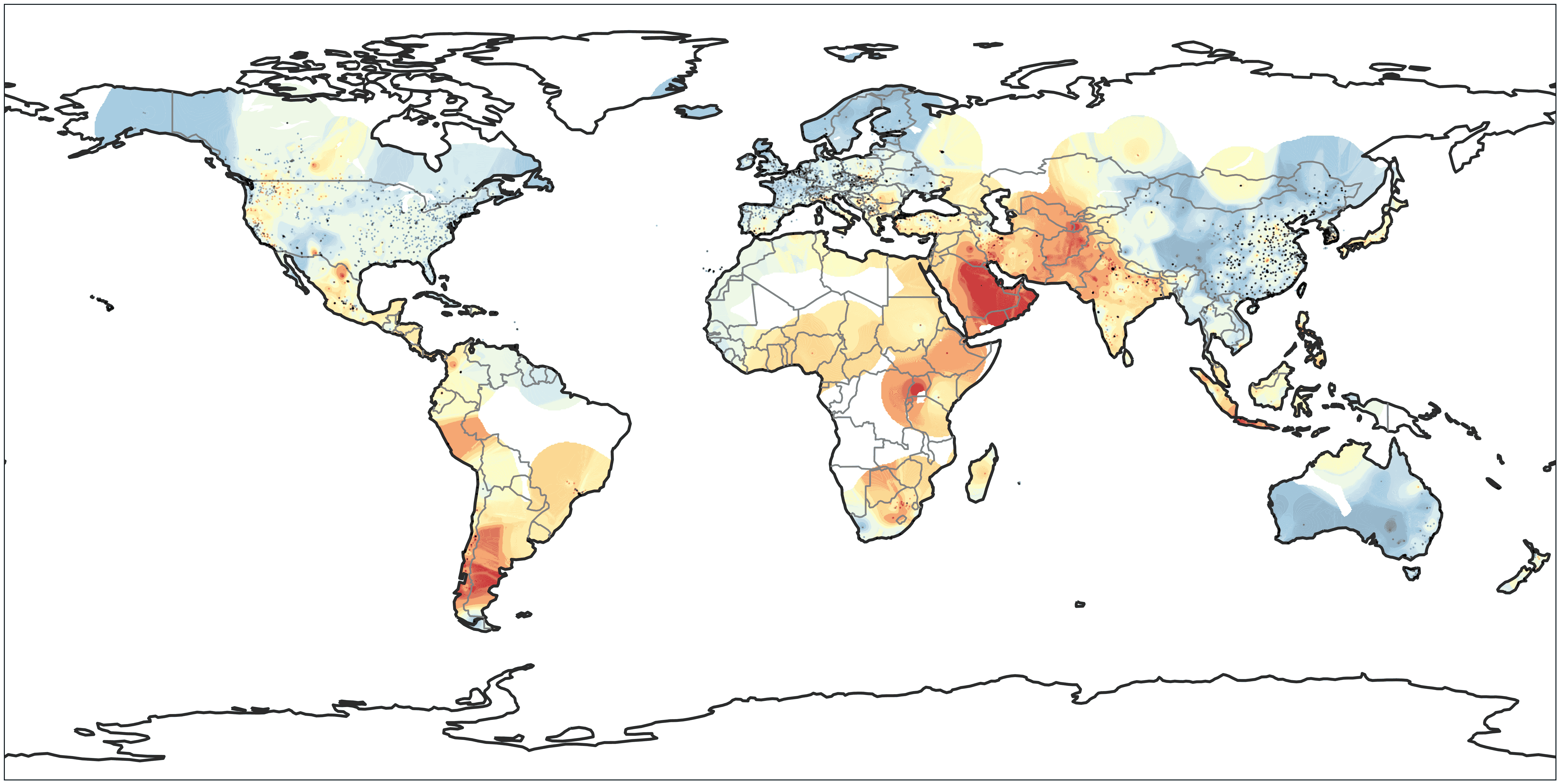}
        \caption{August}
    \end{subfigure}
    \hfill
    \begin{subfigure}{0.32\linewidth}
        \includegraphics[width=\linewidth]{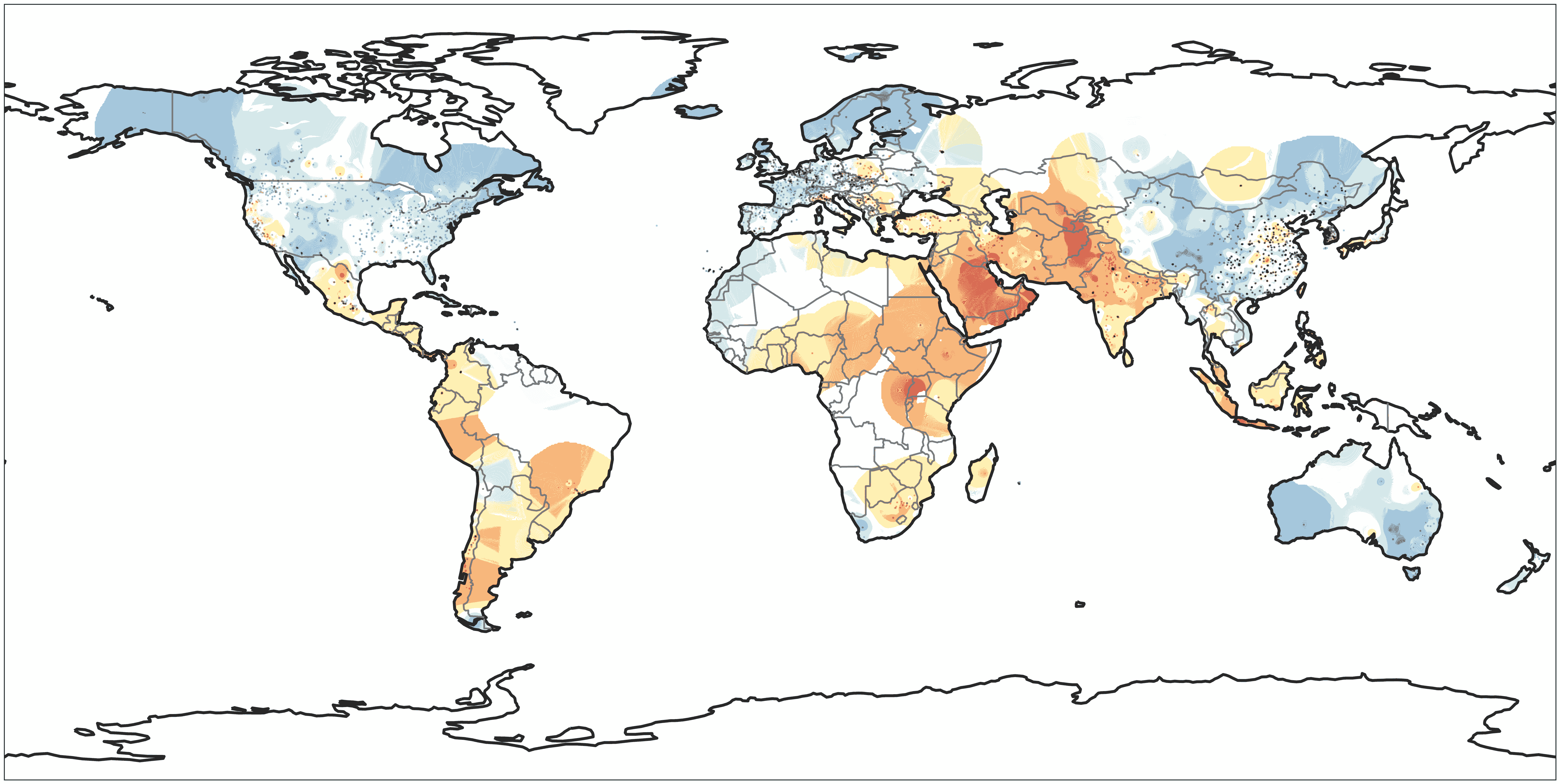}
        \caption{September}
    \end{subfigure}

    \vspace{0.5em}

    \begin{subfigure}{0.32\linewidth}
        \includegraphics[width=\linewidth]{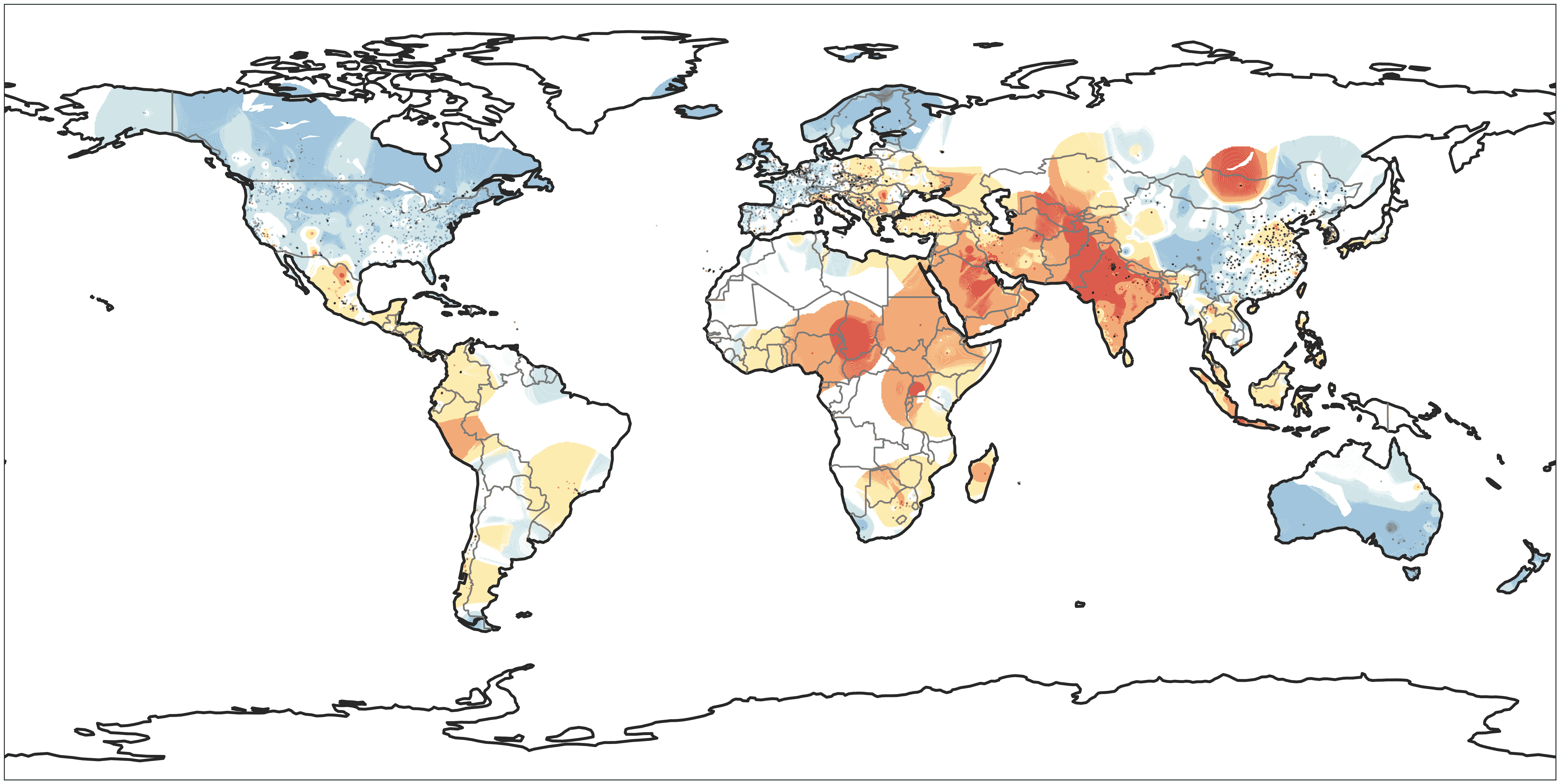}
        \caption{October}
    \end{subfigure}
    \hfill
    \begin{subfigure}{0.32\linewidth}
        \includegraphics[width=\linewidth]{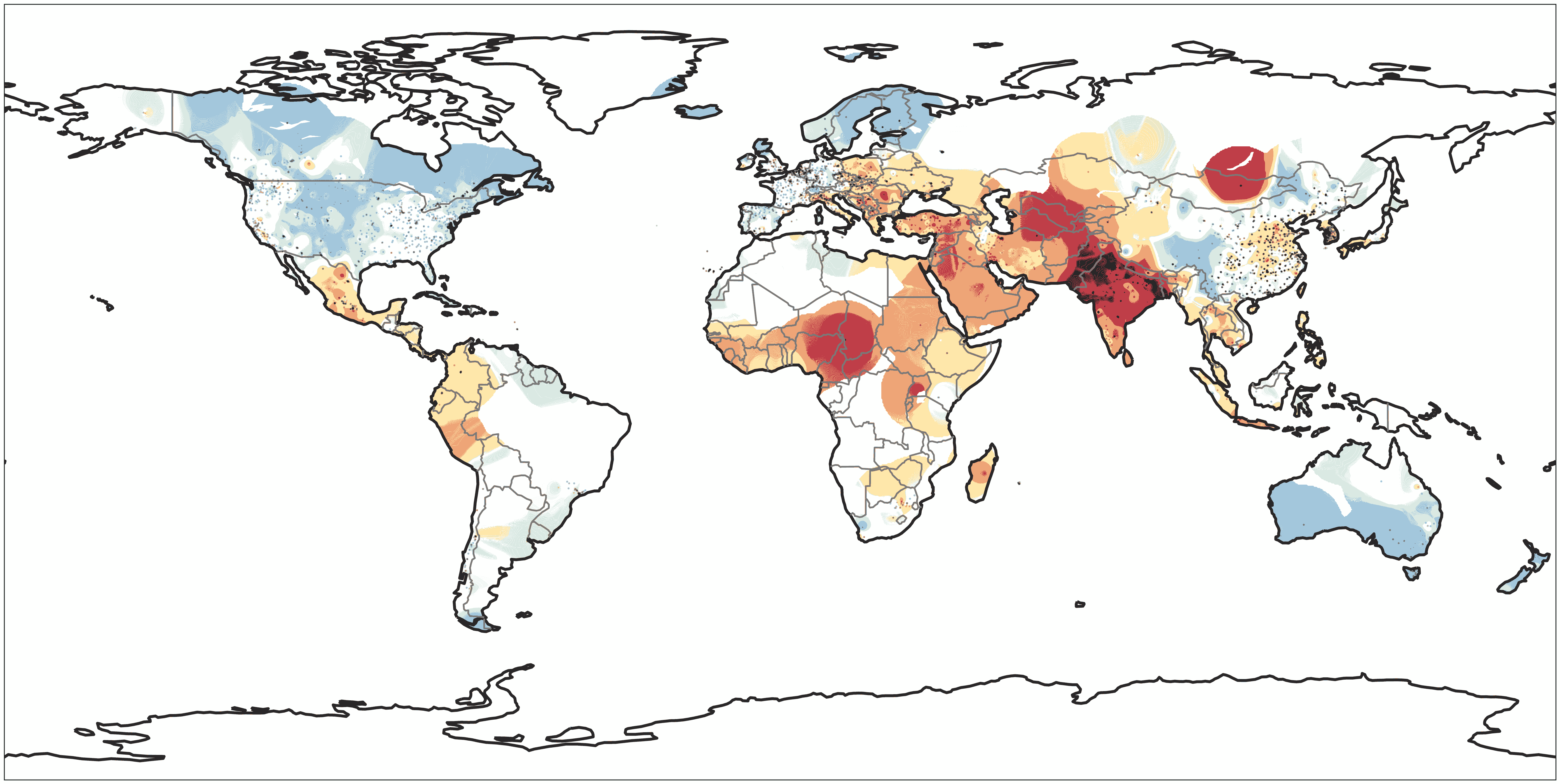}
        \caption{November}
    \end{subfigure}
    \hfill
    \begin{subfigure}{0.32\linewidth}
        \includegraphics[width=\linewidth]{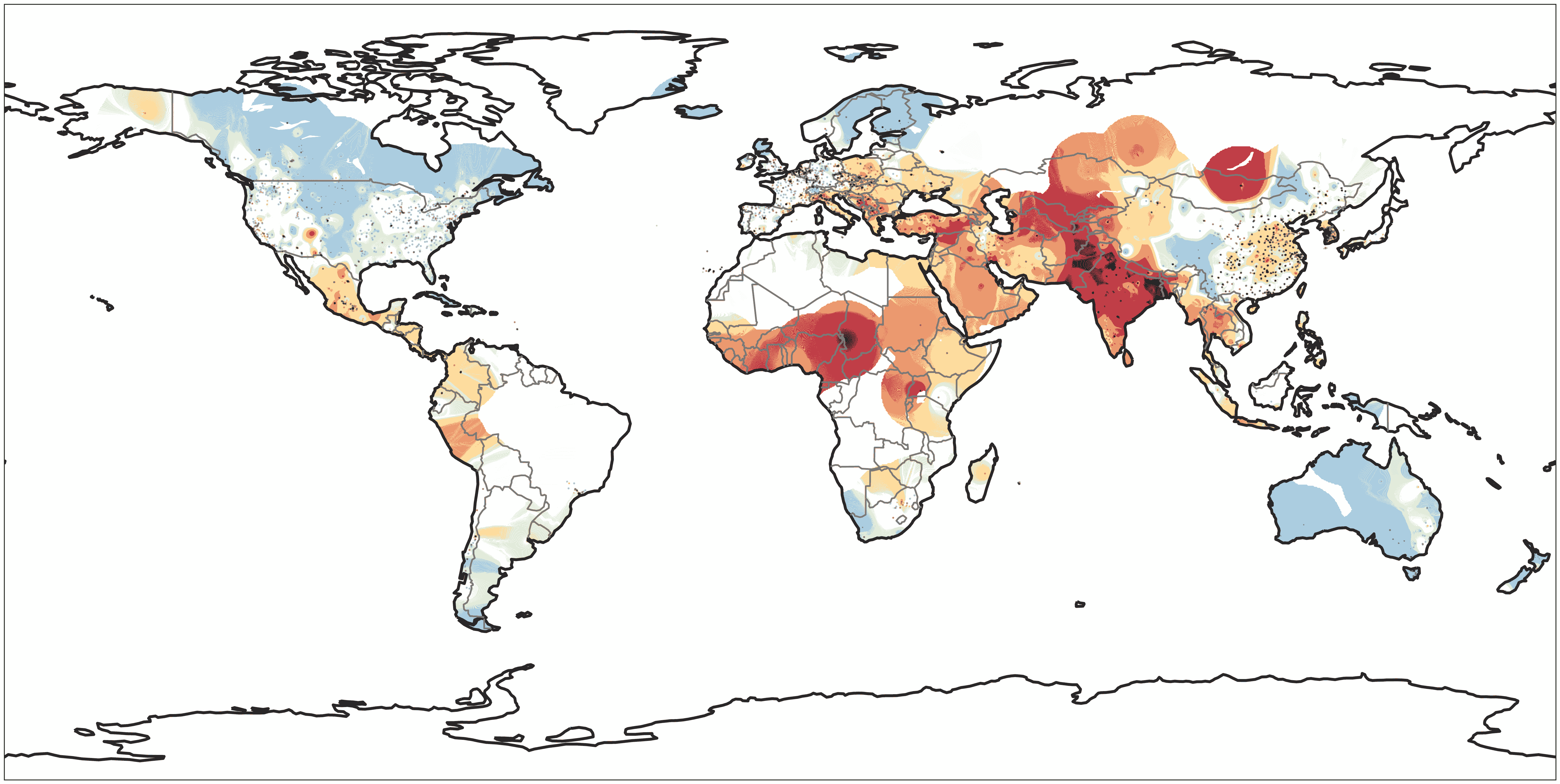}
        \caption{December}
    \end{subfigure}

    \caption{\textbf{Monthly global PM$_{2.5}$ distribution.} The maps show a clear seasonal shift in pollution: high concentrations appear in the Northern Hemisphere during winter (due to heating), while in warmer months, the hotspots move to desert and tropical regions (driven by dust and wildfires).}
    \label{fig:global_monthly_trend}
\end{figure*}

This section presents a comprehensive visualization of the monthly evolution of global PM$_{2.5}$ concentrations (Figure \ref{fig:global_monthly_trend}). From a global perspective, the distribution of air pollutants exhibits strong seasonality, driven by the complex coupling of meteorological dynamics and emission intensities.

\section{Implementation and Training Details}
\label{sec:implementation_details}

\subsection{Experimental Settings}
\label{subsec:experimental_settings}

Our experiments were conducted using the BasicTS framework~\cite{liang2022basicts}, a unified benchmark library for time series forecasting. The implementation was executed on a computational platform equipped with an NVIDIA A800 GPU (80GB memory) running PyTorch with Python 3.11. To ensure reproducibility, we fixed the random seed to 42 for all experimental runs. Regarding the specific architecture of OmniAir, we set the Fourier coordinate encoding dimension to 32 and the identity embedding dimension to 64. To capture station specific pollution characteristics, we employed a 16-dimensional embedding for PM$_{2.5}$ levels. The spatial-temporal backbone utilizes 4 attention heads and consists of 2 stacked layers of air-aware graph diffusion. To ensure a fair comparison across all benchmarks, the number of geographic and semantic neighbors in the topology generator is fixed to a uniform setting for all datasets. 

To standardize the inputs for various baselines, we applied a sliding window approach to generate samples. The input sequence length (look-back window) was set to $30$ time steps, and the forecasting horizon was set to $14$ time steps. 

\subsubsection{Graph Construction}
For Spatio-Temporal Graph Neural Networks (e.g., STGCN, GWNet, PM2.5-GNN), constructing an appropriate adjacency matrix is critical. We employed a distance based method to define the graph topology. Specifically, the adjacency matrix $A_{ij}$ between station $i$ and station $j$ was calculated using a Gaussian kernel with a thresholding strategy:
\begin{itemize}
    \item \textbf{Thresholding:} An edge is established only if the geographical distance between two stations is less than $300$ km.
    \item \textbf{Weighting:} For connected nodes, the edge weight is computed as $w_{ij} = \exp(-\frac{dist(i,j)^2}{\sigma^2})$, where the standard deviation $\sigma$ is set to $100$.
\end{itemize}
This configuration ensures that the graph captures meaningful local geographic dependencies while ignoring negligible long-range connections.

\subsubsection{Training Protocol}
To ensure a fair comparison, all models were optimized using the Adam optimizer to minimize the Mean Absolute Error (MAE). We maintained consistent hyperparameter configurations across all baselines, fixing the batch size at $32$ and the initial learning rate at $1 \times 10^{-3}$. Regularization was implemented via a weight decay of $1 \times 10^{-5}$ to mitigate overfitting. The training process was capped at a maximum of $300$ epochs, incorporating an early stopping mechanism with a patience of $20$ epochs; specifically, training was terminated if the validation loss failed to improve for 20 consecutive epochs, at which point the optimal model weights were restored.
\subsection{Evaluation Metrics}
\label{sec:metrics}

To comprehensively assess the predictive performance of our model across different forecasting horizons, we employ three standard evaluation metrics: Mean Absolute Error (MAE), Root Mean Square Error (RMSE), and Mean Absolute Percentage Error (MAPE). 

Let $Y = \{y_1, y_2, \dots, y_N\}$ denote the ground truth values of the air pollutant concentrations, and $\hat{Y} = \{\hat{y}_1, \hat{y}_2, \dots, \hat{y}_N\}$ represent the corresponding predicted values generated by the model, where $N$ is the total number of test samples across all monitoring stations and time steps.

To ensure robust evaluation under irregular sensor sampling, we implement a masked MAPE metric. A composite validity mask $\mathcal{M} = \mathbf{M}_{null} \odot \mathbb{I}(|\mathbf{Y}| > 5\times 10^{-5})$ is applied to simultaneously exclude missing observations and near zero background noise, preventing numerical divergence. The final metric is normalized by the effective data density $\mathbb{E}[\mathcal{M}]$, eliminating bias arising from the heterogeneous sparsity characteristic of global air quality monitoring.

\paragraph{Mean Absolute Error (MAE)}
MAE measures the average magnitude of the errors in a set of predictions, without considering their direction. It represents the direct difference between the predicted and observed air quality levels.
\begin{equation}
    \text{MAE} = \frac{1}{N} \sum_{i=1}^{N} |y_i - \hat{y}_i|
\end{equation}

\paragraph{Root Mean Square Error (RMSE)}
RMSE is the square root of the average of squared differences between prediction and actual observation. Since the errors are squared before they are averaged, RMSE gives a relatively high weight to large errors. This makes it particularly useful for evaluating whether the model can avoid severe prediction failures (e.g., missing extreme pollution events).
\begin{equation}
    \text{RMSE} = \sqrt{\frac{1}{N} \sum_{i=1}^{N} (y_i - \hat{y}_i)^2}
\end{equation}

\paragraph{Mean Absolute Percentage Error (MAPE)}
MAPE expresses the forecast error as a percentage of the actual value. Unlike MAE and RMSE which are scale-dependent, MAPE provides a relative measure of accuracy, allowing for performance comparison across different pollutant types with varying concentration scales.
\begin{equation}
    \text{MAPE} = \frac{1}{N} \sum_{i=1}^{N} \left| \frac{y_i - \hat{y}_i}{y_i} \right| \times 100\%
\end{equation}

Lower values for all three metrics indicate better model performance.

\section{Detailed Experimental Results}
\label{sec:experimental_results}

We conducted extensive comparative experiments on two real-world datasets, KnowAir and LargeAQ, to evaluate the forecasting performance of our proposed model against seven state-of-the-art baselines. The results are reported in terms of MAE, RMSE, and MAPE across different forecasting horizons.

\subsection{Performance on KnowAir Dataset}
\label{sub:exp_knowair}

Table \ref{tab:knowair_res} summarizes the performance comparison on the KnowAir dataset, which features a 3-hour temporal granularity.

\begin{table*}[htbp!]
    \centering
    \renewcommand{\arraystretch}{1.1} 
    \caption{Performance comparison on KnowAir. \best{Best} and \secondbest{second best} results are highlighted.}
    \label{tab:knowair_res} 
    \resizebox{\textwidth}{!}{%
        \begin{tabular}{lcccccccccccc}
            \toprule
            \multirow{2}{*}{\textbf{Model}} & \multicolumn{3}{c}{\textbf{18h}} & \multicolumn{3}{c}{\textbf{36h}} & \multicolumn{3}{c}{\textbf{72h}} & \multicolumn{3}{c}{\textbf{AVG}} \\
            \cmidrule(lr){2-4} \cmidrule(lr){5-7} \cmidrule(lr){8-10} \cmidrule(lr){11-13}
             & MAE & RMSE & MAPE & MAE & RMSE & MAPE & MAE & RMSE & MAPE & MAE & RMSE & MAPE \\
            \midrule
            
            STGCN & 8.25 & 12.69 & 46.21\% & 9.18 & 13.82 & 54.02\% & 10.11 & 15.56 & 61.35\% & 9.18 & 14.02 & 53.86\% \\
            STID & 8.14 & \secondbest{12.65} & 48.42\% & 9.35 & 14.12 & 58.45\% & 10.49 & 15.42 & 70.08\% & 9.33 & 14.06 & 58.98\% \\
            AGCRN & 8.41 & 13.08 & 48.64\% & 9.58 & 14.43 & 56.82\% & 10.68 & 15.85 & 68.16\% & 9.56 & 14.45 & 57.87\% \\ 
            GWNet  & 9.38 & 13.98 & 56.68\% & 10.62 & 15.35 & 64.85\% & 11.08 & 15.92 & 68.82\% & 10.36 & 15.08 & 63.45\% \\
            PM2.5GNN &\secondbest{8.12} & 12.81 & 46.75\% & \secondbest{9.12} & 14.08 & \secondbest{52.19\%} & \secondbest{9.94} & 15.25 & 63.81\% & \secondbest{9.06} & 14.05 & 54.25\% \\
            GAGNN & 10.15 & 15.24 & 59.26\% & 11.91 & 17.35 & 69.85\% & 12.19 & 19.85 & 73.35\% & 11.25 & 16.81 & 67.49\% \\
            AirFormer & 8.26 & 12.68 & \secondbest{45.49\%} & 9.21 & \secondbest{13.75} & 52.28\% & 10.12 & \secondbest{14.86} & \secondbest{61.12\%} & 9.20 & \secondbest{13.76} & \secondbest{52.96\%} \\
            AirPhyNet & 9.28 & 14.16 & 57.12\% & 10.65 & 15.86 & 67.05\% & 11.62 & 17.21 & 74.65\% & 10.52 & 15.74 & 66.27\% \\
            
            \midrule
            \textbf{Ours} & \best{7.58} & \best{11.85} & \best{42.52\%} & \best{8.51} & \best{12.86} & \best{48.45\%} & \best{9.19} & \best{13.59} & \best{53.95\%} & \best{8.43} & \best{12.77} & \best{48.31\%} \\
            
            \bottomrule
        \end{tabular}%
    }
\end{table*}

\subsection{Performance on LargeAQ Dataset}
\label{sub:exp_largeaq}

Table \ref{tab:largeaq_res} presents the results on the LargeAQ dataset, which is more challenging due to its finer 1-hour granularity and wider spatial coverage.

\begin{table*}[htbp!]
    \centering
    \renewcommand{\arraystretch}{1.1} 
    \caption{Performance comparison on LargeAQ. \best{Best} and \secondbest{second best} results are highlighted.}
    \label{tab:largeaq_res}
    \resizebox{\textwidth}{!}{%
        \begin{tabular}{lcccccccccccc}
            \toprule
            \multirow{2}{*}{\textbf{Model}} & \multicolumn{3}{c}{\textbf{6h}} & \multicolumn{3}{c}{\textbf{12h}} & \multicolumn{3}{c}{\textbf{24h}} & \multicolumn{3}{c}{\textbf{AVG}} \\
            \cmidrule(lr){2-4} \cmidrule(lr){5-7} \cmidrule(lr){8-10} \cmidrule(lr){11-13}
             & MAE & RMSE & MAPE & MAE & RMSE & MAPE & MAE & RMSE & MAPE & MAE & RMSE & MAPE \\
            \midrule
            
            STGCN & 14.85 & 23.12 & 51.95\% & 17.48 & 26.72 & 63.65\% & 19.05 & 28.62 & 73.28\% & 17.13 & 26.15 & 62.96\% \\
            STID & 15.79 & 24.75 & 58.12\% & 18.05 & 27.55 & 69.15\% & 19.12 & 28.85 & 75.82\% & 17.65 & 27.05 & 67.70\% \\
            AGCRN & 15.75 & 24.62 & 56.45\% & 17.95 & 27.38 & 67.12\% & 18.95 & 28.62 & 72.88\% & 17.55 & 26.87 & 65.48\% \\
            GWNet & 15.48 & 23.75 & 57.15\% & 18.18 & 27.42 & 70.92\% & 19.62 & 29.15 & 79.12\% & 17.76 & 26.77 & 69.06\% \\
            AirFormer & 15.52 & 23.15 & 56.35\% & \secondbest{16.48} & \secondbest{24.32} & \secondbest{60.05\%} & \secondbest{16.82} & \secondbest{24.75} & \secondbest{61.68\%} & \secondbest{16.27} & \secondbest{24.07} & \secondbest{59.36\%} \\
            AirPhyNet & 17.02 & 26.28 & 62.08\% & 19.28 & 29.12 & 71.35\% & 20.45 & 30.75 & 75.42\% & 18.92 & 28.72 & 69.62\% \\
            GAGNN & 15.02 & 30.85 & 53.52\% & 20.28 & 37.15 & 75.02\% & 31.18 & 49.65 & 96.05\% & 22.16 & 39.22 & 74.86\% \\
            PM2.5GNN & \secondbest{14.62} & \secondbest{22.88} & \secondbest{50.89\%} & 17.05 & 26.18 & 60.65\% & 18.92 & 28.18 & 73.35\% & 16.86 & 25.75 & 61.63\% \\
            \midrule
            \textbf{Ours} & \best{13.42} & \best{21.22} & \best{42.75\%} & \best{14.15} & \best{22.18} & \best{45.25\%} & \best{14.86} & \best{22.88} & \best{50.58\%} & \best{14.14} & \best{22.09} & \best{46.19\%} \\
            
            \bottomrule
        \end{tabular}%
    }
\end{table*}

\subsection{Ablation Study}
\label{ablation}

\begin{table*}[htbp]
    \centering
    \renewcommand{\arraystretch}{1.1} 
    \caption{Ablation study of different components in OmniAir framework.}
    \label{tab:ablation_study}
    
    \resizebox{\textwidth}{!}{%
        \begin{tabular}{lcccccccccccc}
            \toprule
            \multirow{2}{*}{\textbf{Variant}} & \multicolumn{3}{c}{\textbf{China}} & \multicolumn{3}{c}{\textbf{Europe}} & \multicolumn{3}{c}{\textbf{USA}} & \multicolumn{3}{c}{\textbf{Global}} \\
            \cmidrule(lr){2-4} \cmidrule(lr){5-7} \cmidrule(lr){8-10} \cmidrule(lr){11-13}
             & MAE & RMSE & MAPE & MAE & RMSE & MAPE & MAE & RMSE & MAPE & MAE & RMSE & MAPE \\
            \midrule
            
            \textit{w/o Ext. Feat.} & 7.81 & 15.02 & 24.35\% & 11.89 & 20.05 & 41.28\% & 10.52 & 18.56 & 41.52\% & 11.56 & 20.89 & 34.42\% \\
            \textit{w/o Sem. Graph} & 7.88 & 15.18 & 24.68\% & 12.05 & 20.38 & 41.95\% & 10.68 & 18.82 & 42.15\% & 11.82 & 21.35 & 35.18\% \\
            \textit{w/o Adap. Sparse} & 7.96 & 15.35 & 24.89\% & 12.18 & 20.62 & 42.35\% & 10.79 & 18.95 & 42.56\% & 11.98 & 21.67 & 35.72\% \\
            \textit{w/o ISIE} & 8.05 & 15.52 & 25.18\% & 12.35 & 20.92 & 42.88\% & 10.95 & 19.18 & 43.12\% & 12.18 & 22.05 & 36.25\% \\
            \textit{w/o Dyn. Graph} & 8.28 & 15.95 & 25.85\% & 12.72 & 21.58 & 43.82\% & 11.22 & 19.75 & 44.05\% & 12.58 & 22.78 & 37.52\% \\
            
            \midrule
            \textbf{OmniAir (Full)} & \best{7.72} & \best{14.83} & \best{24.01\%} & \best{11.70} & \best{19.73} & \best{40.71\%} & \best{10.38} & \best{18.38} & \best{41.04\%} & \best{11.28} & \best{20.41} & \best{33.69\%} \\
            \bottomrule
        \end{tabular}%
    }
\end{table*}

To strictly verify the effectiveness and necessity of each component in our proposed framework, we conducted a comprehensive ablation study on the Global dataset and three regional subsets. We designed five specific variants by systematically removing or replacing key modules, including the external environmental features, the semantic graph structure, the adaptive sparsity mechanism, the ISIE module, and the dynamic graph update strategy. The comparative results are summarized in Table \ref{tab:ablation_study}. As observed, the full OmniAir model consistently outperforms all ablated variants across all metrics and regions, demonstrating that these components work synergistically to capture the complex spatio-temporal heterogeneity of global air quality. Detailed analyses of specific modules are provided below.
        
\textbf{Impact of Environmental Context.} To assess the contribution of multi-source static features, we develop the variant \textit{w/o Ext. Feat.}, which removes auxiliary geographical attributes including elevation, wind speed, terrain, and distance to the coastline. This variant relies solely on station coordinates and historical neighborhood statistics. The observed performance degradation highlights the necessity of incorporating rich environmental contexts, as these factors provide essential physical priors that guide the model in distinguishing station specific patterns.

\textbf{Effectiveness of Semantic Topology.} We evaluate the role of the ISIE module and semantic graph construction through two variants. The \textit{w/o ISIE} variant removes the entire Inductive Semantic Identity Encoder module, stripping away the physical priors and degrading the model to a standard transductive setting. Similarly, \textit{w/o Sem. Graph} retains the module but excludes semantic neighbors derived from embedding distances, restricting the graph structure to geographical connections. The inferior performance of these variants demonstrates that physical semantic associations are crucial for capturing non-Euclidean correlations between distant stations that share similar pollution patterns.

\textbf{Dynamic Graph Mechanisms.} To verify the importance of capturing evolving spatial dependencies, we analyze the \textit{w/o Dyn. Graph} and \textit{w/o Adap. Sparse} variants. The former disables the dynamic graph update mechanism, preventing the model from adjusting edge weights based on time-variant features, while the latter removes the adaptive edge learning strategy, utilizing a fixed number of edges without dynamic pruning. The superior performance of the full model confirms that dynamic and adaptive graph structures are essential for handling the rapid evolution of air pollution patterns driven by changing meteorological conditions.

\subsection{Statistical Significance Test}
\label{sec:statistics}
To verify that the performance improvements of OmniAir are robust and not artifacts of random variance, we conducted paired t-tests ($N=5$) on the WorldAir dataset against the top-5 competitive baselines. The analysis confirms that OmniAir achieves statistically significant superiority over all competitors, with p-values consistently well below the 0.05 threshold. Specifically, our model maintains a significant lead ($p < 0.01$) even against the strongest baseline and demonstrates highly significant improvements ($p < 0.001$) compared to domain specific models like AirFormer and AirDualODE, validating the stability of our architecture.

\begin{table}[h]
    \centering
    \caption{Statistical significance analysis (Paired t-test, $N=5$) of OmniAir vs.}
    \label{tab:significance}
    \resizebox{0.45\linewidth}{!}{%
    \begin{tabular}{l|cc}
        \toprule
        \textbf{Comparison (vs. OmniAir)} & \textbf{t-statistic} & \textbf{p-value} \\
        \midrule
        vs. STID (Best Baseline) & -7.23 & $0.0019^{**}$ \\
        vs. Informer & -7.25 & $0.0019^{**}$ \\
        vs. AirFormer & -19.79 & $< 0.001^{***}$ \\
        vs. LightTS & -11.58 & $< 0.001^{***}$ \\
        vs. AirDualODE & -15.96 & $< 0.001^{***}$ \\
        \bottomrule
        \multicolumn{3}{l}{\small \textit{Note: $^{*}$ $p < 0.05$, $^{**}$ $p < 0.01$, $^{***}$ $p < 0.001$.}} \\
    \end{tabular}%
    }
\end{table}

\subsection{Hyperparameter Sensitivity Analysis}
\label{sec:hyperparameter}

To comprehensively evaluate the robustness of OmniAir and identify the optimal configuration for global scale forecasting, we conducted a systematic sensitivity analysis on four pivotal hyperparameters: the hidden dimension of the prediction head ($d_{pred}$), the frequency dimension of the Fourier positional encoding ($d_{fourier}$), and the sparsity thresholds for both geographic ($K_{geo}$) and semantic ($K_{sem}$) graphs,as visualized in Figure \ref{fig:sensitivity}.

The influence of model dimensionality is analyzed in the upper row of Figure \ref{fig:sensitivity}. For the prediction head dimension $d_{pred}$, performance improves substantially as the dimension increases from 32 to 128, indicating that a sufficient channel width is required to decode the complex latent representations into accurate pollutant concentrations. However, further increasing $d_{pred}$ beyond 128 leads to a slight degradation in performance, suggesting that an over parameterized head may induce overfitting on the training noise. A similar trend is observed for the Fourier encoding dimension $d_{fourier}$, where the optimal spatial resolution is achieved at 32. Lower dimensions fail to capture high frequency spatial variations, while excessively high dimensions  introduce high-frequency artifacts that disrupt the learning of global spatial smoothness.

Regarding the graph topology structure, the bottom row of Figure \ref{fig:sensitivity} illustrates the impact of neighbor aggregation sizes. For geographic neighbors $K_{geo}$, the error rate reaches its minimum at $K_{geo}=10$. Using fewer neighbors ($K_{geo}<5$) leads to under utilization of local context, whereas aggregating too many neighbors ($K_{geo}>15$) introduces noise from distant, uncorrelated stations, diluting the valid local signals. Crucially, the analysis of semantic neighbors $K_{sem}$ validates the effectiveness of our inductive design. The inclusion of semantic connections significantly outperforms the baseline with no semantic graph, proving that non-Euclidean correlations are essential, after which the inclusion of less similar stations diminishes the quality of the learned topology. Consequently, we adopt the optimal setting for all main experiments, ensuring a robust trade off between feature expressiveness and noise suppression.

\begin{figure}[htbp!]
    \centering
    \includegraphics[width=0.85\linewidth]{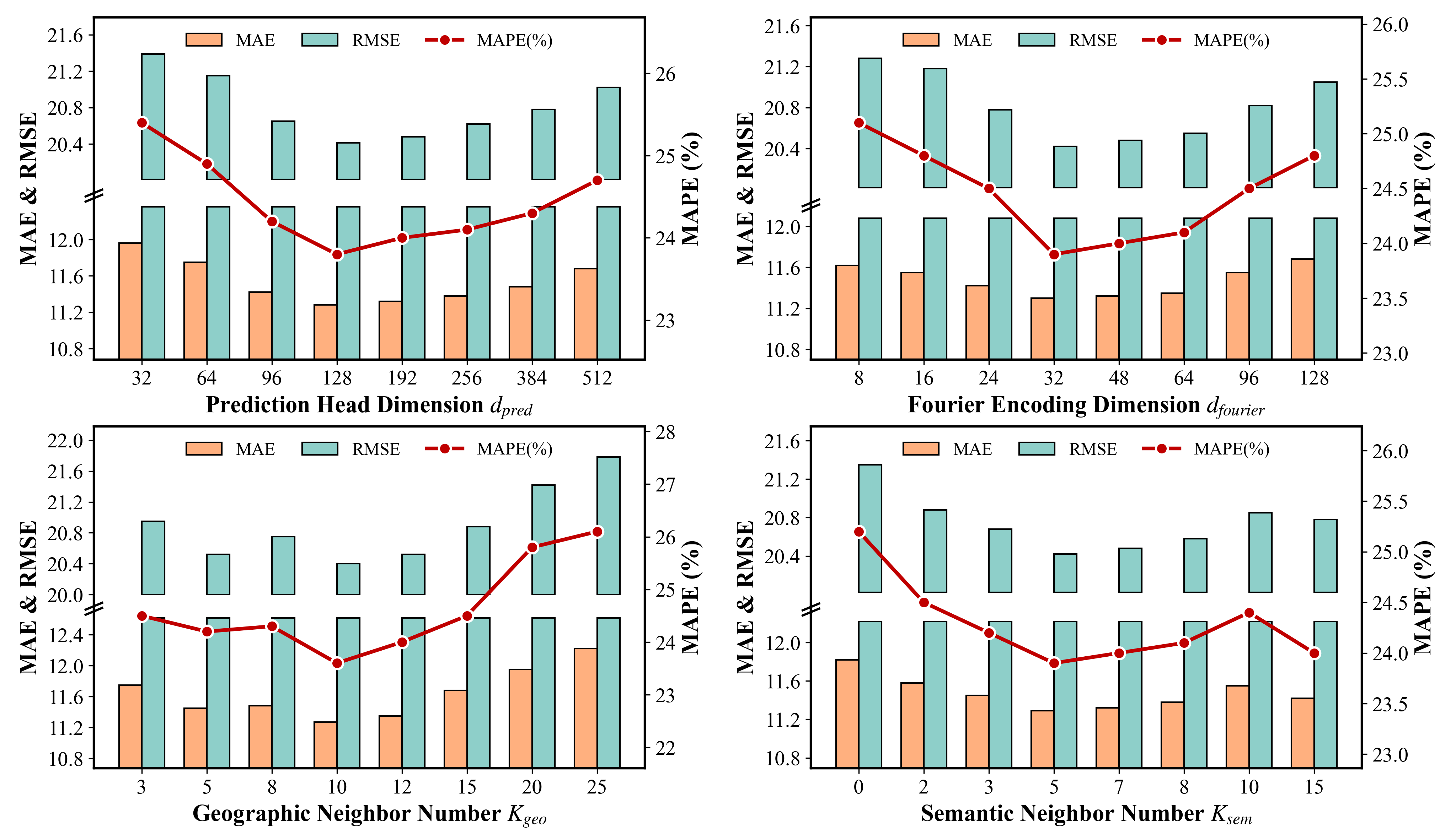}
    \caption{Parameter sensitivity analysis.}
    \label{fig:sensitivity}
\end{figure}

\section{Station Level Feature Visualization}
\label{app:station_features_vis}

To demonstrate the rich geospatial context captured by our static feature engineering pipeline, we present detailed visualizations of two representative metropolitan regions: Tehran (Iran) and Mexico City (Mexico). These case studies illustrate how terrain morphology, atmospheric dynamics, and monitoring station placement interact to shape local air quality patterns.

\begin{figure*}[htbp!]
    \centering
    \begin{subfigure}{0.32\linewidth}
        \includegraphics[width=\linewidth]{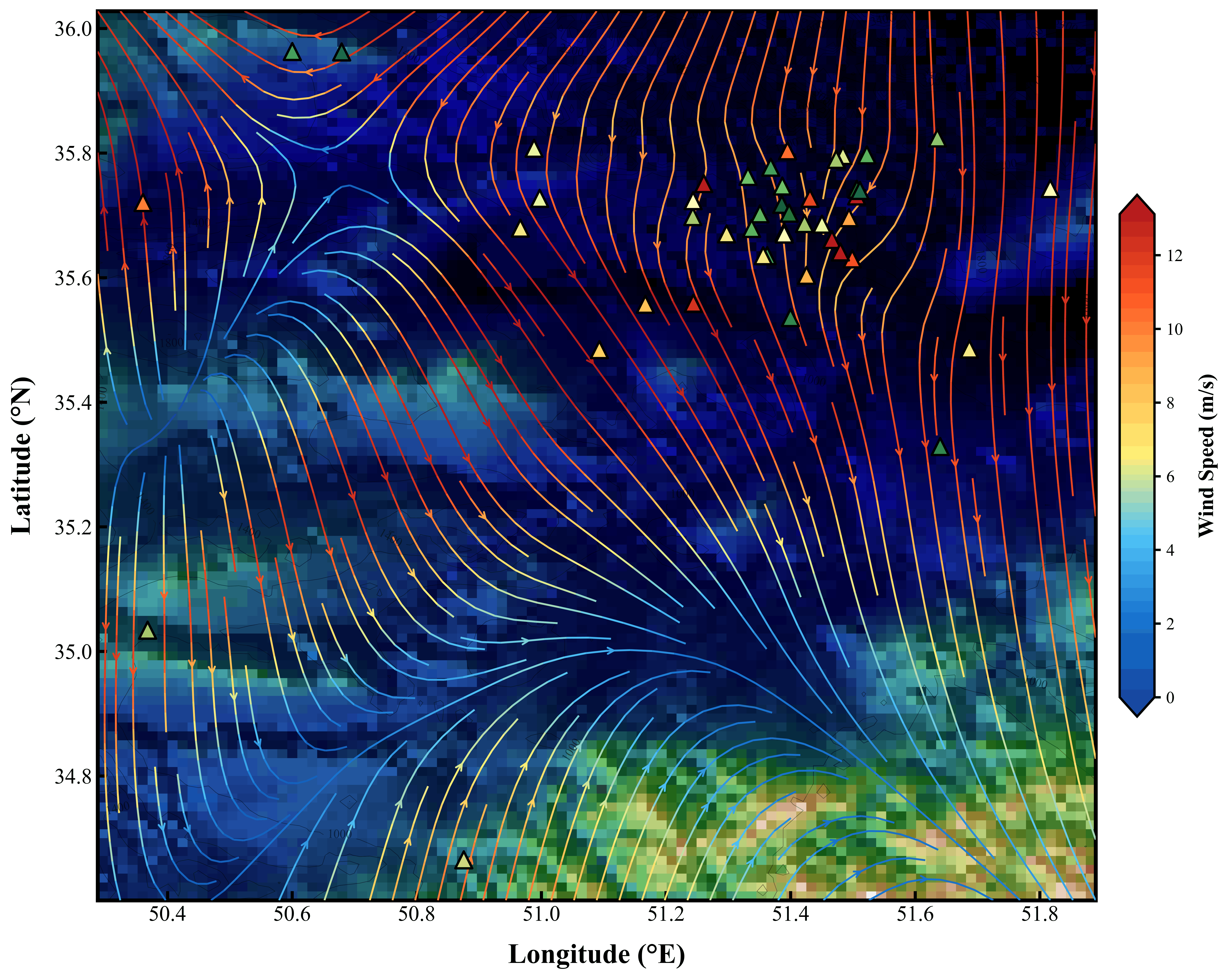}
        \caption{Tehran: Elevation Contour}
        \label{fig:tehran_elev_2d}
    \end{subfigure}
    \hfill
    \begin{subfigure}{0.32\linewidth}
        \includegraphics[width=\linewidth]{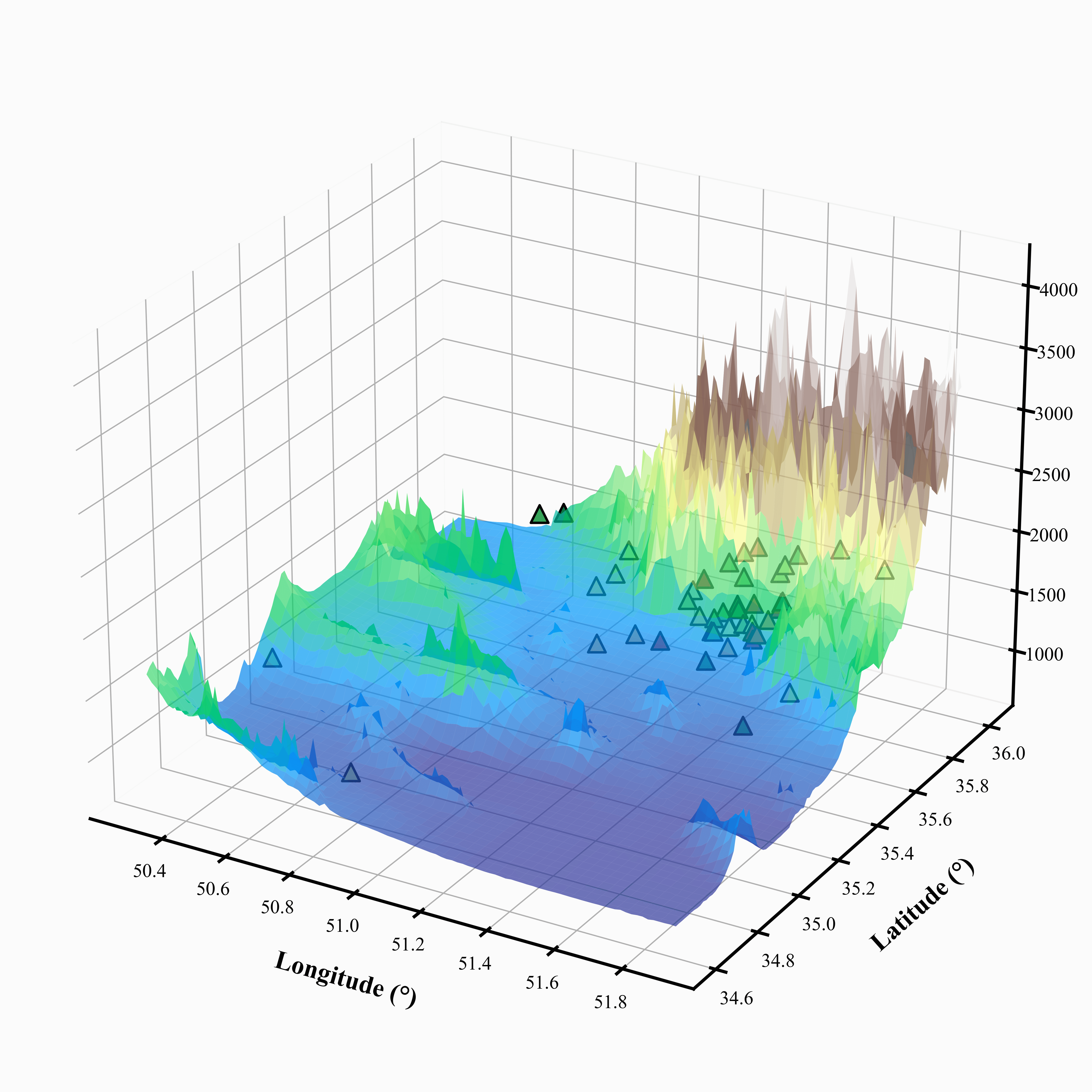}
        \caption{Tehran: 3D Terrain Surface}
        \label{fig:tehran_elev_3d}
    \end{subfigure}
    \hfill
    \begin{subfigure}{0.32\linewidth}
        \includegraphics[width=\linewidth]{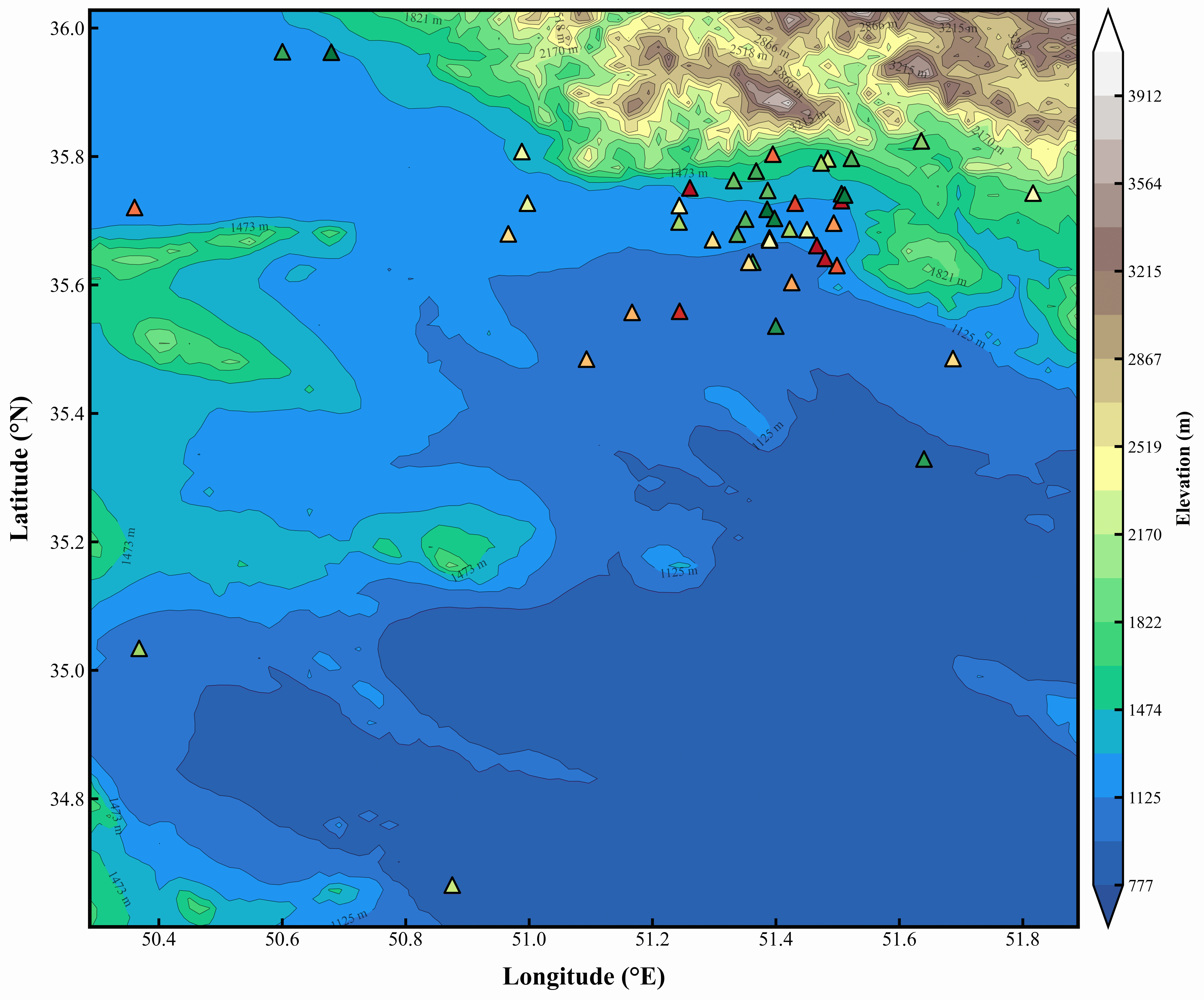}
        \caption{Tehran: Wind Field Streamlines}
        \label{fig:tehran_wind}
    \end{subfigure}
    
    \vspace{0.8em}
    
    \begin{subfigure}{0.32\linewidth}
        \includegraphics[width=\linewidth]{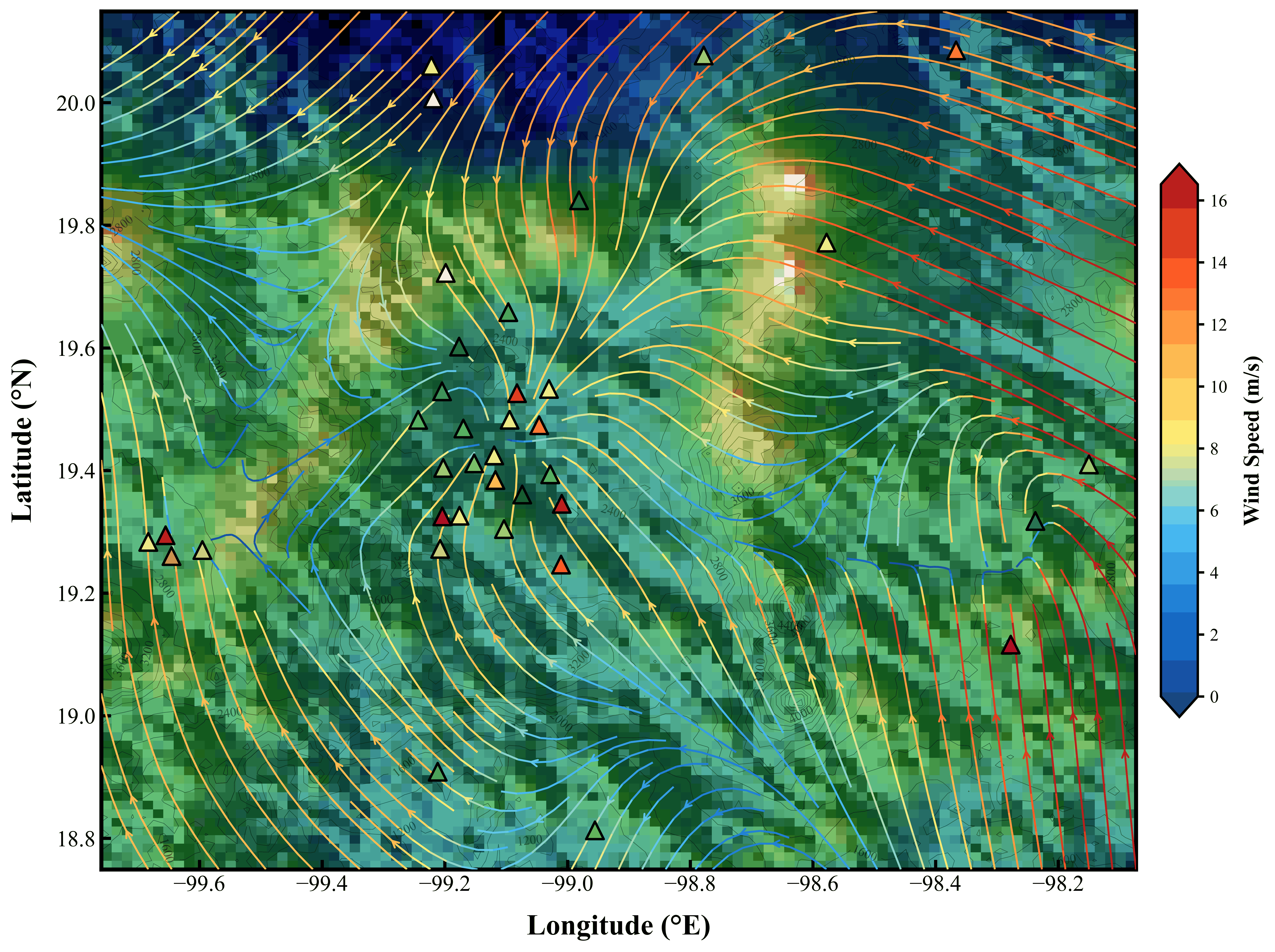}
        \caption{Mexico City: Elevation Contour}
        \label{fig:mexico_elev_2d}
    \end{subfigure}
    \hfill
    \begin{subfigure}{0.32\linewidth}
        \includegraphics[width=\linewidth]{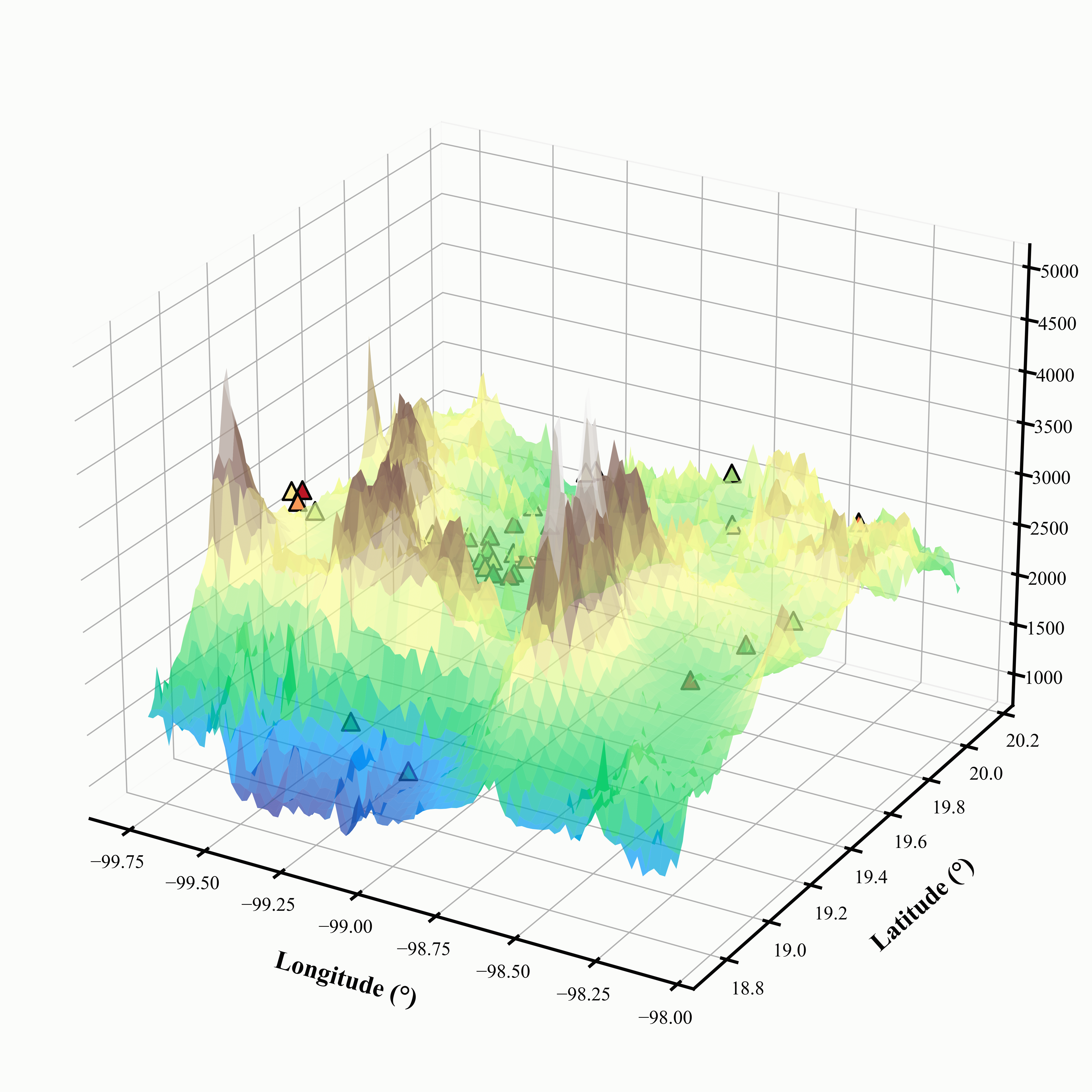}
        \caption{Mexico City: 3D Terrain Surface}
        \label{fig:mexico_elev_3d}
    \end{subfigure}
    \hfill
    \begin{subfigure}{0.32\linewidth}
        \includegraphics[width=\linewidth]{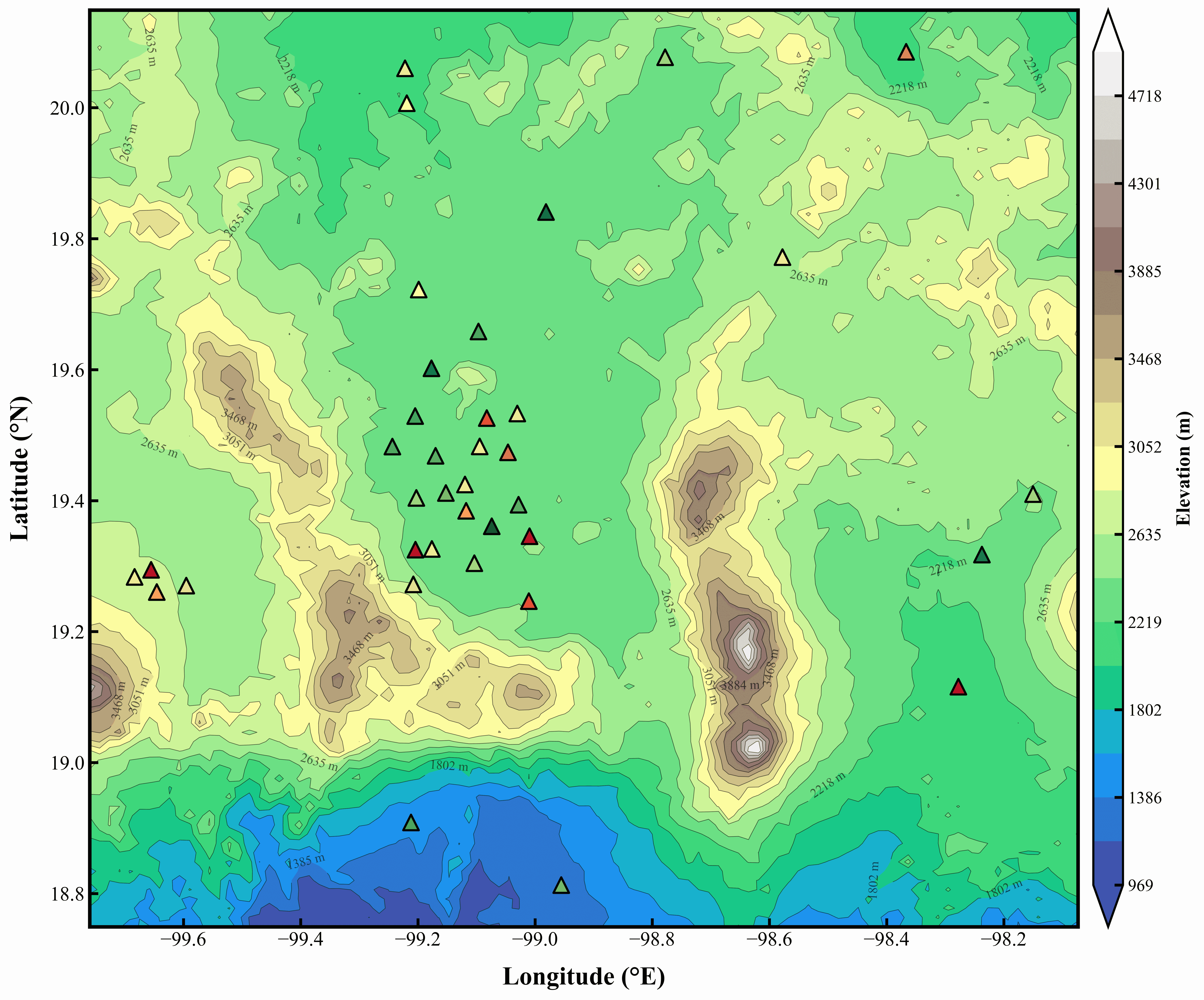}
        \caption{Mexico City: Wind Field Streamlines}
        \label{fig:mexico_wind}
    \end{subfigure}
    
    \caption{\textbf{Multi-dimensional geospatial feature visualization for two metropolitan regions.} Triangle markers ($\triangle$) denote monitoring stations, with colors indicating PM$_{2.5}$ concentration levels (yellow: low, red: high). \textbf{(a--c) Tehran, Iran:} Stations cluster in the northern foothills (1100--1500m) at the base of the Alborz Mountains ($>3500$m), which act as a barrier blocking northward pollutant dispersion. Wind streamlines reveal complex channeling effects through mountain passes. \textbf{(d--f) Mexico City, Mexico:} The metropolitan area sits in a high-altitude basin ($>2200$m) surrounded by volcanic peaks exceeding 4000m, creating a natural bowl that traps pollutants under frequent temperature inversions.}
    \label{fig:station_features}
\end{figure*}

\subsection{Interpretation of Station Level Features}
\label{sub:interp_station}

As illustrated in Figure~\ref{fig:station_features}, integrating multi-source geospatial features enables a comprehensive understanding of each monitoring station's environmental context:

\paragraph{Topographic Influence.}
The 2D contour maps (Figs.~\ref{fig:tehran_elev_2d}, \ref{fig:mexico_elev_2d}) and 3D terrain surfaces (Figs.~\ref{fig:tehran_elev_3d}, \ref{fig:mexico_elev_3d}) reveal how elevation gradients and terrain complexity directly impact pollutant accumulation. In both regions, monitoring stations are predominantly located in topographic basins where cold air pooling and temperature inversions frequently occur. The derived features---\texttt{elevation}, \texttt{terrain\_tpi} (Topographic Position Index), and \texttt{terrain\_roughness}---quantify these morphological characteristics, providing the model with critical information about each station's susceptibility to pollution trapping.

\paragraph{Atmospheric Transport Dynamics.}
The wind field visualizations (Figs.~\ref{fig:tehran_wind}, \ref{fig:mexico_wind}) demonstrate complex advection patterns governing pollutant transport. Streamlines colored by wind speed (0--16 m/s) highlight convergence zones (low speed, blue) where pollutants accumulate, and divergence zones (high speed, red) where dispersion is enhanced. Features such as \texttt{climate\_avg\_wind} and \texttt{climate\_avg\_wind\_dir} encode long-term climatological wind patterns, enabling the model to learn transport-mediated spatial dependencies.

\paragraph{Station-Pollution Correlation.}
The color coded station markers reveal a clear spatial correlation between topographic setting and observed pollution levels. Stations in low-lying areas with weak ventilation (blue streamlines) consistently exhibit higher PM$_{2.5}$ concentrations (red markers), whereas elevated stations with stronger wind exposure show lower readings (yellow markers). This pattern validates the physical relevance of our curated geospatial features and underscores the necessity of station level modeling.

By explicitly encoding these terrain and meteorological attributes, our model effectively learns heterogeneous environment-pollution relationships, enabling accurate predictions even for stations in complex topographic settings.

\section{Broader Impact Statement}
\label{sec:impact}

This research aims to advance global air quality forecasting, which has significant positive societal implications. Accurate air quality predictions can help protect public health by enabling timely warnings for vulnerable populations, support evidence based environmental policy making, and contribute to achieving sustainable development goals. The global scale approach specifically addresses the monitoring inequality between developed and developing regions, potentially democratizing access to air quality intelligence.

However, we acknowledge potential concerns. Model predictions should not replace ground truth measurements but rather complement existing monitoring infrastructure. Users should be aware of the inherent uncertainty in forecasting and avoid over reliance on predictions for critical health decisions without consulting local authorities.

While grid based Large Foundation Models (LFMs) have achieved remarkable success in global weather forecasting, their direct application to station level air quality prediction faces intrinsic limitations due to a fundamental misalignment in task definition and data ontology. The primary divergence lies in the nature of the training data; meteorological LFMs are predominantly trained on reanalysis datasets, which represent a numerical assimilation of atmospheric states rather than ground truth observations. This reliance on smoothed simulation data inherently dampens extreme localized pollution events—such as sudden emission spikes from wildfires or urban traffic—which are critical for health impact assessment but are often diluted in coarse grained grid cells (typically $0.25^\circ$, approx. $25 \times 25$ km). Consequently, these models tend to output a spatially averaged background state rather than the precise, high-variance exposure levels recorded by ground monitors.

Furthermore, the governing dynamics of the two domains differ significantly. Meteorological models primarily focus on solving fluid dynamics equations (e.g., Navier-Stokes) to simulate atmospheric motion. In contrast, air quality forecasting requires the explicit modeling of complex non-linear chemical kinetics (e.g., secondary aerosol formation) and high frequency anthropogenic activities. Most general purpose weather foundation models lack the specialized mechanisms to encode these localized emission sources and chemical interactions. Therefore, to bridge the gap between macro-scale atmospheric circulation and micro-scale human exposure, a dedicated station based inductive framework is essential. This approach preserves the fidelity of local observational data while effectively capturing the fine grained spatial heterogeneity that grid-based simulations inevitably smooth out.

\section{Code and Data Availability}
\label{sec:availability}
We acknowledge and thank OpenStreetMap (\url{https://www.openstreetmap.org/}); Open-Meteo  (\url{https://open-meteo.com/}); and Natural Earth (\url{https://www.naturalearthdata.com/}). To facilitate reproducibility and future research, the complete source code, and the part of curated WorldAir dataset will be made publicly available via GitHub upon acceptance.

\section{Case Study}
\label{sec:case_study}

\begin{figure*}[htbp]
    \centering
    \includegraphics[width=\textwidth]{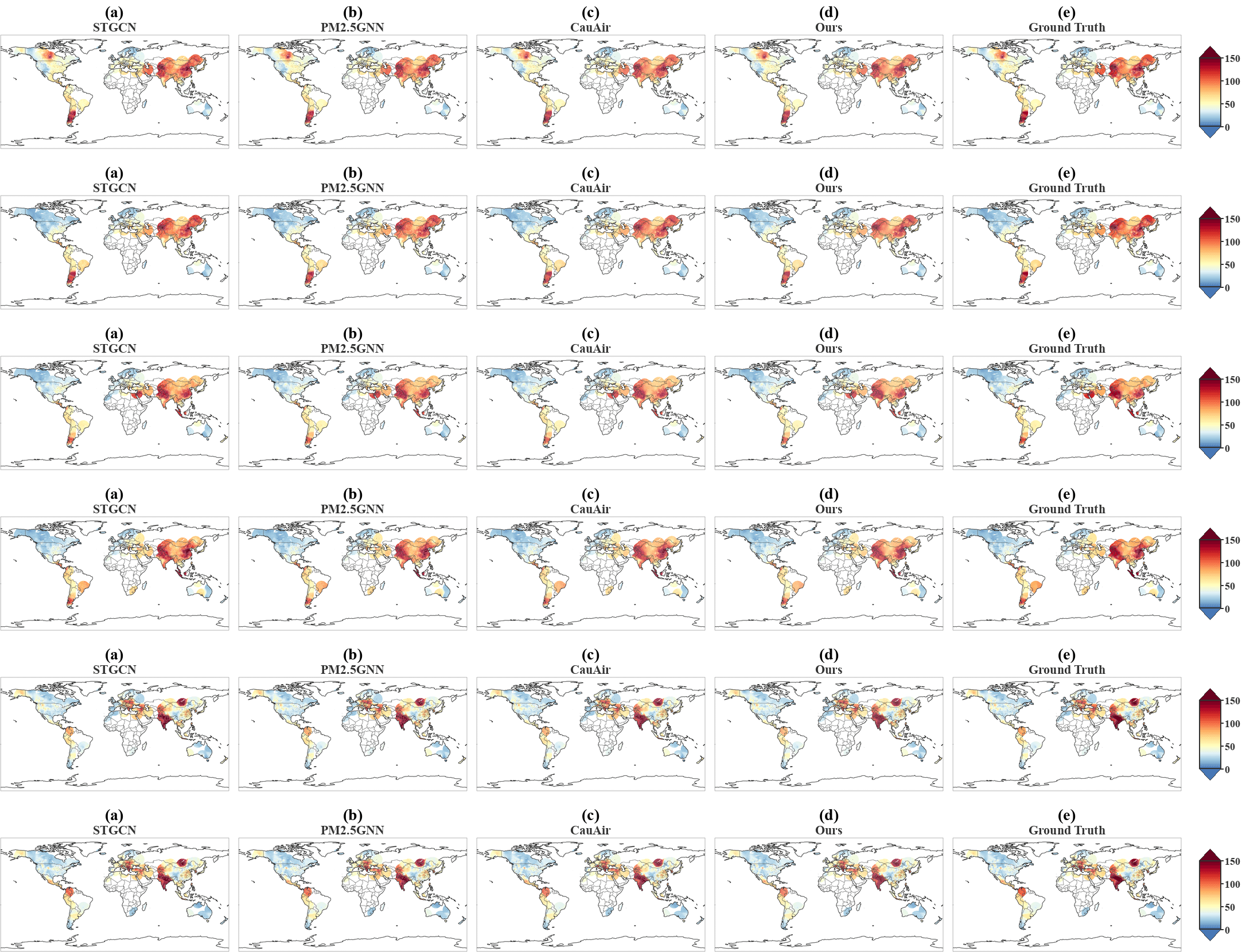}
    \caption{Spatial PM$_{2.5}$ predictions comparison (Part 1). (a) STGCN, (b) PM2.5GNN, (c) CauAir, (d) Ours, (e) Ground Truth.}
    \label{fig:spatial_comp_1}
\end{figure*}

\begin{figure*}[htbp]
    \centering
    \includegraphics[width=\textwidth]{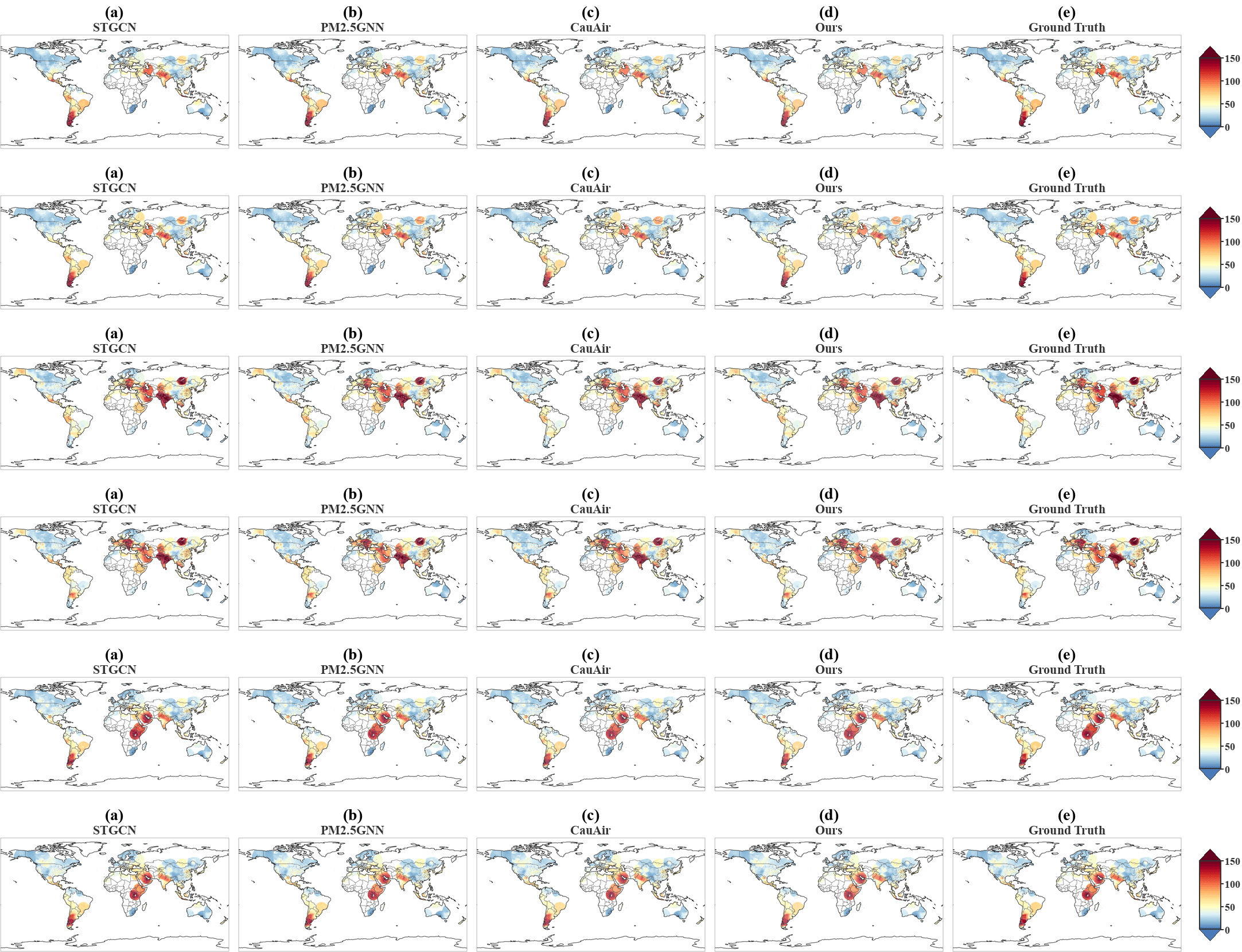}
    \caption{Spatial PM$_{2.5}$ predictions comparison (Part 2). Our method shows better alignment with ground truth in capturing pollution hotspots.}
    \label{fig:spatial_comp_2}
\end{figure*}

\begin{figure*}[htbp]
    \centering
    \includegraphics[width=\textwidth]{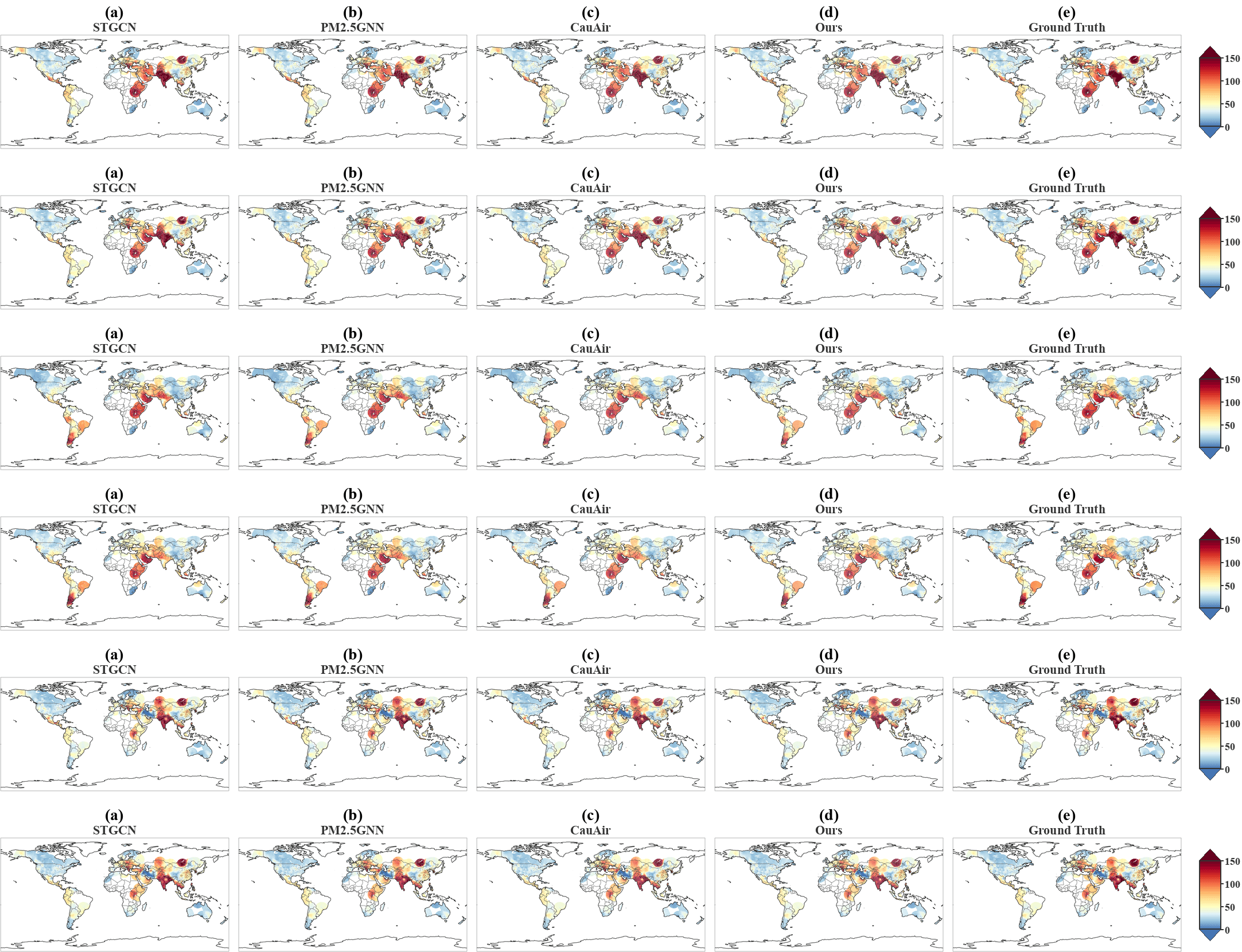}
    \caption{Spatial PM$_{2.5}$ predictions comparison (Part 3). The proposed approach maintains consistent accuracy across seasonal transitions.}
    \label{fig:spatial_comp_3}
\end{figure*}

\begin{figure*}[htbp]
    \centering
    \includegraphics[width=\textwidth]{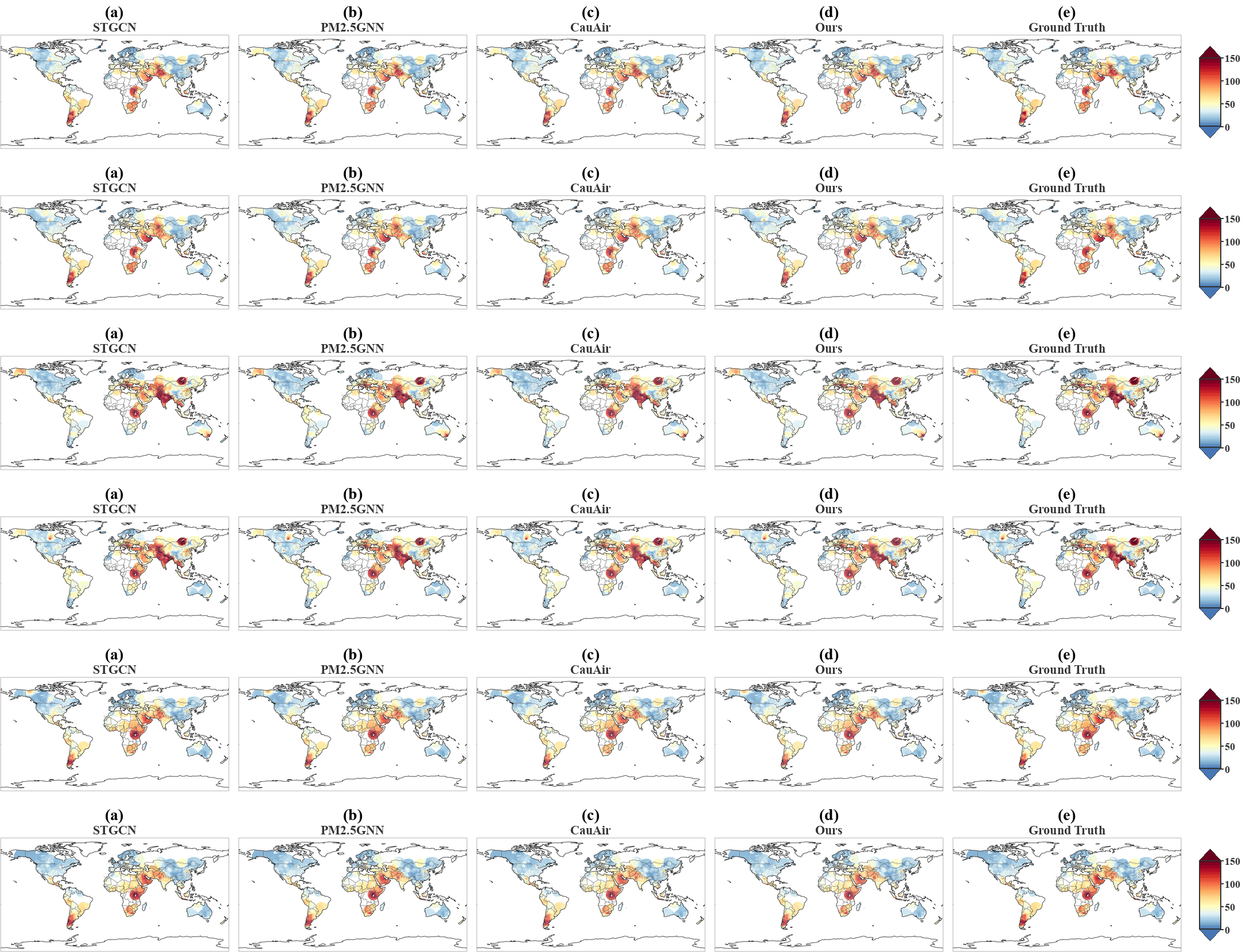}
    \caption{Spatial PM$_{2.5}$ predictions comparison (Part 4). Regional patterns in Europe and North America are well preserved by our method.}
    \label{fig:spatial_comp_4}
\end{figure*}

\begin{figure*}[htbp]
    \centering
    \includegraphics[width=\textwidth]{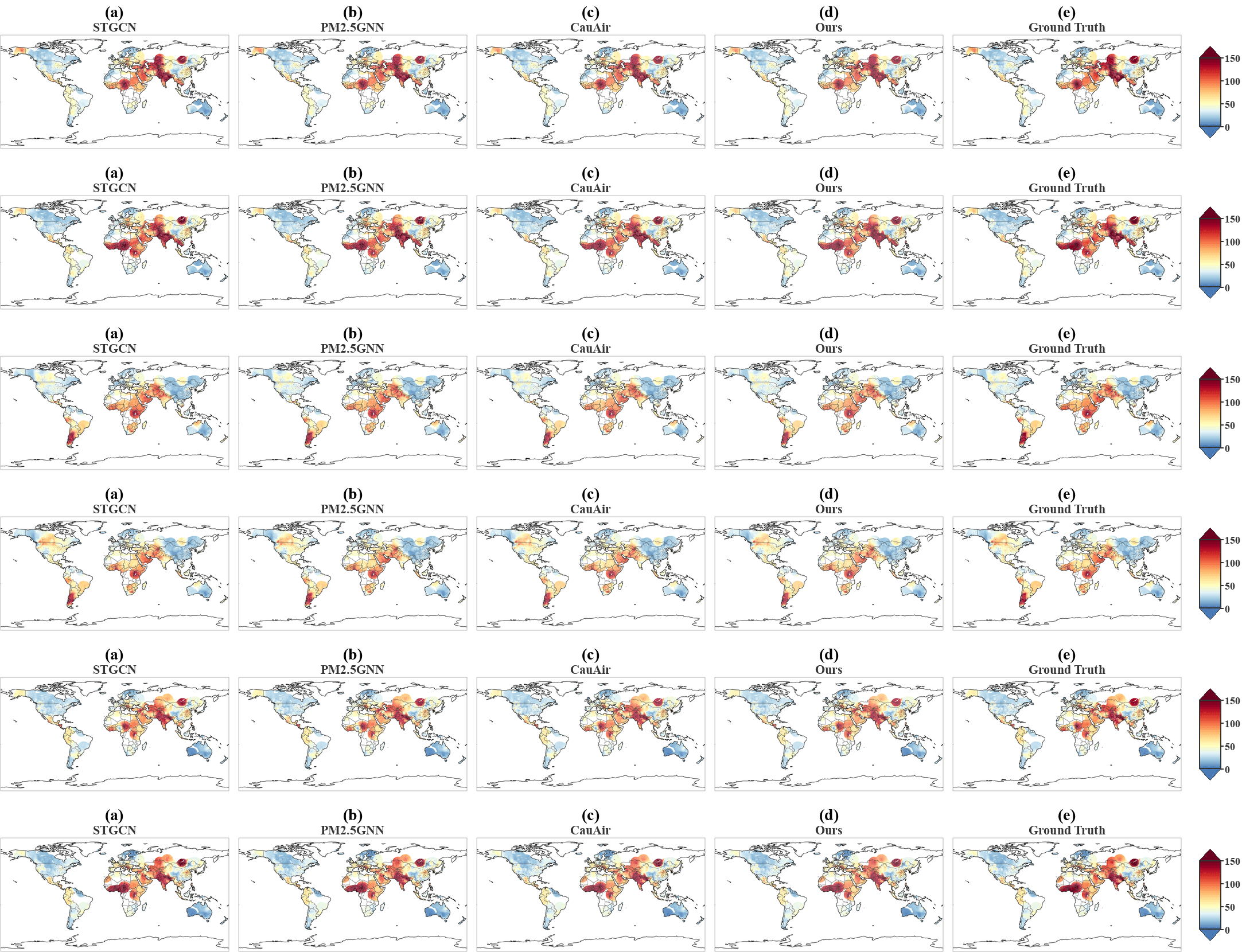}
    \caption{Spatial PM$_{2.5}$ predictions comparison (Part 5). Overall, our method achieves superior visual fidelity compared to baseline approaches.}
    \label{fig:spatial_comp_5}
\end{figure*}

Figs.~\ref{fig:spatial_comp_1}--\ref{fig:spatial_comp_5} present qualitative comparisons of global PM$_{2.5}$ predictions across five methods over 40 bi-weekly periods from 2015 to 2024.

\newpage
\section{Discussion}
\label{sec:discussion}

The Regional Barrier in air quality forecasting is fundamentally a data barrier. Transductive models, which memorize specific station IDs, inherently fail in the Global South where monitoring infrastructure is sparse or non-existent. OmniAir's inductive paradigm—learning station identities from environmental contexts rather than historical sequences—represents a step towards democratizing environmental intelligence. 

By disentangling the invariant physicochemical laws of pollution dispersion from specific geographic locations, our framework enables knowledge transfer from data rich regions to data sparse regions, offering a viable solution for estimating exposure risks in underserved communities without requiring extensive historical data accumulation.

Future work will focus on integrating dynamic multimodal data streams, such as satellite derived aerosol optical depth (AOD) and real time human mobility data, to enhance robustness against such non-stationary events.

\end{document}